\documentclass[letter, 11pt, reqno]{amsart}
\usepackage[margin=1in]{geometry}
\usepackage[T1]{fontenc}
\usepackage{lmodern}
\usepackage[utf8]{inputenc}

\usepackage{amsmath}
\numberwithin{equation}{section}
\usepackage{mathtools}
\usepackage{amssymb}
\usepackage{amsthm}
\usepackage{bbm}
\usepackage[foot]{amsaddr}
\usepackage{mathrsfs}
\usepackage{enumerate}
\usepackage{graphicx}
\usepackage{enumitem}
\usepackage[ruled,vlined]{algorithm2e}
\usepackage{color}
\usepackage{dsfont}
\usepackage{booktabs}
\usepackage[dvipsnames]{xcolor}
\usepackage{hyperref}
\hypersetup{
    colorlinks   = true,
    urlcolor     = blue,
    linkcolor    = blue,
    citecolor   = ForestGreen
}

\usepackage{caption}
\usepackage{subcaption}

\theoremstyle{plain}
\newtheorem{theorem}{Theorem}[section]
\newtheorem{lemma}[theorem]{Lemma}
\newtheorem{proposition}[theorem]{Proposition}
\newtheorem{corollary}[theorem]{Corollary}

\theoremstyle{definition}
\newtheorem{definition}{Definition}[section]
\newtheorem{assumption}[theorem]{Assumption}

\theoremstyle{remark}
\newtheorem{example}{Example}[section]
\newtheorem{remark}[theorem]{Remark}

\newcommand{\RR}{\mathbb{R}}

\newcommand{\Id}{\mathds{1}}
\newcommand{\EE}{\mathbb{E}}
\newcommand{\PP}{\mathbb{P}}

\newcommand{\Sd}{\mathbb{S}}

\newcommand{\ZZ}{\mathbb{Z}}

\newcommand{\diag}{\mathop{\mathrm{diag}}}

\newcommand{\dd}{\mathrm{d}}

\newcommand{\tr}{\mathrm{tr}}
\newcommand{\spanfn}{\mathrm{span}}

\newcommand{\one}{\boldsymbol{1}}

\newcommand{\av}{\boldsymbol{a}}
\newcommand{\bv}{\boldsymbol{b}}

\newcommand{\uv}{\boldsymbol{u}}
\newcommand{\vv}{\boldsymbol{v}}

\newcommand{\Csf}{\mathsf{C}}

\newcommand{\Psf}{\mathsf{P}}
\newcommand{\Qsf}{\mathsf{Q}}

\newcommand{\Tsf}{\mathsf{T}}

\newcommand{\Psiv}{\boldsymbol{\Psi}}

\newcommand{\cA}{\mathcal{A}}
\newcommand{\cB}{\mathcal{B}}
\newcommand{\cC}{\mathcal{C}}
\newcommand{\cD}{\mathcal{D}}
\newcommand{\cE}{\mathcal{E}}
\newcommand{\cF}{\mathcal{F}}
\newcommand{\cG}{\mathcal{G}}
\newcommand{\cH}{\mathcal{H}}

\newcommand{\cL}{\mathcal{L}}
\newcommand{\cM}{\mathcal{M}}
\newcommand{\cN}{\mathcal{N}}

\newcommand{\cR}{\mathcal{R}}

\newcommand{\cV}{\mathcal{V}}
\newcommand{\cW}{\mathcal{W}}
\newcommand{\cX}{\mathcal{X}}
\newcommand{\cY}{\mathcal{Y}}

\newcommand{\sD}{\mathscr{D}}

\newcommand{\sL}{\mathscr{L}}

\newcommand{\sP}{\mathscr{P}}

\newcommand{\E}{\EE}

\newcommand{\opnorm}[1]{\lVert #1\rVert_{\mathrm{op}}}
\newcommand{\Fnorm}[1]{\lVert #1\rVert_{F}}
\newcommand{\abs}[1]{\lvert #1\rvert}
\newcommand{\norm}[1]{\lVert #1\rVert}

\newcommand{\Dd}{\mathrm{D}}

\newcommand{\op}{\mathrm{op}}

\allowdisplaybreaks

\begin{document}

\title[Ballistic limits for adaptive SGD]{High-dimensional limit theorems for adaptive variants of SGD I: Ballistic limits}
\author{Aukosh Jagannath}
\address[Aukosh Jagannath]{Department of Statistics and Actuarial Science, Department of Applied Math Science, and Cheriton School of Computer Science, University of Waterloo, Canada}
\email{a.jagannath@uwaterloo.ca}

\author{Varnan Sarangian}
\address[Varnan Sarangian]{Department of Statistics and Actuarial Science, University of Waterloo, Canada}
\email{v3sarang@uwaterloo.ca}
\date{\today}

\begin{abstract}
    We develop the high-dimensional effective dynamics for adaptive, momentum-based variants of Stochastic Gradient Descent (SGD). These effective dynamics are given by a Markov process. They key observation is that the non-Markovianity in the algorithm due to the exponential moving average and preconditioning can be treated using the method of averaging. Our approach applies to a broad class of algorithms including to Adaptive momentum estimation (ADAM), Momentum orthogonalized by Newton-Schultz (MUON), and its variants. 
    It also applies to a broad class of learning tasks, such at tensor PCA, Gaussian projective models, teacher-student networks, and a multi-layer neural network trained on the XOR problem using a hybrid of MUON and ADAM as commonly done in practice. 
    We demonstrate how to this technique can give precise, quantitative comparisons between algorithms. In particular, we find interesting phenomena, such as the \emph{cross-over phenomenon} where online SGD has better generalization than ADAM early in training until the learning curves cross and ADAM dominates. Furthermore, we prove a version of the \emph{Edge of Stability} phenomenon. In particular, we introduce an analogue of the infinity Laplacian that sharply characterizes the Edge of Stability.
\end{abstract}
\maketitle

\section{Introduction}
\label{sec:intro}

Since its introduction by Robbins--Monro \cite{robbins1951stochastic}, stochastic gradient descent and its many variants have become the go-to optimization algorithms for modern machine learning \cite{goodfellow2016deep}.
While there is a rich classical theory of stochastic approximation algorithms (see below), the modern high-dimensional regime is comparatively less understood, though there has been recent major progress. In this paper, we develop the high-dimensional scaling limits of adaptive, momentum based algorithms which are most commonly used in practice.
In particular, our work applies to Adaptive Moment estimation (ADAM) \cite{kingma2015adam} and Momentum Orthogonalized by Newton-Schultz (MUON) \cite{jordan2024muon} which have emerged as \emph{the} dominant training algorithms in practice \cite{devlin2019bert,liu2025muon,kimi2025k2,loshchilov2019decoupled,schmidt2021crowded}. See Section~\ref{subsec:adaptive-momentum-based-algs} below for a (non-exhaustive) list of algorithms to which our results apply.

There is by now an exhaustive literature for scaling limits for online SGD both in the classical and the high-dimensional scaling regimes. In the classical asymptotic regime, it was established that the algorithm trajectories for online SGD converge to solutions of population gradient flow in the small step-size limit \cite{benaim1999dynamics, benveniste1990adaptive, kushner2003stochastic,ljung1977analysis,robbins1951stochastic}. More refined information is known such as functional central limit theorems \cite{benveniste1990adaptive,kushner2003stochastic,mcleish1976functional} and large deviation principles \cite{dupuis1988large}. In the high-dimensional scaling regime---where the dimension of the parameter space, dimension of the data space, and inverse step size tend to infinity together---it was established that additional corrections to the scaling limit emerge at a critical step size scaling. These corrections can dramatically alter the dynamics. These ODE limits were first derived in the work of Saad--Solla \cite{SaadSolla1995Exact,SaadSolla1995Committee} for teacher-student networks and, more recently, derived in a broad range of models by many others \cite{GoldtEtAl2020TeacherStudent, RefinettiEtAl2021Mixtures,veiga2023phase,WangMattinglyLu2017Scaling}. In \cite{ben2022high,BenArousGheissariJagannath2024Effective}, the effective dynamics approach was introduced to give a rigorous understanding of the scaling limits of a finite collection of summary statistics, where both both ODE limits and SDE limits were developed. These are called \emph{ballistic} and \emph{diffusive} regimes respectively. Ballistic limits in settings where an infinite-dimensional summary statistics are required, such as settings with complex covariance profiles, were studied in \cite{collinswoodfin2024highdim,collinswoodfin2024onepass, paquette2021sgd,paquette2025homogenization}. Universality and non-universality of ballistic and diffusive limits were studied in \cite{GheissariJagannath2025Universality,Goldt-et-al-Gaussian-Equivalence-Conjecture,Goldt-PRX-universality}. There is also a deep related theory regarding sample complexity in these problems, see, e.g., \cite{AbbeBoixAdseraMisiakiewicz2022Staircase, AbbeBoixAdseraMisiakiewicz2023Leap, BenArousGheissariJagannath2021Online, DamianLeeSoltanolkotabi2022Representations}, and works on martingale CLTs \cite{anastasiou2019normal}.
Ballistic and diffusive limits of online SGD have been used to give an analytical understanding of empirical phenomena including scaling laws \cite{ferbach2025dimension,paquette2024scaling}, spectral alignment \cite{BenArousEtAl2025Spectral,BenArousEtAl2026Geometry}, and the lottery ticket hypothesis \cite{BenArousGheissariJagannath2024Effective, FrankleCarbin2019Lottery}.

\begin{figure}[t]
    \centering
    \includegraphics[width=0.8\linewidth]{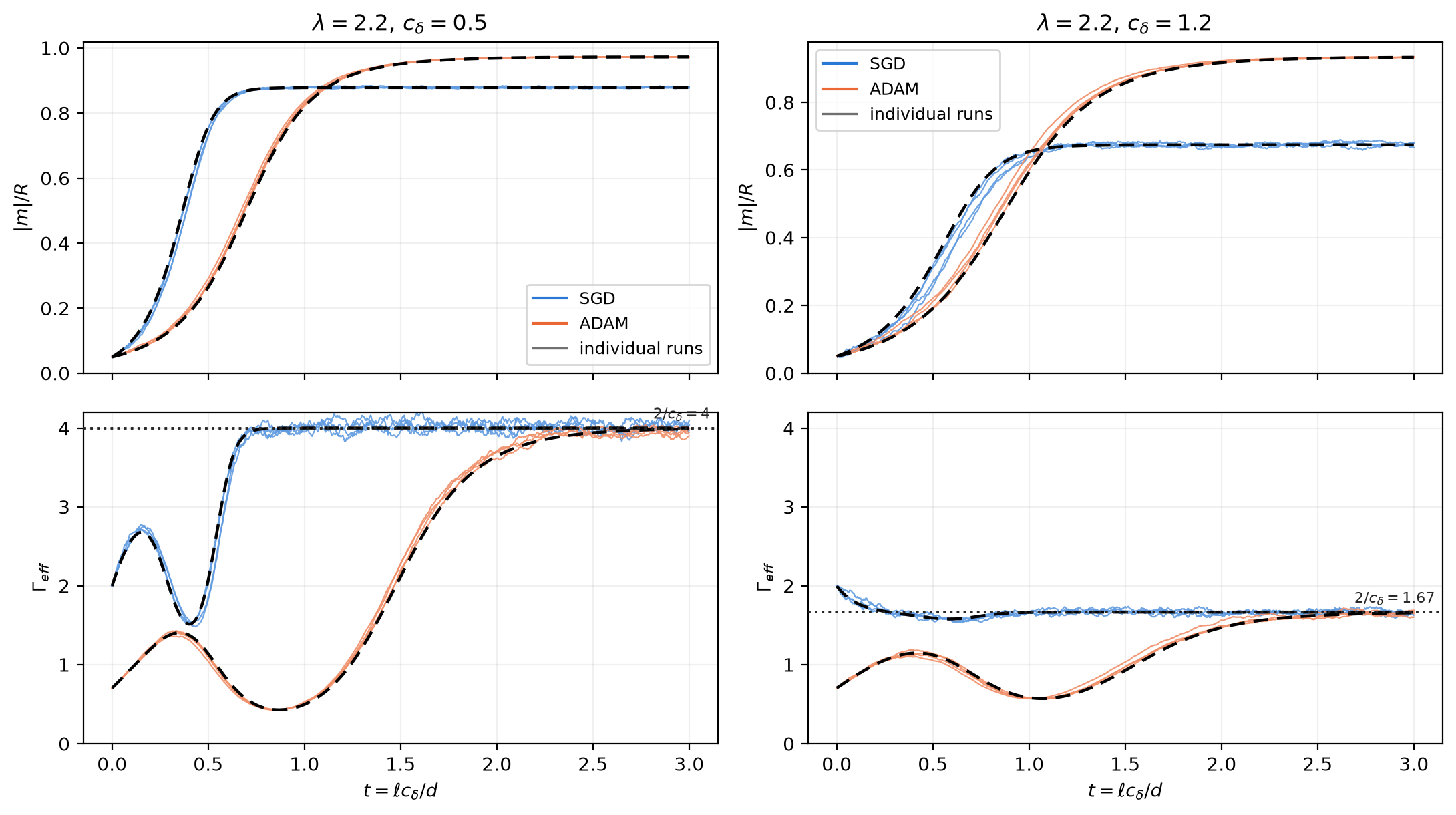}
    \caption{Demonstration of edge of stability phenomena for Matrix PCA ($k=2$, $d=10000$) from a warm initialization $a_0=0.05$, $R_0=1$, comparing online SGD and  ADAM ($\beta_1=0.9$, $\beta_2=0.999$). On the $x$-axis we plot time $t = \ell \delta_n$ up to $T = 3$. On the $y$-axis we plot the cosine alignment $a = |m| / R$ (left) and the population risk (right). The limiting ballistic ODEs (dashed curves) are overlaid on ten algorithmic trajectories (solid lines). The top presents the alignment at different step sizes whereas the bottom presents the effective infinity Laplacian $\Gamma^{\text{eff}}$ which converges to $c_\delta / 2$ once the population risk(s) plateau thus demonstrating the Edge of Stability phenomenon.}
    \label{fig:pca-stability-plot}
\end{figure}

While many of the above mentioned works in the classical asymptotics regime apply to stochastic algorithms broadly speaking, for adaptive, momentum-based variants of SGD, the classical asymptotic regime has been revisited in recent years which include studying continuous-time ballistic limits \cite{barakat2021adam, belotto2020general,dereich2025ode,kovachki2021continuous,ma2022qualitative,malladi2022sdes}, convergence rates \cite{gess2026exponential,reddi2019convergence}, escape times from saddle points \cite{liu2021diffusion,wang2021escaping}, implicit regularization effects \cite{cattaneo2024implicit,cattaneo2026minibatch,ghosh2023implicit}, and martingale central limit theorems \cite{dereich2026clt,gadat2018heavyball}.

In the high-dimensional setting, the picture for adaptive, momentum based algorithms is more sparse. It is typically algorithm and/or model specific. The abovementioned work of Paquette and Paquette \cite{paquette2021momentum} derived limiting risk dynamics for stochastic momentum methods for randomized least squares, including heavy-ball momentum. Related mini-batch results were studied in \cite{lee2022momentum}.  The work of Jones-McCormick and the two of us \cite{jagannath2026high} derived the ballistic limits and diffusive limits for a broad range of learning tasks for SGD with momentum and for SGD with a scalar renormalization (SGD-U). Closely related recent results describe high-dimensional risk dynamics for scalar adaptive learning rates for teacher-student networks \cite{collinswoodfin2024highline},  gradient clipping \cite{marshall2025clipping} for least-squares, and signSGD and stochastic MUON with momentum on matrix least-squares problems \cite{paquette2026phases,xiao2025signsgd}.

In this paper, we develop a unified approach to understanding the scaling limits of adaptive, momentum-based SGD in high dimensions for a broad range of algorithms (including versions of ADAM, MUON, finite Newton--Schulz updates, and SGD-M specified in Section~\ref{sec:setup}) and learning tasks. Here there is momentum and the preconditioner can be a matrix and also may have memory as well. To this end, we build on the effective dynamics approach of \cite{BenArousGheissariJagannath2024Effective}. For online SGD, the key observation in that work is that the parameter sequence is Markovian, allowing scaling limits to be developed using ideas from the ``martingale problem'' approach of Stroock--Varadhan \cite{stroock2006multidimensional}. For adaptive, momentum-based variants, however, the parameter sequence alone is generally not Markovian. We nevertheless obtain closed Markovian limiting dynamics.

To understand how, first observe that the memory in these algorithms is carried in exponential moving averages. The key idea is to treat the moving-average as a second, \emph{fast} time scale in a fast--slow dynamical system where the ``fast'' variable is the momentum and the ``slow'' variable is the parameter. Equipped with this insight, we use the \emph{method of averaging} \cite{pavliotis2008multiscale} --- in particular, we borrow ideas from the martingale approach pioneered by Papanicolaou--Stroock--Varadhan \cite{papanicolaou1977martingale} --- to develop ballistic limits. (In a companion work \cite{upcoming}, we use the homogenization approach to fast-slow systems to develop diffusive limits.) This yields a systematic approach to compute the effective dynamics of adaptive, momentum based algorithm. In particular, this allows us to compare the performance of different algorithms in a systematic fashion.

We demonstrate our results on a broad class of algorithms and learning tasks. In Section~\ref{subsec:adaptive-momentum-based-algs}, we give a detailed explanation of why many popular algorithms, such as ADAM and MUON, fall into our framework, including those that use different optimizers per layer in multi-layer neural networks. 
In Section~\ref{sec:applications}, we illustrate our approach by using the ballistic limits for several algorithms to sharply characterize empirical phenomena using several different data models.
First, in Section~\ref{subsec:app-crossover}, we discuss how these these can be used to exactly quantify comparisons between algorithms. In particular, the first phenomenon we discuss is the \emph{cross-over phenomenon}, where an algorithm that performs better early in training can be overtaken by another at later times. 
Secondly, in Section~\ref{subsec:app-eos}, we demonstrate how our results provide a rigorous characterization of the \emph{Edge of stability} phenomenon \cite{andreyev2026momentum,andreyev2025edge,cohen2024adaptive, cohen2021gradient}. In particular, we provide a generalization of infinity Laplacian, for momentum based algorithms which we call the \emph{effective infinity Laplacian} and prove that it sharply characterizes the Edge of Stability. 

The paper is organized as follows. Section~\ref{sec:setup} states the general framework and main convergence theorem, and explains the high-dimensional and Poisson corrections. Section~\ref{sec:applications} illustrates our effective dynamics by providing rigourous characterizations of the cross-over and edge of stability phenomena for concrete data models and algorithms. Section~\ref{sec:proof-main}--\ref{sec:martingale} details the proof of the main result. We then develop the effective dynamics explicitly in each for Tensor PCA (Section~\ref{sec:ex-tensor-pca}) and the general class of Gaussian projective models (Section~\ref{sec:ex-gpm}). In particular, we work out two specific projective models, namely 
Multiclass Logistic regression (Section~\ref{sec:multi-class-log-reg-model}) and the Gaussian XOR problem (Section~\ref{sec:gaus-xor-two-layer}).
Section~\ref{sec:compatible-algorithms} proves that all of the algorithms in the list provided in Section~\ref{subsec:adaptive-momentum-based-algs} are compatible with our framework.

\section{Setup and Main Results}
\label{sec:setup}

Suppose we are given an i.i.d. sequence of data $Y_1,Y_2,\ldots$ in $\cY_n\subseteq\RR^{d_n}$ with law $P_n$, and a loss function $L_n:\RR^{p_n}\times\cY_n\to\RR$. We seek to understand the evolution of the parameter $\theta\in\RR^{p_n}$ for a family of momentum-based online learning algorithms. Their momentum state space is $\cX_n\subseteq(\RR^b)^{p_n}$, where $b\in\{1,2\}$. We suppress the index $n$ when it is clear.

\begin{figure}[t]
    \centering
    \includegraphics[width=0.8\linewidth]{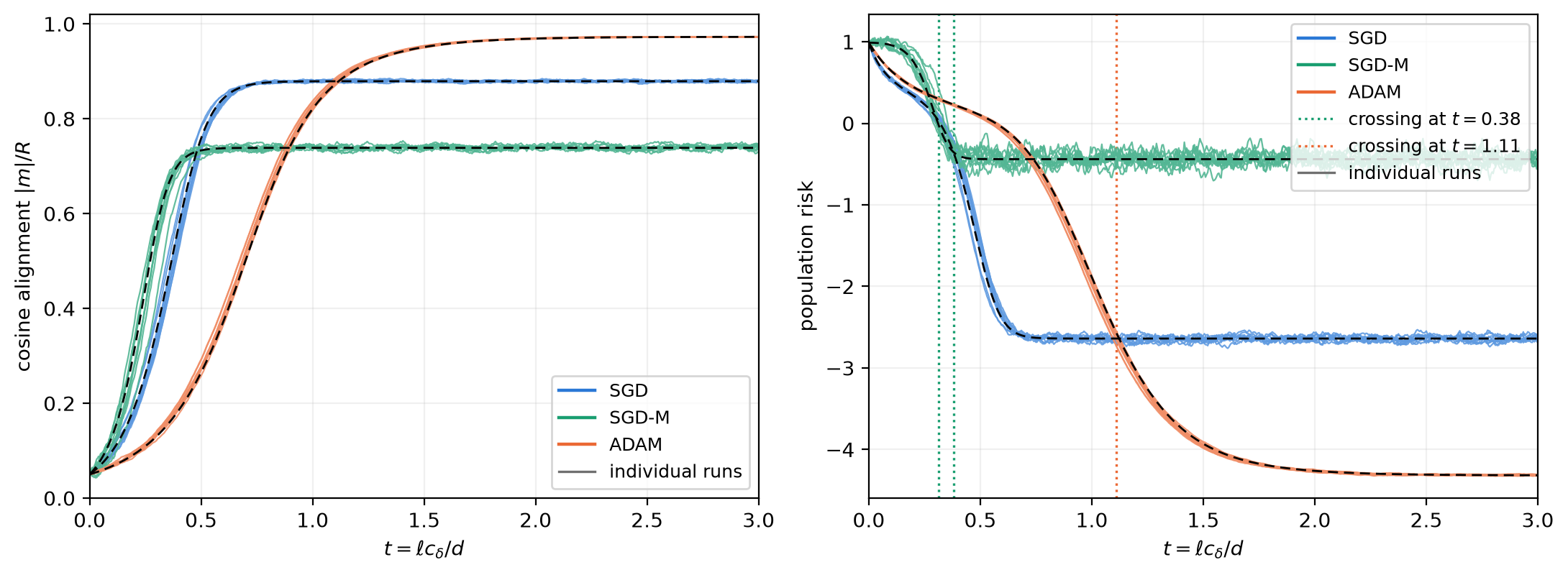}
    \caption{Demonstration of cross-over phenomena for Matrix PCA ($k=2$, $d=10000$) from a warm initialization $a_0=0.05$, $R_0=1$, comparing online SGD, SGD-M ($\beta=0.5$), and ADAM ($\beta_1=0.9$, $\beta_2=0.999$). On the $x$-axis we plot time $t = \ell \delta_n$ up to $T = 3$. On the $y$-axis we plot the cosine alignment $a = |m| / R$ (left) and the population risk (right). The limiting ballistic ODEs (dashed curves) are overlaid on finite-dimensional algorithmic trajectories (solid coloured curves). On the left, observe that we have separate cross-overs between the different pairs of the three algorithms. In particular, notice that on the right, SGD and SGD-M undergo two cross-overs where after the first cross-over, SGD-M briefly attains lower risk before being over-taken by SGD again.}
    \label{fig:pca-crossover-plot}
\end{figure}

\subsection{Adaptive, momentum-based SGD}\label{subsec:adaptive-momentum-based-algs}

Fix a diagonal matrix $B$, a learning rate $\delta>0$, a map $\psi_\theta:\cY\to\cX$, and a search direction $\rho:\cX\to\RR^p$. The diagonal entries of $B$ lie in $[0,1)$, with $\opnorm B\le\beta<1$ uniformly in $n$. The update is
\begin{align}
    \theta_\ell&=\theta_{\ell-1}-\delta\,\rho(X_\ell),
    \label{eq:unified-update-step}\\
    X_\ell&=\Tsf_{\theta_{\ell-1}}(X_{\ell-1},Y_\ell)
    :=BX_{\ell-1}+(I-B)\psi_{\theta_{\ell-1}}(Y_\ell).
    \label{eq:unified-momentum-update}
\end{align}
We call $X_\ell$ the momentum. Throughout, the initialization $\theta_0$ is independent of the subsequent data. For $S\ge1$, $r\ge1$, and $k\ge0$, define the norm
\[
    V_r(x):=1+\norm{x}^r,\qquad
    \norm f_{\cV_{r,S}^k}
    :=\sum\nolimits_{j=0}^k S^j\sup_{x\in\cX}
    \frac{\opnorm{\nabla_x^j f(x)}}{V_r(x/S)},
\]
with the convention $\nabla_x^0f=f$. When $S=1$, this is the usual norm for $C^k$ functions of polynomial growth of order $r$. The scale $S$ allows us to unify coordinate and matrix-valued search directions. In particular with $S=\sqrt p$ this is the polynomial growth extension of the $\cG^k$ norm from \cite{arous2020algorithmic}.

\begin{definition}[Adaptive, momentum-based SGD]\label{def:rho}
    We say that a learning algorithm $(\rho, \psi, B)$ is an \textit{adaptive, momentum-based SGD algorithm} if $\psi$ and $\rho$ satisfy the following,
    \begin{enumerate}[label=\textup{(\roman*)}]
        \item The map $\psi$ has the coordinate form $\psi_{\theta,i}(Y)=\Psi(\partial_iL(\theta,Y))$, $i\in[p]$, where $\Psi\in C^2(\RR;\RR^b)$ has at most quadratic growth.
        \item There is a partition $[p]=\bigsqcup_{\alpha=1}^N I_\alpha$, with $\cX=\bigoplus_{\alpha=1}^N\cX_{I_\alpha}$ and $S_\alpha=\sqrt{|I_\alpha|}$, such that $\rho(x)=(\rho_{I_\alpha}(x_{I_\alpha}))_{\alpha=1}^N$ and
        \begin{equation}\label{eq:rho-scale-bound}
            \max_{1\le\alpha\le N}
            S_\alpha^{-1}\norm{\rho_{I_\alpha}}_{\cV_{1,S_\alpha}^3}
            \le C_\rho.
        \end{equation}
    \end{enumerate}
\end{definition}

Let us record how several first-order, momentum, and preconditioned optimizers fit into \eqref{eq:unified-update-step}--\eqref{eq:unified-momentum-update}. Their structural assumptions are verified in Section~\ref{sec:compatible-algorithms}.

\begin{example}[Online SGD]
    Set $\cX=\RR^p$, $\psi_\theta(Y)=\nabla_\theta L(\theta,Y)$, $\rho(x)=x$, and $B=0$. 
\end{example}

\begin{example}[SGD-M]
    Set $\cX=\RR^p$, $\psi_\theta(Y)=\nabla_\theta L(\theta,Y)$, and $B=\beta I_p$, where $\beta\in[0,1)$. Define the search direction $\rho (x) = x / (1 - \beta )$ so that $M_\ell:=X_\ell/(1-\beta)$ satisfies $M_\ell=\beta M_{\ell-1}+\nabla L(\theta_{\ell-1},Y_\ell)$ and $\theta_\ell=\theta_{\ell-1}-\delta M_\ell$. 
\end{example}

\begin{example}[ADAM]
    Set $\cX=\RR^p\times\RR_{\ge0}^p$, write $x=(m,v)$, and take
    \[
        B=\diag(\beta_1I_p,\beta_2I_p),\qquad
        \psi_\theta(Y)=\bigl(\nabla_\theta L(\theta,Y),
        \nabla_\theta L(\theta,Y)^{\odot2}\bigr),
    \]
    where $\beta_1,\beta_2\in[0,1)$. For a fixed stabilizing constant $\epsilon>0$, the search direction is
    \begin{equation}\label{eq:adam-function}
        \rho(x)=\diag(v+\epsilon)^{-1/2}m.
    \end{equation}
\end{example}

\begin{example}[Exact MUON]\label{ex:muon-algo-def}
    For matrix-valued parameters $\theta\in\RR^{d\times q}$, where $q$ is fixed and $p=dq$, set $\cX=\RR^{d\times q}$, $B=\beta I_p$, and $\psi_\theta(Y)=\nabla_\theta L(\theta,Y)$. For $\epsilon>0$, consider the search direction
    \begin{equation}\label{eq:muon-function}
        \rho(X)=X\left(\frac{X^\top X}{d}+\epsilon I_q\right)^{-1/2}.
    \end{equation}
\end{example}

\begin{example}[MUON with Newton--Schulz iteration]
    Our framework also includes fixed-iteration Newton--Schulz implementations. In the same matrix setting, fix an integer $K\ge1$ and coefficients $\mathsf a>0$, $\mathsf b,\mathsf c\in\RR$. Define the polynomial $\phi(W)=W\bigl(\mathsf a I_q+\mathsf b W^\top W +\mathsf c(W^\top W)^2\bigr)$ and $\Phi_K=\phi^{\circ K}$ representing a $K$-fold composition. For $\epsilon_F>0$, define
    \begin{equation}\label{eq:muon-ns5-function}
        \rho_{\epsilon_F}^{\mathrm{NS}K}(X)
        =\sqrt d\,\Phi_K\!\bigl(N_{\epsilon_F}(d^{-1/2}X)\bigr),
        \qquad
        N_{\epsilon_F}(Z)=Z/\sqrt{\Fnorm Z^2+\epsilon_F}\,.
    \end{equation}
    We abbreviate this algorithm as MUON--NS$K$. The choice $(\mathsf a,\mathsf b,\mathsf c)=(3.4445,-4.7750,2.0315)$ and $K=5$ gives the polynomial used in MUON--NS5.
\end{example}

\begin{example}[Layer-wise combinations]
    We may use different optimizers on different subsets of the parameter. Suppose $\theta=(\theta^{(1)},\ldots,\theta^{(M)})\in\bigoplus_{m=1}^M\RR^{p_m}$, with fixed relative learning rates $c_1,\ldots,c_M>0$. For instance, in a two-layer network the first-layer matrix may use MUON and the second-layer vector ADAM.

    Embed the momentum states in $(\RR^2)^p$ by taking $\psi_\theta(Y)=(\nabla L(\theta,Y),\nabla L(\theta,Y)^{\odot2})$ and ignoring the second-moment coordinates in layers that do not use them. Set
    \[
        B=\bigoplus\nolimits_{m=1}^M B^{(m)},\qquad
        \rho(x)=\bigl(c_m\rho^{(m)}(x^{(m)})\bigr)_{m=1}^M.
    \]
    The moment parameters and stabilizing constants can differ between layers. Taking the union of their block partitions preserves \eqref{eq:rho-scale-bound}, with constant at most $\max_m c_mC_{\rho^{(m)}}$.
\end{example}

\begin{remark}[Decoupled weight decay]\label{rem:weight-decay}
    A differentiable penalty $\cC_\alpha$ gives the decoupled update
    \[
        \theta_\ell=\theta_{\ell-1}
        -\delta\bigl(\rho(X_\ell)+\nabla\cC_\alpha(\theta_{\ell-1})\bigr).
    \]
    For $\cC_\alpha(\theta)=\alpha\norm\theta^2/2$, the added direction is $\alpha\theta$. With the ADAM direction this is the decoupled weight-decay construction of \cite{loshchilov2019decoupled}. Lemma~\ref{lem:decoupled-drift} records the estimate justifying this addition.
\end{remark}

\subsection{Main result}

We study the regime $p_n,d_n\to\infty$ and $\delta_n\to0$. Suppose we are given summary statistics $\uv_n\in C^3(\RR^{p_n};\RR^k)$, where $k$ is fixed and $\uv_n=(u^{(1)},\ldots,u^{(k)})$. Our goal is to analyze the evolution of $\uv_n(\theta_\ell)$.
To establish a scaling limit, we make several assumptions on the tuple $( \uv, L_n , P_n , \rho )$. The first assumption is an upper bound on the norms of the derivatives of both the summary statistics $\uv$ and the loss function $L$. The second assumption is a similar upper bound along projection(s) of the search direction $\rho$. The third assumption is the most important as it asks for the evolution of the summary statistics to close asymptotically.

We now present the first assumption. Let $E_R\subseteq\RR^k$ be the open ball of radius $R$, and write $E_R^*=\uv_n^{-1}(E_R)$.

\begin{assumption}[Localizability on $A$]\label{assume:localize}
    For a set $A\subseteq\RR^p$ and every $R\ge1$, there is a constant $C_{A,R}$, independent of $n$, such that the following hold on $E_R^*\cap A$.
    \begin{enumerate}[label=\textup{(\roman*)},leftmargin=*]
        \item\label{item:A-stat} The summary statistics have uniformly bounded derivatives:
        \[
            \max_{i\le k}\sup_{\theta\in E_R^*\cap A}
            \bigl(\norm{\nabla u^{(i)}(\theta)}
            +\opnorm{\nabla^2u^{(i)}(\theta)}
            +\opnorm{\nabla^3u^{(i)}(\theta)}\bigr)
            \le C_{A,R}.
        \]
        \item\label{item:A-grad} For all deterministic directions $h,h'\in\RR^p$, the derivatives of the loss coordinates' satisfy,
        \begin{equation}\label{eq:P}
            \begin{aligned}
                \sup_{\theta\in E_R^*\cap A}\max_{i\le p}
                \norm{\partial_iL(\theta,Y)}_{L^{16}(P_n)}
                &\le C_{A,R},\\
                \sup_{\theta\in E_R^*\cap A}\max_{i\le p}
                \norm{\nabla_\theta\partial_iL(\theta,Y)[h]}_{L^{16}(P_n)}
                &\le C_{A,R}\norm h,\\
                \sup_{\theta\in E_R^*\cap A}\max_{i\le p}
                \norm{\nabla_\theta^2\partial_iL(\theta,Y)[h,h']}_{L^{16}(P_n)}
                &\le C_{A,R}\norm h\norm{h'}.
            \end{aligned}
        \end{equation}
    \end{enumerate}
\end{assumption}

\begin{figure}[t]
    \centering
    \includegraphics[width=0.8\linewidth]{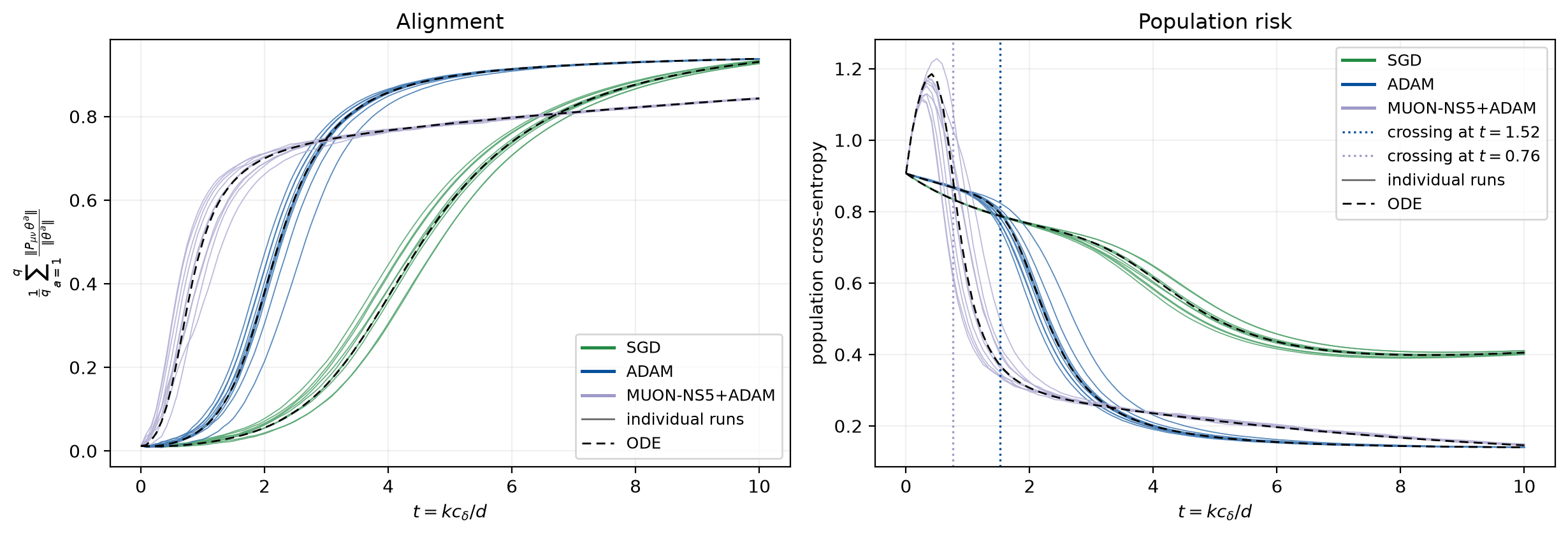}
    \caption{Demonstration of the cross-over phenomena for Gaussian XOR classification ($d=10000$) with softplus activation $\varphi (x) = \eta \log (1 + e^{x / \eta} )$, with $\eta = 0.05$, from random initialization, comparing online SGD, ADAM ($\beta_1=0.9$, $\beta_2=0.95$), and mixed MUON-NS5 first layer $(\beta_M = 0.9$) and ADAM second-layer networks. On the $x$-axis we plot time $t = \ell \delta_n$ up to $T = 10$. On the $y$-axis we plot the average subspace alignment for the first layer $\theta$ with $\spanfn\{ \mu, \nu \}$ (left) and the population risk with predicted cross-over times from the ODEs (right). The limiting ballistic ODEs (dashed curves) are overlaid on ten algorithmic trajectories (solid lines). Notice that on the right, multiple cross-over occurs between MUON-NS5 and the other algorithms in the population risk. On the left, we see that both ADAM and online SGD overtake MUON to attain better alignment despite MUON eventually attaining lower risk.}
    \label{fig:xor-crossover-plot}
\end{figure}

For a fixed $\theta$, define the \emph{fixed-parameter momentum process} to be the Markov chain on $\cX$ with transition kernel
\[
    K_\theta\varphi(x)=\EE_Y[\varphi(\Tsf_\theta(x,Y))],
    \qquad \sL_\theta=K_\theta-I.
\]
The contraction of the momentum recursion gives a unique invariant law $\pi_\theta$. (That this process is ergodic follows since $\beta < 1$. For more on this see Section~\ref{sec:poisson} below.) 
Under the moment conditions of Assumption~\ref{assume:localize} below---which imply that $\psi$ is integrable---it is the law of,
\begin{equation}\label{eq:fixed-momentum}
    \tilde X_0(\theta)=(I-B)\sum\nolimits_{j=0}^\infty B^j\psi_\theta(Y_{-j}),
\end{equation}
where $(Y_{-j})_{j\ge0}$ is an independent past drawn from $P_n$. 

\begin{assumption} \label{assume:A-proj}
    Suppose Assumption~\ref{assume:localize} holds. The stationary search direction has bounded projections along the summary statistics:
    \[
    \sup_{\theta\in E_R^*\cap A}\max_{i\le k}
        \norm{ \langle \nabla u^{(i)}(\theta) , \rho(\tilde X_0) \rangle}_{L^4(\pi_\theta)}
        \le C_{A,R}.
    \]
\end{assumption}

We now turn to our third assumption. For a test function $f=f(\theta,x)$, define the differential operators,
\begin{equation}\label{eq:operators}
    \begin{aligned}
        \cA f(\theta,x)
        &=K_\theta\bigl[ \langle \rho (\cdot ) , \nabla_\theta f(\theta,\cdot)  \rangle \bigr](x), \qquad \cL f(\theta,x)
        =\frac12K_\theta\bigl[
        \langle (\rho \otimes \rho ) (\cdot ), \nabla_\theta^2 f(\theta,\cdot) \rangle \bigr](x).
    \end{aligned}
\end{equation}
Furthermore, define the averaged operators by,
\begin{equation}\label{eq:averaged-operators}
    \begin{aligned}
        \tilde\cA_\theta f
        &=\pi_\theta(\cA f)
        =\pi_\theta\bigl[ \langle \rho (\cdot ) , \nabla_\theta f(\theta,\cdot)  \rangle\bigr], \qquad
        \tilde\cL_\theta f =\pi_\theta(\cL f)
        =\frac12\pi_\theta\bigl[
        \langle (\rho \otimes \rho ) (\cdot ), \nabla_\theta^2 f(\theta,\cdot) \rangle \bigr].
    \end{aligned}
\end{equation}
For a statistic $u$, define its momentum corrector $\chi_{1,\theta}^u$ as the centred solution of the Poisson equation
\begin{equation}\label{eq:main-first-corrector}
    -\sL_\theta\chi_{1,\theta}^u
    =-(I-\pi_\theta)\cA u,\qquad
    \pi_\theta\chi_{1,\theta}^u=0.
\end{equation}
The well-posedness of the Poisson equation is established in Section~\ref{sec:poisson}. Write $\chi_{1,\theta}^{\uv}=(\chi_{1,\theta}^{u^{(1)}},\ldots,\chi_{1,\theta}^{u^{(k)}})$.

\begin{assumption}[Closability on $A$]\label{assume:closable}
    Assumption~\ref{assume:localize} holds on $A$, the learning rates satisfy $\delta_n=O(p_n^{-1})$, and there is a locally Lipschitz function $\Psiv:\RR^k\to\RR^k$ such that, for every $R\ge1$,
    \begin{equation}\label{eq:closure}
        \sup_{\theta\in E_R^*\cap A}
        \|(-\tilde\cA_\theta+\delta_n\tilde\cL_\theta)\uv_n(\theta)
        -\delta_n\tilde\cA_\theta\chi_{1,\theta}^{\uv_n}
        -\Psiv(\uv_n(\theta))\|\rightarrow0.
    \end{equation}
    We call $\Psiv$ the \emph{effective drift}.
\end{assumption}

We now present the fourth and final assumption. In the previous assumptions, we required all of the estimates to hold potentially on a smaller region of the parameter space. Thus for the localization arguments to be sound, we require that the parameter stays in such regions of the parameter space beyond the time scale for which the scaling limit holds. To this end,
given a set $A \subseteq \RR^p$ and with $E_{R}^{*}$ defined prior to Assumption~\ref{assume:localize}, define the exit times $\tau_A=\inf\{\ell\ge0:\theta_\ell\notin A\}$ and $\tau_R=\inf\{\ell\ge0:\theta_\ell\notin E_R^*\}$.

\begin{assumption}\label{assume:A-subset-timescale}
    For open sets $A_n\subseteq\RR^{p_n}$, every $R\ge1$, and every fixed $T>0$,
    \[
    \PP\bigl(\tau_{A_n}\le\tau_R\land\lfloor T/\delta_n\rfloor\bigr)
        \rightarrow0.
    \]
\end{assumption}

For most algorithms and examples we consider, it suffices to take $A_n=\RR^{p_n}$ in which case Assumption~\ref{assume:A-subset-timescale} is tautological. This is the case, for example, for Muon and its variants, SGD-M, among others. The only algorithm for which we need to take $A_n\neq \RR^{p_n}$ is Adam. This is due to the fact that Adam depends on a choice of coordinates, i.e., it is a coordinate-wise algorithm. In this case, we will take $A_n$ to be the collection of \emph{coordinate-delocalized} sets in $\RR^{p_n}$, i.e., for fixed $\zeta \in [0, 1/2 )$, take $A_{n, \zeta} := \{ \theta \in \RR^{p_n} : \| \theta \|_{\infty} < p_n^{-1/2 + \zeta} \}$. We show that Assumption~\ref{assume:A-subset-timescale} holds for the coordinate-delocalized sets for all of the data models we consider, including, e.g., Gaussian projective models. See Sections~\ref{sec:pca-moment-deloc} and \ref{sec:deloc} below.

We are now ready to present out main result. Let $\uv_n(t)$ be the linear interpolation of $\uv_n(\theta_\ell)$ at times $t=\ell\delta_n$. 

\begin{theorem}\label{thm:main-ballistic}
    Suppose Assumptions~\ref{assume:localize}--\ref{assume:A-subset-timescale} hold for open sets $A_n$ and learning rates $\delta_n$, and $\uv_n(\theta_0)\Rightarrow\nu$. Fix $T > 0$. If $(\rho, \psi , B)$ is an adaptive, momentum-based SGD algorithm in the sense of Definition~\ref{def:rho}, the interpolated process $\uv_n$ converges weakly to $\uv$ in $C ([0, T ] ; \RR^k)$, where $\uv$ is the solution of,
    \begin{equation}\label{eq:main-adam-sde}
        \dd\uv_t=\Psiv(\uv_t)\,\dd t,\qquad \uv_0\sim\nu.
    \end{equation}
\end{theorem}

\begin{figure}[t]
    \centering
    \includegraphics[width=0.85\linewidth]{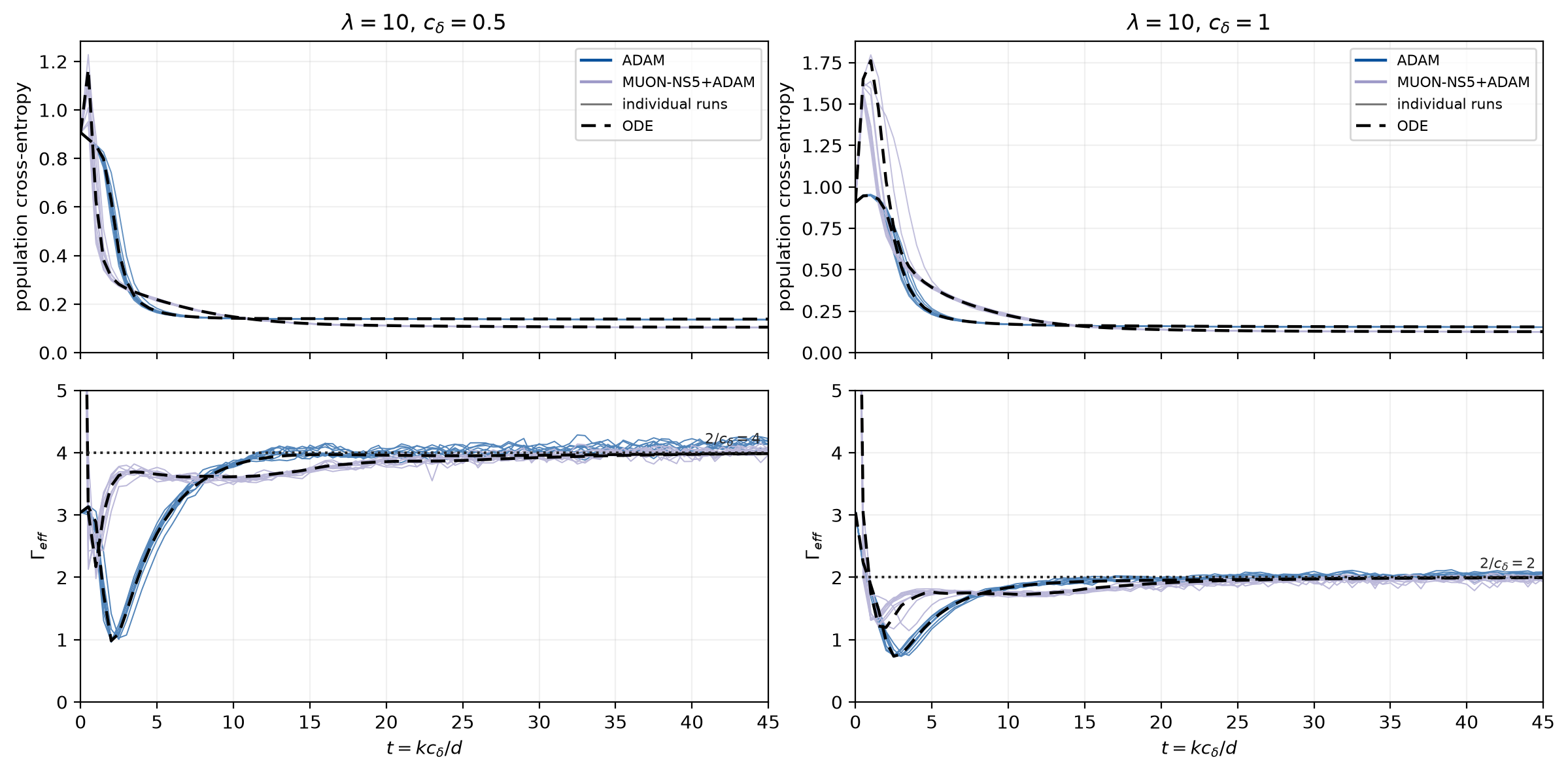}
    \caption{Demonstration of edge of stability for Gaussian XOR classification ($d=10000$) with softplus activation $\varphi (x) = \eta \log (1 + e^{x / \eta} )$, with $\eta = 0.05$, from random initialization, comparing ADAM ($\beta_1=0.9$, $\beta_2=0.95$), and mixed MUON-NS5 first layer $(\beta_M = 0.9$) and ADAM second-layer networks for $c_\delta \in \{0.5, 1\}$. The limiting ballistic ODEs (dashed curves) are overlaid on ten algorithmic trajectories (solid lines). On the $x$-axis we plot time $t = \ell \delta_n$ up to $T = 45$. On the $y$-axis we plot the population risk (top) and the effective infinity Laplacian $\Gamma^{\text{eff}}$ (bottom). Observe that $\Gamma^{\text{eff}}$ converges to $c_\delta / 2$ as the population risk plateaus for both algorithms.}
    \label{fig:xor-stability-plot}
\end{figure}

Let us compare this result with the classical theory of online SGD and the high-dimensional effective dynamics of \cite{BenArousGheissariJagannath2024Effective}. For online SGD, define
\[
    \cA_{\mathrm{SGD}}u =  \langle \EE_Y[ \nabla L (\theta , Y ) ] , \nabla u(\theta) \rangle,\qquad
    \cL_{\mathrm{SGD}}u=\frac12 \langle \EE_Y[ \nabla L (\theta , Y ) ^{\otimes 2} ] , \nabla^2u(\theta) \rangle.
\]
In the classical small-step limit at fixed dimension, the second-order term vanishes and the limiting dynamics converge to population gradient flow along the direction of the statistics. In high dimensions, the effective evolution instead follows,
\[
    \dd u\approx
    (-\cA_{\mathrm{SGD}}+\delta\cL_{\mathrm{SGD}})u\,\dd t\,.
\]
Thus there exists a \textit{critical scaling} for how fast $\delta$ tends to zero in which the second-order contribution can survive if the dimension grows at the same rate. This also implies a \textit{sub-critical} regime in which $\delta$ tends to zero fast enough where the second-order correction still vanishes and thus we recover the population dynamics above.

For momentum-based algorithms, the parameter process alone is generally not Markovian. The exponential moving average evolves on a faster time scale than the parameter. Averaging it against $\pi_\theta$ produces the limiting Markovian dynamics, with effective operator
\begin{equation}\label{eq:effective-operator-comparison}
    \dd u\approx
    [(-\tilde\cA_\theta+\delta\tilde\cL_\theta)u
    -\delta\tilde\cA_\theta\chi_{1,\theta}^u]\,\dd t\,.
\end{equation}
The first-order term $-\tilde\cA_\theta u$ is now the stationary mean search direction projected onto $\nabla u$. The high-dimensional correction now has two pieces: the stationary mean of the original high-dimensional correction observed for online SGD $\delta\tilde\cL_\theta u$, and a new correction, $-\delta\tilde\cA_\theta\chi_{1,\theta}^u$, from the momentum.

In the examples we consider, the momentum correction can be interpreted as an autocovariance. To see this, observe that $\chi_{1,\theta}^u$ solves $-\sL_\theta \chi_{1,\theta}^u = (I - \pi_\theta)\cA(\theta, \cdot)u(\theta)$, inverting $-\sL_\theta$ via the Neumann series shows that to leading order, the Poisson correction reduces to the autocovariance 
$$
    -\delta\tilde{\cA}_\theta \chi_{1,\theta}^u(\theta) \approx \frac{\delta}{2}\sum_{\ell=1}^\infty \EE_{\pi_\theta}\bigl[\langle \nabla_\theta^2 u(\theta) , \bar \rho(\tilde{X}_0) \otimes \bar \rho(\tilde{X}_\ell) \rangle + \langle \nabla_\theta^2 u(\theta) , \bar \rho(\tilde{X}_\ell) \otimes \bar \rho(\tilde{X}_0) \rangle \bigr],
$$
with $\bar \rho (X) = \rho (X) - \pi_\theta \rho (X)$. In principle, incorporating momentum into the algorithm imposes an additional high-dimensional correction that introduces a notion of ``temporal effects'' as the cost of averaging, on top of the classical correction arising from $\delta \cL$.

To reconcile with \cite{jagannath2026high}, we consider the special case of Polyak momentum (SGD-M) where $\rho(x) = (1-\beta)^{-1}x$ is linear. It was observed that both high-dimensional correction terms reduce to scalar multiples of the online SGD operator $\cL_{\mathrm{SGD}} u$. More explicitly, we have that,
$$
    \delta\tilde{\cL}_\theta u = \frac{1}{1-\beta^2}\delta\cL_{\mathrm{SGD}} u + o(1), \qquad \delta\tilde{\cA}_\theta \chi_{1,\theta}^u = \frac{2\beta}{(1-\beta)^2(1+\beta)}\delta\cL_{\mathrm{SGD}} u + o(1),
$$
from which summing the two contributions yields the total high-dimensional correction,
\[
    \delta\bigl(\tilde{\cL}_\theta u - \tilde{\cA}_\theta \chi_{1,\theta}^u\bigr) = \frac{1}{(1-\beta)^2}\delta\cL_{\mathrm{SGD}} u + o(1),
\]
recovering the $(1-\beta)^{-1}$ drift rescaling and $(1-\beta)^{-2}$ corrector rescaling established in Theorem 2.3 of \cite{jagannath2026high}. For nonlinear adaptive preconditioners such as ADAM and MUON, however, the autocovariances do not collapse to a multiple of the lag-zero variance.

\subsection*{Statement of AI Use} The manuscript and all proofs were entirely written by us. All proofs and proof ideas were developed on our own. ChatGPT 5.6 was used by one of us (AJ) to find a hole in an earlier proof of Lemma~\ref{lem:poisson-first-deriv-pde}. Claude Opus 5.5 was used to generate the code for the figures produced.

\subsection*{Acknowledgements}
The authors thank R. Gheissari and M. Li for numerous helpful comments on the manuscripts. A.J. and V.S.
acknowledge the support of the Natural Sciences and Engineering Research Council of Canada
(NSERC), the Canada Research Chairs program, the Canadian Foundation for Innovation- John
Evan’s Leaders fund, and the Ontario Research Fund. V.S. was partially supported by the Ontario Graduate Scholarship. Cette recherche a \'et\'e enterprise gr\^ace, en
partie, au soutien financier du Conseil de Recherches en Sciences Naturelles et en G\'enie du Canada
(CRSNG), du Programme des chaires de recherche du Canada, du Foundation canadienne pour
l’innovation FLJE, et les Fonds pour la recherce en Ontario. [RGPIN-2020-04597, DGECR-2020-
00199,CRC-2022-00142,CFI-JELF Project 43994].

\section{Applications: Phenomena in High-Dimensional Learning}
\label{sec:applications}

In this section, we demonstrate how Theorem~\ref{thm:main-ballistic} can sharply characterize phenomena in high-dimensional learning. We defer the full limiting ODE systems and their analysis to Sections~\ref{sec:ex-tensor-pca}--\ref{sec:gaus-xor-two-layer}, and touch on two applications of our theory: (1) precise comparisons between training algorithms and (2) sharp population-risk thresholds related to the edge of stability. We focus on the critical scaling regime in which the high-dimensional corrections contribute to the limiting dynamics.

To keep our discussion concrete, we first illustrate our results on the following learning problems and provide the effective dynamics in Sections~\ref{sec:ex-tensor-pca}--\ref{sec:gaus-xor-two-layer}. We emphasize, however, that these models are only representative instances of the broader family of high-dimensional, non-convex data models for which Theorem~\ref{thm:main-ballistic} and subsequent techniques apply.

\begin{example}[Spiked Tensor PCA] \label{ex:app-tensor-pca}
    At each step $\ell \ge 1$, we observe an i.i.d.\ symmetric order-$k$ tensor (with $k \ge 2$) of the form
    $$
        Y^{\ell} = \lambda \theta_*^{\otimes k} + W^{\ell},
    $$
    where $\theta_* \in \Sd^{d-1}$ is a hidden signal direction, $\lambda > 0$ is the signal-to-noise ratio (SNR), and $W^{\ell}$ is a symmetric Gaussian noise tensor normalized so that $\EE[\langle W, x^{\otimes k}\rangle \langle W, y^{\otimes k}\rangle] = \langle x, y\rangle^k$ for $x, y \in \RR^d$. Our goal is to infer $\theta_{*}$ from the samples. We do so by minimizing the squared loss $L(\theta, Y) = \Fnorm{Y - \theta^{\otimes k}}^2$ at the critical scaling $\delta_d = c_\delta / d$ for $c_\delta > 0$.

    The natural summary statistics to consider are the correlation and squared parameter norm, i.e. $\uv(\theta) = (m(\theta), R^2(\theta) )$ with $m(\theta) := \langle \theta, \theta_*\rangle$ and $R^2(\theta) := \norm{\theta}^2$, from which recovery of $\theta_{*}$ may be measured by the cosine alignment $a(\theta ) := m (\theta ) / R (\theta )$. The effective dynamics are provided in Section~\ref{sec:ex-tensor-pca}.
\end{example}

\begin{example}[Multi-class Logistic Regression]  \label{ex:app-mlr}
    Fix integers $q \ge 2$ and $k \ge 1$. At each step $\ell \ge 1$, we observe an i.i.d. pair $Y_\ell = (J_\ell, X_\ell)$ drawn from a $k$-component Gaussian mixture in $\RR^d$,
    $$
        X_\ell = \mu^{J_\ell} + \lambda^{-1/2} Z_\ell, \qquad J_\ell \sim \operatorname{Cat}(\mathsf{p}_1, \dots, \mathsf{p}_k), \qquad Z_\ell \sim \cN(0, I_d),
    $$
    with SNR $\lambda > 0$, component means $\mu^1, \dots, \mu^k \in \RR^d$ of norm $O(1)$, and one-hot class label $y(J_\ell) \in \{0, 1\}^q$. We train a linear classifier $\theta = (\theta^1, \dots, \theta^q) \in \RR^{d \times q}$ at critical step size $\delta_d = c_\delta / d$ by minimizing the cross-entropy loss,
    $$
        L(\theta, Y) = \Upsilon(\theta^\top X; y(J)) := \log \sum\nolimits_{a=1}^q e^{\langle \theta^a, X \rangle} - \sum\nolimits_{a=1}^q y_a(J) \langle \theta^a, X \rangle,
    $$
    optionally with decoupled weight decay (see Remark~\ref{rem:weight-decay}). The natural collection of statistics to consider are the family of correlations and parameter norms generated by the fixed-dimensional Gram matrix $G = (\theta , \mu )^\top (\theta , \mu ) \in \RR^{(q + k) \times (q + k)}$. We also track the population cross-entropy risk $\cR(\theta) := \EE_Y[L(\theta, Y)]$ whose dynamics are also characterized by $G$. The effective dynamics are provided in Section~\ref{sec:multi-class-log-reg-model}.
\end{example}

\begin{example}[Gaussian XOR Classification with Two-Layer Feedforward Networks] \label{ex:app-xor}
    Take two orthonormal signal directions $\mu, \nu \in \RR^d$, and consider a four-component Gaussian mixture $Y = (J, X)$ with $J \sim \operatorname{Unif}(\{1, 2, 3, 4\})$ and
    $$
        X = \mu^J + \lambda^{-1/2} Z, \qquad Z \sim \cN(0, I_d), \qquad \lambda > 0,
    $$
    where the cluster means and binary labels $y(J) \in \{0, 1\}$ are
    $$
        (\mu^1, y(1)) = (\mu, 1), \quad (\mu^2, y(2)) = (-\mu, 1), \quad (\mu^3, y(3)) = (\nu, 0), \quad (\mu^4, y(4)) = (-\nu, 0).
    $$
    Given an activation function $\varphi : \RR \to \RR$, we train a two-layer neural network $f(\theta^\top X, w) = \sum_{a=1}^q w_a \varphi(\langle \theta^a, X\rangle)$ of fixed hidden width $q \ge 4$ at step size $\delta_d=c_\delta/d$, with parameters $\Theta = (\theta, w) \in \RR^{d \times q} \times \RR^q$ under the cross-entropy loss,
    \[
        L(\Theta; Y) = \Upsilon(\theta^\top X, w; y(J)) := \log\bigl(1 + e^{f(\theta^\top X, w)}\bigr) - y(J) f(\theta^\top X, w).
    \]
    The effective dynamics are provided in Section~\ref{sec:gaus-xor-two-layer}.
\end{example}

\begin{figure}
    \centering
    \includegraphics[width=0.9\linewidth]{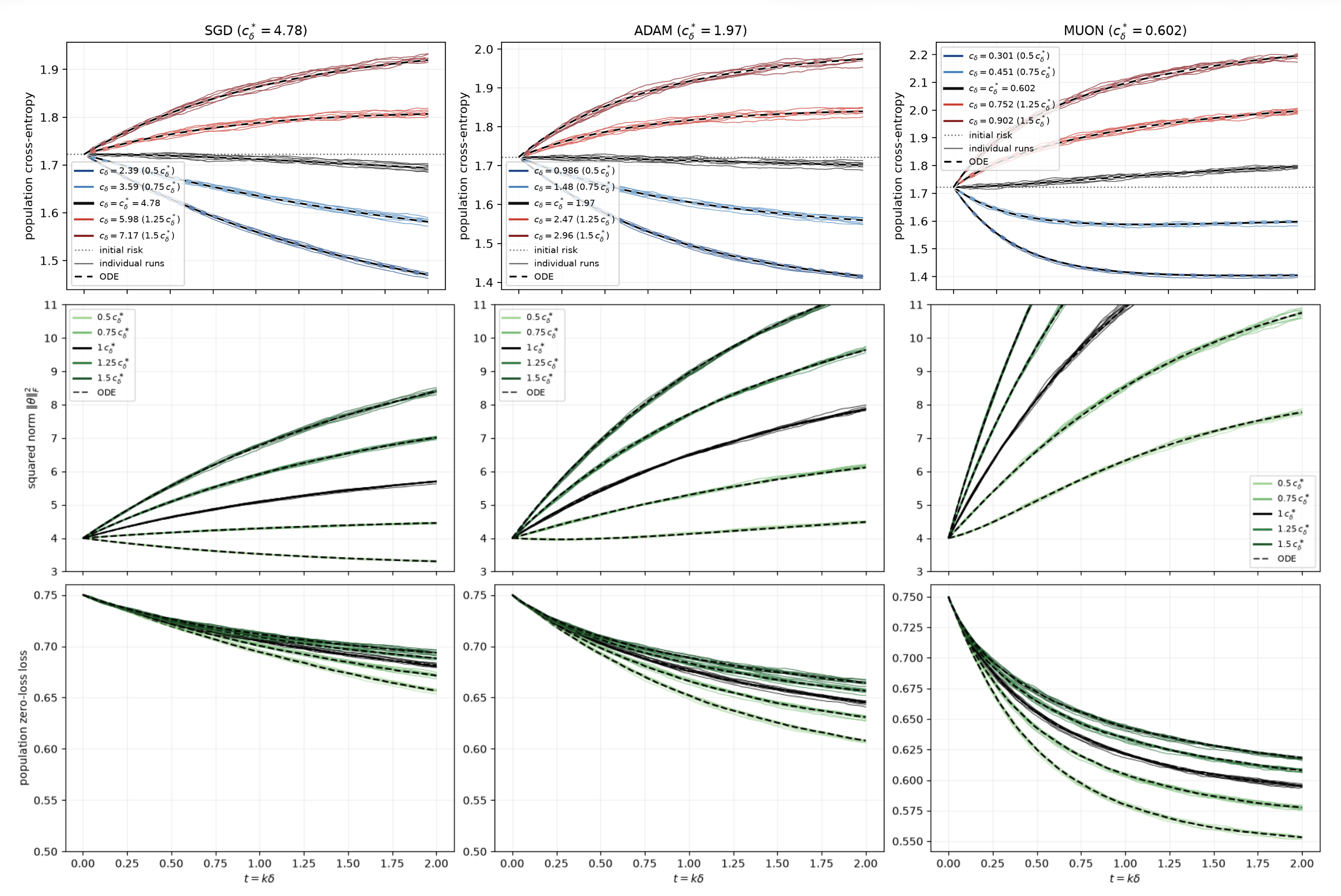}
    \caption{Phase transition at initialization for Multi-class Logistic Regression with $q = k = 4$ classes and $d = 10000$, $\lambda = 1$, and $\kappa = 0.2$, from $Q_0=I_q$, $M_0=0$. Columns correspond to online SGD, ADAM, and MUON, rows show population cross-entropy, total squared norm of the parameters, and population zero-one loss. Each algorithm uses
    $c_\delta/c_\delta^\ast\in\{0.5,0.75,1,1.25,1.5\}$, where $c_\delta^\ast=c_{\delta,\mathrm{alg}}^{\mathrm{desc}}$ is the predicted initial-risk threshold  ($c_\delta^\ast\approx4.78,1.97,0.602$, respectively). Solid curves show individual runs and dashed curves show the overlayed ODE predictions.}
    \label{fig:logreg-init-plot-risk}
\end{figure}

\subsection{Comparing training algorithms and the cross-over phenomenon}
\label{subsec:app-crossover}

A basic question is to understand the relative performance of different optimization algorithms and gain insights into when to choose one algorithm over another. The framework we develop provides a systematic tool to answer this question in high dimensions. As a consequence of our work, we find that comparing algorithms is delicate because their relative performance can reverse over the course of training, a behavior we call the \emph{cross-over phenomenon}.

Suppose that two algorithms are run on the same learning task with the same initialization and data. The cross-over phenomenon occurs when one algorithm performs strictly better at short training times, whereas the other achieves strictly better performance for all sufficiently large times. For continuous learning curves such as the ballistic limits, this reversal entails at least one finite crossing time $t_{\mathrm{crs}}>0$. Since each online step processes a fresh sample, training time is proportional to the number of samples processed when the algorithms use a common step size. Thus, in the regimes illustrated here, online SGD can outperform adaptive momentum algorithms such as ADAM if training is stopped early, whereas the adaptive methods can achieve better performance at sufficiently long training times.

Related reversals have been observed empirically \cite{keskar2017switching,wilson2017marginal}. In particular, \cite{keskar2017switching} describes settings in which adaptive methods initially perform better before SGD overtakes them. Furthermore, \cite{wu2025early} rigorously shows how early stopping in overparameterized logistic regression can substantially influence statistical performance. Our examples give explicit regimes in which the ordering is reversed: SGD performs better initially, and ADAM overtakes it later.

Figure~\ref{fig:pca-crossover-plot} illustrates this behavior by comparing the cosine alignment $a_t=m_t/R_t$ of online SGD, Polyak momentum (SGD-M), and ADAM on Spiked Matrix PCA ($k=2$, Example~\ref{ex:app-tensor-pca}) for a specific step size $c_\delta$ and signal-to-noise ratio $\lambda$. We observe that initially both SGD and SGD-M escape from the warm initialization for a period of time before ADAM crosses the SGD trajectories to achieve better performance at longer timescales. 

Using our framework, we can explain why this cross-over occurs and predict the parameter regimes in which it occurs. Recall from Section~\ref{sec:setup} that, in the critical scaling regime, the limiting trajectory combines the \emph{stationary averaged drift} $-\tilde\cA_\theta\uv$ and the \emph{high-dimensional correction} $\delta(\tilde\cL_\theta\uv-\tilde\cA_\theta\chi_{1,\theta}^{\uv})$. In Tensor PCA, the first term drives the parameter toward the signal, while the second increases its squared norm and can degrade alignment. Momentum and adaptive normalization transform these two contributions differently, creating a trade-off between initial speed and long-time recovery.

\begin{table}[t]
    \centering
    \begin{tabular}{lcc}
        \toprule
        Algorithm & Initial rate $\dot a_0/a_0$ at $R_0=1$ & Long-time alignment ($k=2$)\\
        \midrule
        Online SGD & $2k\Lambda_0-2kc_\delta$
        & $\displaystyle\sqrt{\left(1-\frac{c_\delta}{\lambda}\right)_+}$\\[2mm]
        SGD-M & $\displaystyle\frac{2k\Lambda_0}{1-\beta}-\frac{2kc_\delta}{(1-\beta)^2}$
        & $\displaystyle\sqrt{\left(1-\frac{c_\delta}{(1-\beta)\lambda}\right)_+}$\\[3mm]
        ADAM & $\displaystyle\sqrt{k}\,\kappa_1\Lambda_0-\frac{c_\delta\kappa_2}{2}$
        & $\displaystyle\sqrt{\left(1-\frac{c_\delta\zeta}{\lambda^{3/2}}\right)_+}$\\[2mm]
        \bottomrule
    \end{tabular}
    \caption{Initial alignment velocities for tensor order $k\ge2$, with $\Lambda_0=\lambda a_0^{k-2}(1-a_0^2)$, and long-time alignment in Matrix PCA from $a_0>0$. Here $(x)_+=\max\{x,0\}$, SGD-M uses unnormalized Polyak momentum, and the ADAM row takes $\epsilon\downarrow0$ after the high-dimensional limit. The positive constants $\kappa_1,\kappa_2$ depend only on $(\beta_1,\beta_2)$, and $\zeta=\kappa_2/(2\sqrt2\kappa_1)$.}
    \label{tab:pca-algo-summary}
\end{table}

To see this mechanism in detail, we compare the ballistic limits on Spiked Tensor PCA for online SGD \cite[Propositions~3.1--3.3]{BenArousGheissariJagannath2024Effective}, SGD-M \cite[Propositions~3.1--3.2]{jagannath2026high}, and ADAM. Proposition~\ref{prop:pca-sgdm-sgdu} gives their common form, and Corollary~\ref{cor:adam-pca-eps-zero} justifies the ADAM limit $\epsilon\downarrow0$ with $\beta_1\in[0,1)$ and $\beta_2\in(0,1)$ fixed. Table~\ref{tab:pca-algo-summary} records two quantities extracted from these ODEs: the initial relative alignment rate and the limiting alignment for $k=2$. The first decomposes into a positive signal term proportional to the initial signal strength $\Lambda_0=\lambda a_0^{k-2}(1-a_0^2)$ and a noise penalty proportional to $c_\delta$. The second follows from the global convergence result in Proposition~\ref{prop:adam-pca-global}.

\subsubsection*{Initial speed and the effect of the step size.} In the weak-alignment limit $a_0\downarrow0$ for Matrix PCA, the initial rates become affine functions of $c_\delta$:
\begin{equation}\label{eq:pca-k2-init-lines}
    \begin{aligned}
        \frac{\dot a_{\mathrm{SGD}}(0)}{a_0}&\rightarrow4(\lambda-c_\delta),\qquad
        \frac{\dot a_{\mathrm M}(0)}{a_0}\to
        \frac4{1-\beta}\left(\lambda-\frac{c_\delta}{1-\beta}\right),\qquad
        \frac{\dot a_{\mathrm A}(0)}{a_0}\to
        \sqrt2\kappa_1\lambda-\frac{\kappa_2}{2}c_\delta.
    \end{aligned}
\end{equation}
Their zeros give the weak-alignment thresholds for positive initial growth:
\begin{equation}\label{eq:pca-k2-escape-thresholds}
    c_{\delta,\mathrm M}^{\mathrm{esc}}=(1-\beta)\lambda,
    \qquad c_{\delta,\mathrm{SGD}}^{\mathrm{esc}}=\lambda,
    \qquad c_{\delta,\mathrm A}^{\mathrm{esc}}
    =2\sqrt 2(\kappa_1/\kappa_2)\lambda.
\end{equation}

When $\kappa_2<8$, $\kappa_1<2\sqrt2$, and $\kappa_2<2\sqrt2\kappa_1$, ADAM has the smallest sensitivity to the noise penalty as $c_\delta$ grows, and the largest initial-growth threshold. On the other hand, as $c_\delta\downarrow0$, ADAM has the smallest signal drift. Thus the initial velocity of online SGD, respectively SGD-M, overtakes that of ADAM when $c_\delta$ drops below
\begin{equation}\label{eq:pca-k2-crossovers}
    \begin{aligned}
        c_{\delta,\mathrm{A,SGD}}^*
        &:=\frac{8-2\sqrt2\kappa_1}{8-\kappa_2}\lambda,\qquad c_{\delta,\mathrm{A,M}}^*
        :=\frac{8/(1-\beta)-2\sqrt2\kappa_1}
        {8/(1-\beta)^2-\kappa_2}\lambda.
    \end{aligned}
\end{equation}

\subsubsection*{Why ADAM overtakes online SGD at long training times.} Although online SGD initially gains alignment faster when $c_\delta<c_{\delta,\mathrm{A,SGD}}^*$, the last column of Table~\ref{tab:pca-algo-summary} describes what happens later in training. Whenever $\lambda>\zeta^2$ and $c_\delta<\lambda$, ADAM converges to a solution with strictly greater alignment than SGD. Combined with the initial ordering, this forces the learning curves to cross. We summarize both regimes, together with the corresponding population-risk comparison, in the following proposition.

\begin{proposition}\label{prop:app-pca-crossovers}
    Fix $k=2$, $\lambda>0$, $\beta\in(0,1)$, and $(\beta_1,\beta_2)\in[0,1)\times(0,1)$ with $\kappa_1<2\sqrt2$ and $\kappa_2<8$. Set
    \[
        \zeta=\frac{\kappa_2}{2\sqrt2\kappa_1},\qquad
        c_*:=\frac{8-2\sqrt2\kappa_1}{8-\kappa_2},\qquad
        c_{\delta,\mathrm A}^{\mathrm{eq}}:=\frac{\lambda^{3/2}}{\zeta}.
    \]
    Let $(R_t^{\mathrm{alg}},a_t^{\mathrm{alg}})$ denote the limiting trajectories of online SGD, SGD-M, and ADAM, initialized at $R_0=1$ and $a_0\in(0,1)$.
    \begin{enumerate}[label=\textup{(\roman*)},leftmargin=*]
        \item If $\lambda>\zeta^2$ and $\lambda\le c_\delta<c_{\delta,\mathrm A}^{\mathrm{eq}}$, then ADAM converges to the learned equilibrium with positive alignment, whereas online SGD and SGD-M lose alignment:
        \[
            a_t^{\mathrm A}\rightarrow
            \sqrt{1-\frac{\zeta c_\delta}{\lambda^{3/2}}}>0,
            \qquad a_t^{\mathrm{SGD}},a_t^{\mathrm M}\rightarrow0.
        \]
        \item If $\lambda>\zeta^2$ and $0<c_\delta<\min\{c_*\lambda,\lambda\}$, then for all sufficiently small $a_0>0$, online SGD strictly dominates ADAM on some interval $(0,t_0]$. There is a finite crossing time $t_{\mathrm{crs}}>t_0$, after which $a_t^{\mathrm A}>a_t^{\mathrm{SGD}}$, and
        \begin{equation}\label{eq:pca-trajectory-crossover}
            \lim_{t\to\infty}a_t^{\mathrm A}
            =\sqrt{1-\frac{\zeta c_\delta}{\lambda^{3/2}}}
            >\sqrt{1-\frac{c_\delta}{\lambda}}
            =\lim_{t\to\infty}a_t^{\mathrm{SGD}}.
        \end{equation}
        \item For the population risk $\cR_{\mathrm{alg}}(t)=(R_t^{\mathrm{alg}})^4-2\lambda(m_t^{\mathrm{alg}})^2$, if $\lambda>\zeta^2$ and $0<c_\delta<\lambda$, then
        \begin{equation}\label{eq:app-pca-risk-crossover-limits}
            \begin{aligned}
                \lim_{t\to\infty}\cR_{\mathrm A}(t)
                &=-\lambda^2+2\zeta c_\delta\sqrt\lambda <-\lambda^2+2\lambda c_\delta
                =\lim_{t\to\infty}\cR_{\mathrm{SGD}}(t).
            \end{aligned}
        \end{equation}
        If also $c_\delta<c_*$, then for all sufficiently small $a_0>0$, SGD initially has strictly smaller risk, so the population risks also cross at a finite time and ADAM has smaller risk after their last intersection. Both risks initially decrease under the additional condition $c_\delta<\min\{1,\zeta^{-1}\}$.
    \end{enumerate}
\end{proposition}

\subsubsection{Cross-over phenomena for population risks}
\label{subsubsec:app-risk-crossover}

The same framework compares population risks in convex classification (Multi-class Logistic Regression, Example~\ref{ex:app-mlr}) and nonconvex multilayer feature learning (Gaussian XOR, Example~\ref{ex:app-xor}). In both models, the population cross-entropy $\cR(\theta):=\EE_Y[L(\theta,Y)]$ is determined by the summary statistics; see \eqref{eq:gpm-pop-risk}. Differentiating along their limiting trajectories by Lemma~\ref{lem:gaussian-risk-chain-rule} gives
\begin{equation}\label{eq:app-universal-risk-ode}
    \frac{\dd}{\dd t}\cR\bigl(\uv_t^{\mathrm{alg}}\bigr)
    =-\cF_{\mathrm{alg}}\bigl(\uv_t^{\mathrm{alg}}\bigr)
    +c_\delta\cD_{\mathrm{alg}}\bigl(\uv_t^{\mathrm{alg}}\bigr).
\end{equation}
Here $\cF_{\mathrm{alg}}$ is the limiting population drift, while $c_\delta\cD_{\mathrm{alg}}$ is the combined high-dimensional correction, including the temporal correlations caused by momentum. The formulas are given in \eqref{eq:gpm-risk-F+D}.

At a common initialization $\uv_0$, the initial velocities are affine functions of $c_\delta$. Whenever $\cF_{\mathrm{alg}}(\uv_0)$, $\cD_{\mathrm{alg}}(\uv_0)>0$, the algorithm initially decreases risk precisely when $c_\delta$ is below
\begin{equation}\label{eq:app-general-closs}
    c_{\delta,\mathrm{alg}}^{\mathrm{desc}}
    :=\frac{\cF_{\mathrm{alg}}(\uv_0)}{\cD_{\mathrm{alg}}(\uv_0)}.
\end{equation}
For example, Lemma~\ref{lem:mlr-init-master} gives these coefficients for logistic regression. If two algorithms satisfy $\cF_1(\uv_0)>\cF_2(\uv_0)>0$ and $\cD_1(\uv_0)>\cD_2(\uv_0)>0$, equality of their initial velocities occurs at
\begin{equation}\label{eq:app-general-cdelta-init-star}
    c_{\delta,1,2}^*
    :=\frac{\cF_1(\uv_0)-\cF_2(\uv_0)}
    {\cD_1(\uv_0)-\cD_2(\uv_0)}.
\end{equation}
Consequently, for $0<c_\delta<\min\{c_{\delta,1,2}^*,c_{\delta,2}^{\mathrm{desc}}\}$, we have $\dot\cR_1(0)<\dot\cR_2(0)<0$, so algorithm $1$ achieves strictly smaller risk on an interval immediately after initialization.

To turn this local comparison into a cross-over, one also needs a long-time comparison. A sufficient condition, when the two finite limiting risks exist, is
\begin{equation}\label{eq:app-general-crossover-criterion}
    \lim_{t\to\infty}\cR\bigl(\uv_t^{2}\bigr)
    <\lim_{t\to\infty}\cR\bigl(\uv_t^{1}\bigr).
\end{equation}
Continuity then gives a finite crossing, after which Algorithm $2$ has lower risk. Refer to the right panel of Figures~\ref{fig:pca-crossover-plot} and ~\ref{fig:xor-crossover-plot} for simulations demonstrating cross-overs and predicted ODEs for the Spiked Tensor PCA and Gaussian XOR classification model(s) in the population risk.

\subsection{The edge of stability and critical step-size scaling}
\label{subsec:app-eos}

A key observation in neural network optimization is the \emph{Edge of Stability} (EoS) phenomenon, which was identified for full-batch algorithms such as gradient descent \cite{cohen2021gradient,jastrzebski2019sharpest,jastrzebski2020break} and subsequently studied for adaptive methods \cite{cohen2024adaptive}. For gradient descent, the \emph{sharpness}, defined as the largest eigenvalue of the training-loss Hessian, typically grows toward and hovers near a threshold $2/\delta$ where $\delta$ is a fixed learning rate. The loss can then fluctuate over short time scales while continuing to decrease over longer ones \cite{cohen2021gradient}. For adaptive methods, the corresponding criteria follows from a preconditioned Hessian, and the relevant threshold instead depends on properties of the preconditioner and momentum convention \cite{cohen2024adaptive}. Several complementary theoretical explanations of these dynamics have been developed \cite{agarwala2023second,arora2022understanding,chen2023beyond,damian2023self,lyu2022normalization,zhu2023minimalist}.

Modern large-scale training, however, relies on stochastic gradient methods, motivating a parallel body of work \cite{agarwala2025conservative,andreyev2026momentum,andreyev2025edge,lee2023new,mulayoff2024exact,wu2018how}. Here, along with related sharpness analyses, it was shown that the sample-level gradient fluctuations and how they interact with the Hessian can change the stability criteria and associated comparator. Thus, when studying the risk, the relevant comparator is not the largest Hessian eigenvalue, but instead an analogue of the (normalized) infinity Laplacian. Its importance for SGD was first highlighted in \cite{lee2023new} and distinguished from an instability criterion in \cite{andreyev2025edge}. For online SGD, its population-risk version is given by,
$$
    \mathrm{\Gamma}^N_d(\theta)
    :=
    \frac{\EE_Y[
    \langle \nabla_\theta L(\theta,Y),
    \nabla_\theta^2\cR_d(\theta)\nabla_\theta L(\theta,Y)
    \rangle]}
    {\|\nabla_\theta\cR_d(\theta)\|^2},
$$
whenever $\|\nabla_\theta\cR_d(\theta)\|>0$, which also admits a threshold near $2 / \delta$, for which the risk stagnates.

\begin{figure}
    \centering
    \includegraphics[width=0.9\linewidth]{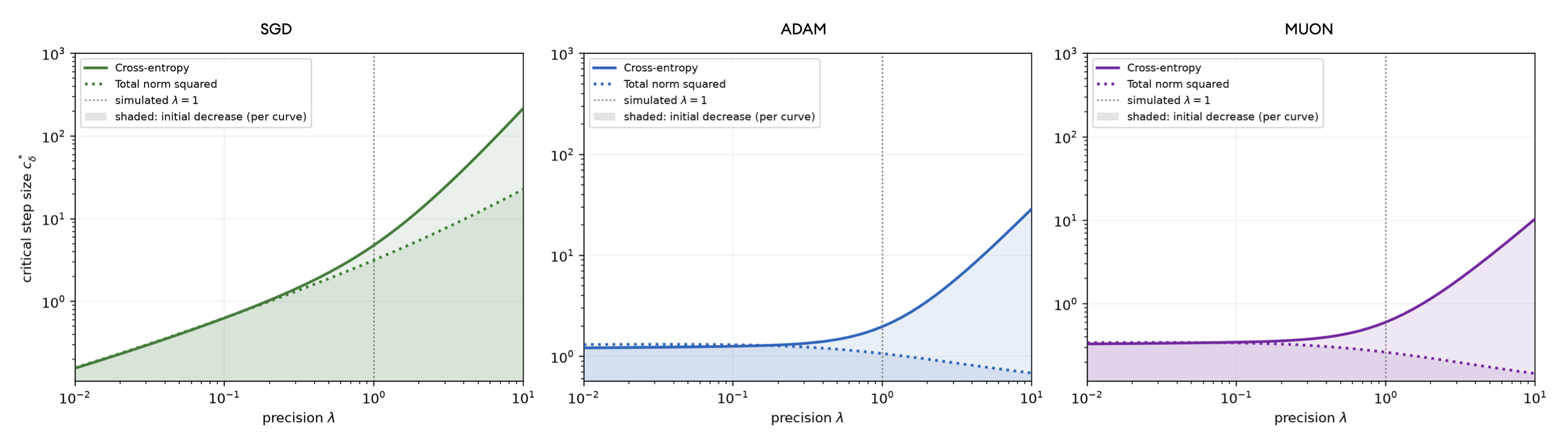}
    \caption{Phase diagrams of the initial risk for Multi-class Logistic Regression in the setting of Figure~\ref{fig:logreg-init-plot-risk}. We vary $\lambda$ and view the critical step size $c_* (\lambda )$ as a function of $\lambda$.
    Solid curves mark the step-size at which $\dot{\cR}(0)=0$ (i.e. the critical value), while coloured dotted curves mark those at which $ \partial_t \|\theta_t\|_F^2 |_{t=0}=0$ is the total squared parameter norm.
    Below each boundary, the corresponding statistic initially decreases, 
    above it, it initially increases. Shading indicates the decreasing regions.
    The separation of the boundaries at $\lambda=1$ exhibits regimes
    where the norm initially grows while the population risk decreases.}
    \label{fig:logreg-phase-boundary}
\end{figure}

The difference between the full-batch and stochastic regimes becomes especially stark in the high-dimensional critical scaling regime. For the models we consider, the population-risk Hessian has bounded operator norm, so that $\delta_d\|\nabla_\theta^2\cR_d(\theta)\|_{\mathrm{op}}$ tends to zero as dimension increases. Thus, from a full-batch perspective, high-dimensional online algorithms operate far below the classical Hessian barrier of two, which would naively suggest that the algorithm always forces risk descent for any $c_\delta > 0$. Instead, precisely in line with the above perspective, the high-dimensional fluctuations can still contribute to the risk criteria. Our framework sharply characterizes this high-dimensional risk threshold for a broad family of adaptive, momentum-based algorithms, including ADAM and MUON, whenever the population-risk closes with respect to the statistics. In particular, we demonstrate how our ballistic limits allow us to establish a high-dimensional analogue of the $2/\delta$ threshold that arises from the Edge of Stability phenomena.

For online SGD, the operators from \eqref{eq:operators} satisfy
\begin{equation}\label{eq:app-sgd-operators-explicit}
        \cA_{\mathrm{SGD}}\cR_d(\theta)
        =\norm{\nabla_\theta\cR_d(\theta)}^2,\qquad \delta_d\cL_{\mathrm{SGD}}\cR_d(\theta)
        =\frac{\delta_d}{2}\EE_Y\bigl[ \langle \nabla_\theta L (\theta , Y) ,
        \nabla_\theta^2\cR_d(\theta) \nabla_\theta L (\theta , Y) \rangle
        \bigr].
\end{equation}
Consequently, the infinity Laplacian is twice the ratio of the second-order and first-order operators:
$$
    \bigl(-\cA_{\mathrm{SGD}}+\delta_d\cL_{\mathrm{SGD}}\bigr)\cR_d
    =-\norm{\nabla_\theta\cR_d}^2
    (1-\frac{\delta_d}{2} \mathrm{\Gamma}^N_d(\theta) ).
$$
At $\delta_d=c_\delta/d$, the relevant normalized statistic is $\mathrm{\Gamma}^N_d(\theta) /d$.

\begin{proposition}\label{prop:app-eff-inf-lap-window}
    Consider online SGD on Examples~\ref{ex:app-tensor-pca}--\ref{ex:app-xor}, or more broadly for Gaussian projective models (see Section~\ref{sec:ex-gpm}). If $\uv_d(\theta_d)\to\uv$ and $\cF_{\mathrm{SGD}}(\uv)>0$, then
    $$
        \frac1d\mathrm{\Gamma}^N_d(\theta_d)
        \rightarrow\Gamma_{\mathrm{SGD}}^{\mathrm{eff}}(\uv)
        :=\frac{2\cD_{\mathrm{SGD}}(\uv)}{\cF_{\mathrm{SGD}}(\uv)}.
    $$
\end{proposition}

Along the limiting SGD trajectory, the resulting risk equation is
\begin{equation}\label{eq:app-sgd-risk-ode-gamma}
    \frac{\dd}{\dd t}\cR(\uv_t^{\mathrm{SGD}})
    =-\cF_{\mathrm{SGD}}(\uv_t^{\mathrm{SGD}})
    \left[1-\frac{c_\delta}{2}
    \Gamma_{\mathrm{SGD}}^{\mathrm{eff}}(\uv_t^{\mathrm{SGD}})
    \right].
\end{equation}
Thus the rescaled boundary $2/c_\delta$ has a direct interpretation in terms of the original population-gradient and Hessian quantities.

For algorithms carrying momentum, the operators are averaged with respect to the fixed-parameter chain, and we also have the momentum correction that arises from the Poisson equation arguments (see Section~\ref{thm:main-ballistic}). In the models considered here, the risk-operator limits give
\begin{equation}\label{eq:app-homogenized-risk-approx}
    -\tilde\cA_\theta\cR_d
    +\delta_d\bigl(\tilde\cL_\theta\cR_d
    -\tilde\cA_\theta\chi_{1,\theta}^{\cR_d}\bigr)
    \rightarrow-\cF_{\mathrm{alg}}(\uv)+c_\delta\cD_{\mathrm{alg}}(\uv).
\end{equation}
To interpret the new momentum correction, let $(\tilde X_h)_{h\in\ZZ}$ be the stationary fixed-parameter momentum chain, and write $\rho_h=\rho(\tilde X_h)$ and $\bar \rho_h=\rho_h-\pi_\theta\rho$. For our examples, this new correction, combined with Lemma~\ref{lem:risk-operator-split} justifying the separate limit, give the autocovariance quantity,
\begin{equation}\label{eq:app-poisson-autocov-approx}
    \begin{aligned}
        -\delta_d\tilde\cA_\theta\chi_{1,\theta}^{\cR_d}
        =\frac{\delta_d}{2}\sum\nolimits_{h\ge1}\EE_{\pi_\theta}\bigl[
        \langle \nabla_\theta^2\cR_d(\theta), \bar \rho_0 \otimes \bar \rho_h \rangle
        + \langle \nabla_\theta^2\cR_d(\theta), \bar \rho_h \otimes \bar \rho_0 \rangle \bigr]+o(1).
    \end{aligned}
\end{equation}
The centered temporal covariances distinguish this term from the original high-dimensional correction. More precisely, the three operators converge separately:
\begin{equation}\label{eq:app-three-operator-limits}
    \begin{aligned}
        \tilde\cA_\theta\cR_d&\to\cF_{\mathrm{alg}}(\uv),\qquad
        d^{-1}\tilde\cL_\theta\cR_d
        \to\cD_{\mathrm{alg}}^{(0)}(\uv),\qquad
        -d^{-1}\tilde\cA_\theta\chi_{1,\theta}^{\cR_d}
        \to\cD_{\mathrm{alg}}^{(\mathrm{mem})}(\uv),
    \end{aligned}
\end{equation}
with $\cD_{\mathrm{alg}}=\cD_{\mathrm{alg}}^{(0)}+\cD_{\mathrm{alg}}^{(\mathrm{mem})}$.

Whenever $\cF_{\mathrm{alg}}(\uv)>0$, define the \emph{effective infinity Laplacian} by
\begin{equation}\label{eq:app-gamma-eff}
    \Gamma_{\mathrm{alg}}^{\mathrm{eff}}(\uv)
    :=\frac{2\cD_{\mathrm{alg}}(\uv)}{\cF_{\mathrm{alg}}(\uv)}
    =\frac{2\cD_{\mathrm{alg}}^{(0)}(\uv)}{\cF_{\mathrm{alg}}(\uv)}
    +\frac{2\cD_{\mathrm{alg}}^{(\mathrm{mem})}(\uv)}{\cF_{\mathrm{alg}}(\uv)}.
\end{equation}
Factoring \eqref{eq:app-universal-risk-ode} gives
\begin{equation}\label{eq:app-risk-ode-factored}
    \frac{\dd}{\dd t}\cR(\uv_t^{\mathrm{alg}})
    =-\cF_{\mathrm{alg}}(\uv_t^{\mathrm{alg}})
    \left[1-\frac{c_\delta}{2}
    \Gamma_{\mathrm{alg}}^{\mathrm{eff}}(\uv_t^{\mathrm{alg}})
    \right].
\end{equation}
Below the boundary $\Gamma_{\mathrm{alg}}^{\mathrm{eff}}=2/c_\delta$, the averaged descent contribution dominates and the risk decreases. At the boundary, the instantaneous risk derivative vanishes, and above it, the risk increases. When $\cD_{\mathrm{alg}}(\uv)>0$ as well, the equivalent state-dependent step-size threshold is
\begin{equation}\label{eq:app-state-cdesc}
    c_{\delta,\mathrm{alg}}^{\mathrm{desc}}(\uv)
    :=\frac{\cF_{\mathrm{alg}}(\uv)}{\cD_{\mathrm{alg}}(\uv)}
    =\frac2{\Gamma_{\mathrm{alg}}^{\mathrm{eff}}(\uv)}.
\end{equation}
The same identity explains long-time convergence of the algorithm. Rearranging \eqref{eq:app-risk-ode-factored} gives
\begin{equation}\label{eq:app-dni-balance-identity}
    \Gamma_{\mathrm{alg}}^{\mathrm{eff}}(\uv_t^{\mathrm{alg}})
    =\frac2{c_\delta}\left[
    1+\frac{\frac{\dd}{\dd t}\cR(\uv_t^{\mathrm{alg}})}
    {\cF_{\mathrm{alg}}(\uv_t^{\mathrm{alg}})}
    \right].
\end{equation}
If the trajectory approaches an equilibrium, and the limiting descent coefficient $\cF_{\text{alg}}$ is strictly positive, then as the risk derivative tends to zero we have that $\Gamma_{\mathrm{alg}}^{\mathrm{eff}}\to2/c_\delta$. We demonstrate this concretely in the Matrix PCA example (also see Figure~\ref{fig:pca-stability-plot}).

\begin{table}[t]
    \centering
    \begin{tabular}{lc}
        \toprule
        Algorithm & $c_{\delta, \text{alg}}^{\text{desc}}$ at zero init: $Q_0=M_0=0$\\
        \midrule
        Online SGD & $2\lambda^2$\\[2mm]
        Unnormalized SGD-M & $2(1-\beta)\lambda^2$\\[2mm]
        ADAM ($q=2$, $\epsilon_A\downarrow0$)
        & $\displaystyle\frac{\kappa_1(\beta_1,\beta_2)}{\kappa_2(\beta_1,\beta_2)}\lambda^{3/2}$\\[3mm]
        Exact MUON ($q=2$, $\epsilon_M\downarrow0$)
        & $\displaystyle\sqrt{\frac{2(1-\beta_M)}{1+\beta_M}}\lambda^{3/2}$\\[2mm]
        \bottomrule
    \end{tabular}
    \caption{Exact initial-descent thresholds $c_{\delta,\mathrm{alg}}^{\mathrm{desc}}$ for logistic regression with $k=q$, $y(j)=e_j$, orthonormal means, uniform mixture weights, and no weight decay. The SGD and SGD-M rows hold for every $q\ge2$; the ADAM and MUON rows are binary. Momentum parameters are fixed, with $\beta_2\in(0,1)$ for ADAM. Regularizers tend to zero after the high-dimensional limit at zero initialization. The constants $\kappa_1,\kappa_2$ are those of Corollary~\ref{cor:adam-pca-eps-zero}.}
    \label{tab:eos-desc-thresholds}
\end{table}

\begin{proposition} \label{prop:app-pca-eoss-contraction}
    Fix $k=2$, $\lambda,c_\delta>0$, and consider online SGD or ADAM with $\beta_1\in[0,1)$ and $\beta_2\in(0,1)$ when $\epsilon_A \downarrow 0$. For initialization $(m_0, R_0 )$, write $\cB(m,R):=R^4(R^2-m^2)+m^2(R^2-\lambda)^2$.

    For every $t>0$, $\cB(m_t,R_t)>0$, and the effective ratios along the respective trajectories are
    \begin{equation}\label{eq:app-pca-dni-explicit}
        \Gamma_{\mathrm{SGD}}^{\mathrm{eff}}(m_t,R_t)
        =\frac{2R_t^4}{\cB(m_t,R_t)},\qquad
        \Gamma_{\mathrm A}^{\mathrm{eff}}(m_t,R_t)
        =\frac{\kappa_2R_t^3}{\sqrt2\kappa_1\cB(m_t,R_t)}.
    \end{equation}
    Therefore for any regime with $c_\delta > 0$, we have that,
    \begin{equation}\label{eq:app-pca-dni-limit}
        \lim_{t\to\infty}\Gamma_{\mathrm{SGD}}^{\mathrm{eff}}(m_t,R_t)
        =\lim_{t\to\infty}\Gamma_{\mathrm A}^{\mathrm{eff}}(m_t,R_t)
        =\frac2{c_\delta}.
    \end{equation}
\end{proposition}

Thus the same asymptotic risk condition can occur whether the limiting estimator retains a nonzero signal alignment or loses alignment. In Figure~\ref{fig:xor-stability-plot} we illustrate the same phenomena emerging for the Gaussian XOR problem for two-layer networks when the algorithms reach convergence.

We conclude with a separate application to illustrate how a risk threshold at initialization can result in a phase transition. In particular, following from the preceding discussion we determine a phase transition at $t = 0$ to determine what values of the step size $c_\delta$ force the risk to initially increase, decrease, or stagnate. To do so, we consider the Multi-class Logistic Regression with uniform mixture weights $\mathsf p_j=1/k$ and orthonormal means. Write $Q=\theta^\top\theta$ and $M=(\theta^\top\mu^1,\ldots,\theta^\top\mu^k)$, and initialize at $Q_0=\tau^2I_q$, $M_0=0$.

\begin{proposition}\label{prop:app-mlr-eos}
    Consider Multi-class Logistic Regression with $\mathsf p_j=1/k$, orthonormal means, and initialization $Q_0=\tau^2I_q$, $M_0=0$.
    \begin{enumerate}[label=\textup{(\roman*)},leftmargin=*]
        \item For online SGD with $\ell^2$ weight decay parameter $\kappa \ge 0$, the exact threshold at $\tau=1$ is
        \begin{equation}\label{eq:app-mlr-sgd-tau1}
            c_{\delta,\mathrm{SGD}}^{\mathrm{desc}}(1,\lambda)
            =\frac{2\bigl(\lambda^2/(qk)+h(\lambda)^2
            +\kappa\lambda h(\lambda)\bigr)}
            {h(\lambda)(2/q-h(\lambda))},
        \end{equation}
        where $h(\lambda)=h(1,\lambda)\in(0,1/q]$ is defined in \eqref{eq:mlr-h-lambda-def}.

        \item If $k=q$ and $y(j)=e_j$, the zero-initialization thresholds are given by Table~\ref{tab:eos-desc-thresholds}.
    \end{enumerate}
    In each case, $\dot\cR_{\mathrm{alg}}(0)$ is negative, zero, or positive according as $c_\delta$ is smaller than, equal to, or larger than $c_{\delta,\mathrm{alg}}^{\mathrm{desc}}$.
\end{proposition}

The closed-form thresholds in Table~\ref{tab:eos-desc-thresholds} distinguish the SNR dependence of the algorithms at zero initialization. SGD has threshold $2\lambda^2$, whereas the binary ADAM and Exact MUON thresholds scale as $\lambda^{3/2}$. Thus, for fixed momentum parameters, these adaptive methods tolerate larger step-size constants at low SNR, while SGD tolerates larger constants at high SNR. Figure~\ref{fig:logreg-init-plot-risk} illustrates the initial-risk transition from $Q_0=I_q$, $M_0=0$ at $\lambda=1$. The threshold $c_{\delta,\mathrm{alg}}^{\mathrm{desc}}$ separates initial risk descent from ascent. The additional statistics distinguish this transition from other aspects of learning. In particular, the norm can increase while the risk decreases, and the classification error decreases even for the step sizes above the risk threshold. Figure~\ref{fig:logreg-phase-boundary} provides the corresponding phase diagram to compare the initial risk-ascent and norm-growth boundaries.

\section{Proof of Dynamics}
\label{sec:proof-main}

In this section, we prove Theorem~\ref{thm:main-ballistic}. Specifically, we establish that $\uv_{n}\Rightarrow\uv$ weakly as random variables on $C([0,T];\RR^{k})$ equipped with the sup-norm metric, where $\uv$ solves \eqref{eq:main-adam-sde}. To this end, fix $R, T > 0$, let $E_R := \{x \in \RR^{k} : \norm{x} < R\}$ be the open Euclidean ball of radius $R$, and recall that $\uv_n(t)$ is the linear interpolation of $(\uv_n(\theta_\ell))_{\ell\ge0}$ at the times $t=\ell\delta_n$. Define the stopping time $\tau_R := \inf\{t\ge0:\norm{\uv_n(t)} \ge R\}$ and corresponding stopped process $\uv_n^R(t) := \uv_n(t\wedge\tau_R)$.

Let $E_{R}^{*}:=\uv_{n}^{-1}(E_{R})$ denote the open pre-image of $E_{R}$ in parameter space. For a function $f$ we use the shorthand $F_{\ell}$ for $f(Y_{\ell})$. For a function $f= f(x,\theta)$, we denote $f_\ell=f(\theta_\ell,X_\ell)$. We denote the martingale part of $f$ as $\cM^f$ whose martingale difference is $\Delta \cM^f_\ell=f(\theta_\ell,X_\ell)-\E[f(\theta_{\ell-1},X_\ell) | \cF_{\ell-1}]$. We denote the increment by $\Delta f_\ell = f(\theta_\ell,X_\ell)-f(\theta_{\ell-1},X_{\ell-1})$. It is also helpful to define the parameter increment by $\Delta_\theta f_\ell = f(\theta_\ell,X_\ell)-f(\theta_{\ell-1},X_\ell)$. Under the assumption that $\PP(\tau_A \le ( \tau_R / \delta ) \land \lfloor T / \delta \rfloor) \to 0$, Assumption~\ref{assume:closable} holds on $E_R^* \cap A_n$ for all $\ell < (\tau_R / \delta ) \land \lfloor T / \delta \rfloor$.

\subsection{Step 1: Method of Averaging}\label{subsec:step1}

We begin by applying the Method of Averaging, to replace the full process with an averaged proxy. Start with the Taylor expansion of a single statistic $u$ from $\uv$. Since $\theta_\ell=\theta_{\ell-1}-\delta D_\ell$ with $D_\ell=\rho(X_\ell)$,
\begin{align}
    u (\theta_{\ell} ) &= u \left( \theta_{\ell - 1} \right) - \delta \left\langle D_\ell, \nabla u (\theta_{\ell - 1}) \right\rangle + \frac{\delta^{2}}{2} \left\langle D_\ell^{\otimes 2}, \nabla^{2} u (\theta_{\ell - 1}) \right\rangle + R^u_\ell, \label{eq:taylor-u}
\end{align}
and the remainder satisfies,
\begin{equation}\label{eq:Ru-bound}
    \abs{R^u_\ell}\le\frac{\delta^3}{6}\,\norm{D_\ell}^3\sup_{\theta \in E_R^*}\opnorm{\nabla^3u(\theta)}.
\end{equation}

Rewriting the leading terms of \eqref{eq:taylor-u} by Doob's decomposition into previsible and martingale difference terms with respect to $(\cF_\ell)$ gives for any $\ell \le \tau_R / \delta$ steps,
$$
    u_\ell - u_{\ell - 1} = \delta \left[ A_\ell - A_{\ell - 1} \right] + \delta \left[ H_\ell - H_{\ell - 1} \right] + R^u_\ell,
$$
where the increments $A_\ell - A_{\ell - 1}$ and $H_\ell - H_{\ell - 1}$ may be explicitly written as,
\begin{align*}
    A_\ell - A_{\ell - 1} &= \left[ \left( - \cA + \delta \cL \right) (X_{\ell - 1} ) \right] u (\theta_{\ell - 1} ) \\
    H_\ell - H_{\ell - 1} &= -\left\langle D_\ell - \EE[D_\ell\mid\cF_{\ell-1}], \nabla u (\theta_{\ell - 1}) \right\rangle + \frac{\delta}{2} \left\langle D_\ell^{\otimes 2} - \EE[D_\ell^{\otimes 2} \mid\cF_{\ell-1}], \nabla^{2} u (\theta_{\ell - 1}) \right\rangle
\end{align*}
The problem now is that the previsible term implicitly depends on the entire parameter trajectory $(\theta_\ell)_\ell$ through $X_{\ell-1}$, whereas the statistic depends only $\theta_{\ell - 1}$. We need to show that the previsible component is well-approximated by its averaged counterpart. To this end, define the corrector $\chi$ as the solution to the Poisson equation
\begin{equation}\label{eq:poisson-chi}
        -\sL_\theta \chi^u (\theta , x ) = (I - \pi_{\theta}) [ ( - \cA + \delta \cL \Big) (x) ] u (\theta )
\end{equation}
As $u$ is clear from context we will suppress the dependence of $u$. We may explicitly write the solution \(\chi\) as
\[
    \chi (\theta, x) = \mathscr{P}_{\theta} \left[  \left( -\mathcal{A} + \delta \mathcal{L} \right)  (x) \right]\,u(\theta), \qquad \mathscr{P}_{\theta} = (-\mathscr{L}_{\theta})^{-1} \left( I - \pi_{\theta} \right).
\]
Substituting the Poisson equation in place of the difference,
\begin{equation}\label{eq:poisson-substituted}
    A_\ell - A_{\ell - 1} =( - \tilde \cA_{\theta_{\ell-1}} + \delta \tilde \cL_{\theta_{\ell-1}} ) u (\theta_{\ell - 1} )
    - \sL_\theta \chi (\theta_{\ell - 1}, X_{\ell - 1} )
\end{equation}
By definition, the one-step conditional mean under $K_\theta$ may be written as
$$
    K_{\theta} \chi (\theta , x ) = \EE_Y \left[ \chi (\theta , \Tsf_{\theta} ( x , Y ) ) \right] \implies K_{\theta_{\ell - 1}} \chi (\theta_{\ell - 1} , X_{\ell - 1} ) = \EE \left[ \chi (\theta_{\ell - 1} , X_\ell )\mid\cF_{\ell-1} \right],
$$
where we use that $X_\ell = \Tsf_{\theta_{\ell -1}} ( X_{\ell - 1} , Y_\ell )$. Define the martingale $\cM^\chi$ by the martingale difference $\Delta \cM^{\chi}_\ell= \chi  (\theta_{\ell - 1} , X_{\ell} ) - \EE \left[ \chi (\theta_{\ell - 1} , X_\ell )\mid\cF_{\ell-1} \right]$ . We may then re-write the above in a (nearly) discrete Ito type form as
\begin{equation}\label{eq:chi-three-way}
    \begin{aligned}
        -\sL_\theta \chi (\theta_{\ell - 1} , X_{\ell - 1} )  &= \Delta M^\chi_\ell +\Delta \chi_\ell+\Delta_\theta \chi_\ell
    \end{aligned}
\end{equation}
The first term is a martingale, the second term is telescoping, however the third term needs further estimates. To this end, first observe that by linearity of the Poisson equation, we may write $\chi=\chi_1+\delta\chi_2$, where $\chi_1$ and $\chi_2$ are centred solutions to
\[
    -\sL \chi_1 = (I-\pi_\theta)\cA(\theta,x)u\qquad -\sL \chi_2 = (I-\pi_\theta)\cL(\theta,x)u
\]
We now expand the $\chi_1$ part of the last term to first order, retaining the entire $\delta\chi_2$ parameter shift in the remainder, along $\theta_\ell-\theta_{\ell-1}=-\delta D_\ell$:
\begin{equation}\label{eq:corrector}
    \Delta_\theta \chi_\ell
    =-\delta\,B_u(\theta_{\ell-1},X_\ell)+\cR^{(1)}_{\chi,\ell},\qquad B_u(\theta_{\ell-1},X_\ell)=\bigl\langle D_\ell,\nabla_\theta\chi_1(\theta_{\ell-1},X_\ell)\bigr\rangle,
\end{equation}
with the remainder precisely given by the difference,
\begin{equation}\label{eq:Rchi}
    \cR^{(1)}_{\chi,\ell}:=\chi(\theta_\ell,X_\ell)-\chi(\theta_{\ell-1},X_\ell)-\nabla_\theta\chi_1(\theta_{\ell-1},X_\ell)[\theta_\ell-\theta_{\ell-1}] .
\end{equation}
Repeating the above averaging argument for $B_u$, we may write
\begin{equation}\label{eq:doob-B}
    B_u(\theta_{\ell-1},X_\ell)=b_u(\theta_{\ell-1},X_{\ell-1})+\Delta \cM^{B_u}_\ell
    \qquad b_u(\theta,x):=K_\theta B_u(\theta,\cdot)(x).
\end{equation}
whose stationary mean is $\pi_\theta b_u=\tilde\cA_\theta\chi_1$. Let $\xi$ solve the Poisson equation
\[
    -\sL_\theta\xi(\theta,x)=(I-\pi_\theta)\,b_u(\theta,x), \qquad \pi_\theta \xi (\theta , \cdot ) = 0.
\]
Substituting and repeating the same telescoping versus parameter-shift argument gives
\begin{equation}\label{eq:psi-three-way}
    \begin{aligned}
        b_u(\theta_{\ell-1},X_{\ell-1})&=\tilde\cA_{\theta_{\ell-1}}\chi_1(\theta_{\ell-1},\cdot)+ \Delta\cM^{\xi}_\ell
        + \Delta \xi_\ell +\Delta_\theta\xi_\ell
    \end{aligned}
\end{equation}
Grouping all of the terms together, \eqref{eq:taylor-u}--\eqref{eq:psi-three-way} yield the one-step increment identity
\begin{equation}\label{eq:one-step}
    \begin{aligned}
        u(\theta_\ell)-u(\theta_{\ell-1})=\delta \Psi^u_n(\theta_{\ell-1}) +
        \delta \Delta\cM^u_\ell +\delta ( \Delta \chi_\ell+\delta\Delta \xi_\ell)
        + ( R^u_\ell+\delta\,\cR^{(1)}_{\chi,\ell}-\delta^2\,\Delta_\theta\xi_\ell ),
    \end{aligned}
\end{equation}
where the averaged drift is defined in Assumption~\ref{assume:closable}. The total martingale increment is
\begin{equation}\label{eq:total-mart}
    \Delta \cM^u_\ell =  ( H_{\ell} - H_{\ell - 1})
    + \Delta \cM^{\chi}_\ell
    -\delta(\Delta \cM^{B_u}_\ell+\Delta \cM^\xi_\ell)
\end{equation}
Observe that the first term is the ``original'' increment and the remainders are terms that arise due to the corrections in the method of averaging. Summing \eqref{eq:one-step} over $\ell\le r$ steps, we obtain the \emph{stopped process decomposition}
\begin{equation}\label{eq:stopped-decomp}
    u(\theta_{r\wedge[ \tau_R / \delta ]})=u(\theta_0)+\delta\sum\nolimits_{\ell=1}^r \Psi_n^u(\theta_{\ell-1})+\sum\nolimits_{\ell=1}^r\Xi^u_\ell+E^u_r,\qquad \Xi^u_\ell:=\delta\,\Delta\cM^u_\ell,
\end{equation}
with the error term,
\begin{equation}\label{eq:error-def}
    E^u_r=\sum\nolimits_{\ell=1}^r \Bigl(R^u_\ell+\delta\,\cR^{(1)}_{\chi,\ell}-\delta^2\,\Delta_\theta\xi_\ell\Bigr)+\delta\bigl(\chi_0-\chi_{r\wedge[ \tau_R / \delta ]}\bigr)-\delta^2\bigl(\xi_0-\xi_{r\wedge[ \tau_R / \delta ]}\bigr),
\end{equation}
where $\chi_0=\chi(\theta_0,X_0)$. The four error terms in \eqref{eq:error-def} are controlled as follows (the proof is given in Section~\ref{sec:proof-bounds}).

\begin{theorem}[Negligibility of the remainders]\label{thm:main}
    Let Assumption~\ref{assume:localize} hold and fix $T>0$. Then, with every supremum taken over $1\le r\le \lfloor T / \delta \rfloor$
    \begin{align}
        \EE\Bigl[ \sup_r\Bigl|\sum_{\ell=1}^r ( R^u_\ell + \delta \cR^{(1)}_{\chi,\ell} - \delta^2 \Delta_\theta\xi_\ell ) \Bigr|\Bigr]&\le C_{K,T}\,\delta^{1/2},\label{eq:main0}\\
        \EE\Bigl[ \sup_r\Bigl|\delta\bigl(\chi_0-\chi_{r\wedge[ \tau_R / \delta ]}\bigr)-\delta^2\bigl(\xi_0-\xi_{r\wedge[ \tau_R / \delta ]}\bigr)\Bigr|\Bigr]&\le C_{K,T} \delta^{3 / 8}.\label{eq:main3}
    \end{align}
\end{theorem}

\subsection{Step 2: Tightness of the Limiting Process}\label{subsec:step2}

We now turn to tightness. Define the martingale and its predictable bracket as,
\begin{equation}\label{eq:N-C}
    \cN^K_n(t)=\sum\nolimits_{\ell\le t/\delta}\Xi_\ell,\qquad \Csf^K_n(t)=\sum\nolimits_{\ell\le t/\delta}\EE\bigl[\Xi_\ell\Xi_\ell^\top\bigm|\cF_{\ell-1}\bigr].
\end{equation}

This next estimate allows for us to establish that the martingale component vanishes in \eqref{eq:stopped-decomp}. The proof is provided in Section~\ref{sec:martingale}.
\begin{lemma} \label{lem:mart-increment}
    If Assumption~\ref{assume:localize} holds and $\delta = O(1/p)$, then for all $1 \le \ell \le \lfloor T / \delta \rfloor \wedge (\tau_R / \delta)$,
    \begin{equation}\label{eq:mart-increment}
        \bigl\| \Delta\cM_\ell\bigr\|_{L^{8/3}} \le C_{R,T}\bigl(\log(1/\delta) + \sqrt{p}\beta^\ell\bigr),
    \end{equation}
\end{lemma}

Recall that we may write the process \( u (\theta_{\ell})\) as,
\[
    u (\theta_{\ell}) = u (\theta_{0}) + \delta \sum_{\ell ' \leq \ell} \Psi^{u} ( \theta_{\ell' - 1}) + \delta \sum_{\ell ' \leq \ell} \Delta \cM_{\ell'}^{u} + o(1),
\]
where the last term vanishes as \(\delta \to 0\) in \(L^{1}\) uniformly for \(\ell \leq \tau_{R} / \delta\) steps. The linear interpolation \(\uv_{n} (t )\) for \(t \in [0, T]\) may then be written as,
\[
    \uv_{n} (t) = \uv_{n} (0) + \av_{n} (t) + \bv_{n} (t) + o(1),
\]
where we write \(\av_{n}\) as the continuous linear interpolation of the drift. To show tightness, let us define,
\[
    \vv_{n} (t) = \uv_{n} (0) + \av_{n} (t) + \bv_{n} (t).
\]
As the \(o(1)\) error above is uniform in \(t\), it suffices to show the claimed tightness of limit points of \(\vv_{n}\) instead of \(\uv_{n}\). We do so by showing the Kolmogorov continuity criterion:

\begin{proposition}[Tightness]\label{prop:tightness}
    Let Assumptions~\ref{assume:localize}--\ref{assume:closable} hold and suppose that $(\uv_n(\theta_0))_n$ is tight. Then the sequence $(\vv_{n} (s \land \tau_{R}))_n$ is tight in $C([0,T];\RR^{k})$. In particular, we have that for all \(0 \leq s, t \leq T\),
    \begin{equation} \label{eq:tight-kolmo-bound}
        \mathbb{E} \lVert \vv_{n} (t \land \tau_{R}) - \vv_{n} (s \land \tau_{R}) \rVert^{8/3} \leq C_{K, T} (t - s)^{4/3},
    \end{equation}
    from which the sequence \(\vv_{n} (s \land \tau_{R})\) is uniformly \(\gamma\)-Holder for all \(\gamma < 1 / 8\).
\end{proposition}

\begin{proof}
    For all \(s, t\), we have the decomposition,
    \[
        \lVert \vv_{n} (t ) - \vv_{n} (s) \rVert \leq \lVert \av_{n} (t) - \av_{n} (s) \rVert + \lVert \bv_{n} (t) - \bv_{n} (s) \rVert.
    \]
    We control each of these terms in turn, and we will do so coordinate-wise. Fix the statistic \(u\) of the collection \(\boldsymbol{u}\) and further abbreviate the corresponding component of \(\av\) and \(\bv\) by \(a, b\). For the previsible term, we have that,
    \begin{equation} \label{eq:tight-previsible}
        \mathbb{E} \left| a ( t \land \tau_{R} ) - a ( s \land \tau_{R} ) \right|^{8/3} \leq \mathbb{E} | \delta \sum\nolimits_{\ell} ( ( - \tilde{\mathcal{A}_{\theta}} + \delta \tilde{\mathcal{L}_{\theta}} ) u - \delta\tilde{\mathcal{A}_{\theta}} \chi_1  )_{\ell} |^{8 / 3},
    \end{equation}
    where these sums are over steps \(\ell\) ranging from \([s / \delta ] \land \tau_{R} / \delta \) to \([t / \delta ] \land \tau_{R} / \delta \). Let \(\Psiv = (\Psi_{j})_{j \leq k}\) denote the coordinate-level drift terms. Then by Assumption~\ref{assume:closable},
    \begin{align*}
        \mathbb{E} | \delta \sum\nolimits_{\ell} ( ( - \tilde{\mathcal{A}_{\theta}} + \delta \tilde{\mathcal{L}_{\theta}} ) u - \delta\tilde{\mathcal{A}_{\theta}} \chi_1  )_{\ell} |^{8 / 3} & \lesssim_{K} \mathbb{E} | \delta \sum\nolimits_{\ell} ( \Psi (\uv_{n})  )_{\ell} |^{8 / 3} + o((t - s)^{8 / 3}) \\[2mm]
        &\lesssim_{K} (t - s)^{8 / 3} ( \lVert \Psi \rVert^{8/3}_{L^{\infty} (E_{R}^{*})} + o(1) ) \lesssim_{K} (t - s)^{8/3,}
    \end{align*}
    by continuity of \(\Psi\) in \(E_{R}\). For the martingale term, we notice that by Burkholder's inequality,
    \[
        \mathbb{E} | b ( t \land \tau_{R} ) - b ( s \land \tau_{R} ) |^{8/3} = \mathbb{E} | \delta \sum\nolimits_{\ell} \Delta \cM_{\ell} |^{8/3} \lesssim \mathbb{E} ( \delta^{2} \sum\nolimits_{\ell} \Delta \cM_{\ell}^{2} )^{4/3}.
    \]
    Taking the power \(p = 3 / 4\) of both sides of the expression converts this inequality with respect to \(L^{p}\) norms. In particular, we may write,
    \begin{align*}
        \left\| b(t\land\tau_R)- b(s\land\tau_R)\right\|_{L^{8/3}}^2
        &\lesssim\delta^2\sum\nolimits_\ell\norm{\Delta\cM_\ell}_{L^{8/3}}^2
        \lesssim_{K,T}\delta^2\sum\nolimits_\ell\bigl(\log(1/\delta)^2 + p\beta^{2\ell}\bigr)\lesssim_{K,T}\delta^{1/2}|t-s|,
    \end{align*}
    by Lemma~\ref{lem:mart-increment}. Raising both sides by \(p = 4 / 3\) provides the estimate,
    \[
        \mathbb{E} \left| b ( t \land \tau_{R} ) - b ( s \land \tau_{R} ) \right|^{8/3} \lesssim_{K} \delta^{2/3} | t - s |^{4/3}.
    \]
    Apply this bound for \(s = 0\) and \(t = T\) shows that the martingale component of \(\vv_{n}\) is vanishing in the limit for each \(K\) whenever \(\delta_{n} \to 0\). Combining both the previsible and martingale components yields \eqref{eq:tight-kolmo-bound}. By Kolmogorov's continuity theorem, it follows that \( (\vv_{n} (s \land \tau_{R}))_{n}\) is tight with the claimed H\"{o}lder regularity for the limit points.
\end{proof}

\subsection{Step 3: Identification of Limit Points using Martingale Problem}\label{subsec:step3}

By Proposition~\ref{prop:tightness}, the stopped processes are tight. Moreover, Assumption~\ref{assume:closable} identifies the limiting generator by,
\(
    \mathsf L
    =\sum_{i=1}^{k}\Psi_i \partial_i.\)
Combined with Assumption~\ref{assume:closable}, we therefore have that
\[  \sup\nolimits_{0 \le t \le T} \| \delta \sum\nolimits_{\ell \le (\tau_R \land t ) / \delta} \Psi_{n} ( u (\theta_{\ell - 1}) ) - \int^{t \land \tau_R}_0 \Psi ( \uv_n^R ( s ) ) \, \text{d}s \| \leq T \, \sup\nolimits_{\theta \in E_{R}^{*}} | \Psi_{n} \left( \theta \right) - \Psi \left( \theta \right) | \to 0.
\]
We can now complete the proof: Fix $R,T>0$. Since $(\uv_n^R)_n$ is tight, and every subsequential limit stopped at $\tau_R \land T$ satisfies,
$$
    \varphi (\vv_{t}) - \varphi (\vv_{0}) - \int^{t}_{0} \mathsf{L} \varphi (\vv_{s}) \, \dd s = 0,
$$
for every $\varphi \in C_0 ( \RR^{k} )$ by Ito's formula. Since $\Psiv$ is locally Lipschitz, this uniquely characterizes the solutions to \eqref{eq:main-adam-sde} stopped at $\tau_R$. By lower semicontinuity of $\tau_R$ \cite[Lemma~11.1.2]{stroock2006multidimensional} and standard localization arguments \cite[Lemma~11.1.1]{stroock2006multidimensional}, it follows that $\uv_n \Rightarrow \uv$ in $C ([0, T] ; \RR^{k} )$ as desired. \qed

\section{Well-posedness of the Poisson equation} \label{sec:poisson}

In this section we establish the well-posedness of the Poisson equation for the fixed-parameter momentum chain $X_n^\theta$ on $\cX$ defined in \eqref{eq:unified-momentum-update}--\eqref{eq:fixed-momentum}. We begin by identifying its unique invariant probability measure $\pi_\theta$, after which we study the existence, uniqueness, regularity, continuous dependence, and differentiable dependence of solutions.  

Recall from Section~\ref{sec:setup} the space $\cV_{r,S}^k$ with $V_r(x) = 1 + \norm{x}_2^r$ for $S, r \ge 1$ and $k \ge 0$. Given a probability measure $\mu$, let $\cV_{r,S,\mu}^k := \{f \in \cV_{r,S}^k : \mu(f) = 0\}$. Throughout the following, we abuse notation by writing $\beta = \opnorm{B} \in [0, 1)$ and make repeated use of the inequalities
\(
    V_r(a + b) \lesssim_r V_r(a) + V_r(b) \lesssim_r V_r(a) V_r(b).
\) 
We begin with existence and uniqueness.

\begin{lemma} \label{lem:pi-ergodic}
    Suppose Assumption~\ref{assume:localize} holds, we have that the fixed-parameter momentum chain $X_\ell = B X_{\ell - 1} + (1 - B ) \psi_\theta (Y_{\ell} )$ with $Y_\ell \sim P_n$ i.i.d. admits a unique, invariant law $\pi_\theta$.
\end{lemma}

\begin{proof}
    First, by Assumption~\ref{assume:localize} and Definition~\ref{def:rho}, we have that quadratic growth of $\psi_{\theta, i}$ implies that $\| \psi_{\theta, i} \|_{L^r} \le C_R$ for each coordinate $i$ and $1 \le r \le 8$. Furthermore, by Minkowski's inequality for $r \ge 2$, we have the vector bound $\| \psi_\theta (Y) \|_{L^r}^2 \le \sum_{i=1}^{p_n} \| \psi_{\theta, i} (Y) \|^2_{L^{r}} \le C_R p$. For $1 \le r < 2$, the same follows by monotonicity of the norm.
    Next, for a two-sided i.i.d. sequence $(Y_j )_{j \in \ZZ}$, the sequence $\tilde X_\ell = (1 - B ) \sum^\infty_{j=0} B^j \psi_\theta (Y_{\ell - j} )$ satisfies,
    $$
    \| \tilde X_\ell \|_{L^r} \le \sum\nolimits^\infty_{j = 0} \beta^j \| \psi_\theta (Y_{\ell - j} ) \|_{L^r} \le (1 - \beta )^{-1} C_R \sqrt{p},
    $$
    thus the series converges. We note that invariance is trivial since $B \tilde {X}_{\ell - 1} + (I - B ) \psi_\theta (Y_\ell ) = \tilde X_\ell$ and $Y_j$ are i.i.d. are unchanged by shifting $\ell$. Finally, to prove uniqueness couple the chains $X_\ell^x$ and $\tilde X_\ell$ with the same data $(Y_j )_{j=1}^\ell$ where $X_0^x \sim \nu$ for another invariant measure $\nu$ and $\tilde X_0 \sim \pi_\theta$. Then $\| X_\ell^x - \tilde X_\ell \| \le \beta^\ell \| x - \tilde X_0 \| \to 0$ almost surely. By Slutsky's, $X^x_\ell = \tilde X_\ell + (X_\ell^x - \tilde X_\ell ) \Rightarrow \pi_\theta$ weakly, whereas $X^x_\ell \Rightarrow \nu$ for every $\ell \ge 0$ trivially. Thus $\pi_\theta = \nu$ by uniqueness of weak limits.
\end{proof}

The preceding lemma establishes that the frozen momentum dynamics possess a canonical stationary distribution. Moreover, the geometric coupling between arbitrary trajectories and the stationary process is the key ingredient in solving the Poisson equation.

\begin{lemma} \label{lem:poisson-wellposed}
    Suppose for $r, k \ge 1$ and $S \ge 1$ that $\EE\norm{\psi_\theta}_2^r < \infty$ and $f \in \cV_{r,S}^k$. The Poisson equation
    \begin{equation}\label{eq:general-poisson}
        \begin{cases}
            -\sL_\theta u_\theta = f_\theta - \pi_\theta f_\theta, \\
            \pi_\theta u_\theta = 0
        \end{cases}
    \end{equation}
    has a unique solution in $\cV_{r,S,\pi_\theta}^k$. In particular, for $f \in \cV_{r,S,\pi_\theta}^k$,
    \[
        \norm{(-\sL_\theta)^{-1} f}_{\cV_{r,S}^k} \le \frac{C_r\bigl(1 + S^{-r}\EE\norm{\psi_\theta}_2^r\bigr)}{1 - \beta^r} \norm{f}_{\cV_{r,S}^k}.
    \]
    Furthermore, if $f \in \cV_{r,S_I}^k$ depends only on coordinates indexed by $I \subseteq [p]$ and we set $S = S_I := \abs{I}^{1/2}$, then
    \begin{equation}\label{eq:poisson-few-coords}
        \norm{(-\sL_\theta)^{-1} f}_{\cV_{r,S_I}^k} \le \frac{C_r}{1 - \beta^r} ( 1 + \sup_{i \in I} \norm{\psi_i}_{L^r}^r ) \norm{f}_{\cV_{r,S_I}^k}.
    \end{equation}
\end{lemma}
\begin{proof}
    We suppress $\theta$ dependence for readability and assume without loss of generality that $\pi_\theta f = 0$. Given $x \in \cX$, couple $X_n^x$ and $\tilde{X}_n$ using the data stream $(Y_\ell)_{\ell = -\infty}^n$ so that
    \[
        \norm{X_n^x - \tilde{X}_n} \le \beta^n \norm{x - \tilde{X}_0}, \qquad \tilde{X}_0 \sim \pi_\theta.
    \]
    Let $\EE_x$ denote expectation under the chain initialized at $x$. We first estimate $K^n f$. We have,
    \[
        \abs{K_\theta^n f(x)} = \abs{\EE_x[f(X_n) - f(\tilde{X}_n)]}
        \lesssim_r \beta^n \norm{f}_{\cV_{r,S}^1} \bigl(V_r(x/S) + \EE V_r(\tilde{X}_0/S)\bigr)
        \lesssim_r \beta^n \EE\bigl[V_r(\tilde{X}_0/S)\bigr] V_r(x/S) \norm{f}_{\cV_{r,S}^1}.
    \]
    Since $\EE V_r(\tilde{X}_0/S) \le (1 - \beta^r)^{-1}\bigl(1 + S^{-r}\EE\norm{\psi_\theta}_2^r\bigr)$ and $\opnorm{S^\ell \nabla_x^\ell K_\theta^n f(x)} \le \beta^{\ell n} K_\theta^n \opnorm{S^\ell \nabla_x^\ell f}(x)$, we obtain
    \[
        \norm{K_\theta^n f}_{\cV_{r,S}^k} \lesssim_r \frac{\beta^n}{1 - \beta^r}\bigl(1 + S^{-r}\EE\norm{\psi_\theta}_2^r\bigr) \norm{f}_{\cV_{r,S}^k}.
    \]
    Consequently, the Neumann series $u(x) = \sum_{n=0}^\infty K_\theta^n(f(x) - \pi_\theta f)$ converges absolutely in $\cV_{r,S}^k$ and satisfies the stated bound.

    For uniqueness, if $u_1, u_2 \in \cV_{r,S,\pi_\theta}^k$ both solve \eqref{eq:general-poisson}, their difference $w = u_1 - u_2$ satisfies $\sL_\theta w = 0$ and $\pi_\theta w = 0$, hence $w = K_\theta^n w$ and
    \[
        \norm{w}_{\cV_{r,S}^0} = \norm{K_\theta^n w}_{\cV_{r,S}^0} \le C_{r,S} \beta^n \norm{w}_{\cV_{r,S}^1} \to 0 \qquad \text{as } n \to \infty.
    \]
    Finally, for $r \ge 2$, if $f(x) = \bar{f}(x_I)$ depends only on $I \subseteq [p]$ and $S_I = \abs{I}^{1/2}$, Minkowski's inequality gives
    \[
        \norm{\psi_I}_{L^r} = \| \sum_{i \in I} \abs{\psi_i}^2 \|_{L^{r/2}}^{1/2} \le ( \sum_{i \in I} \norm{\psi_i}_{L^r}^2 )^{1/2} \le S_I \sup_{i \in I} \norm{\psi_i}_{L^r},
    \]
    so $S_I^{-r} \EE\norm{\psi_I}_2^r \le \sup_{i \in I} \norm{\psi_i}_{L^r}^r$, yielding \eqref{eq:poisson-few-coords}.
\end{proof}

We now turn to continuous dependence. In the following, let $\mathscr{W}_1$ denote the class of $1$-Wasserstein signed measures, i.e., those finite signed measures $\mu$ with $\abs{\mu}(V_r(\cdot/S)) < \infty$.

\begin{lemma}\label{lem:weak-compactness}
    Suppose that $f_n, f \in \cV_{r,S}^1$ with $f_n \to f$ pointwise and $\sup_n \norm{f_n}_{\cV_{r,S}^1} < \infty$. Then
    \[
        (-\sL)^{-1} f_n \to (-\sL)^{-1} f
    \]
    locally uniformly.
\end{lemma}

\begin{proof}
    We use a weak compactness argument. Let $u_n = (-\sL)^{-1} f_n$ and $u = (-\sL)^{-1} f$. By boundedness of $(-\sL)^{-1}$ we have that $u_n$ are locally bounded and locally uniformly equicontinuous. Thus by Arzel\`a--Ascoli, it is precompact in the topology of locally uniform convergence. Take a subsequence on which $u_{n_i} \to v$ locally uniformly. Since $\tilde{X}_0 \in L^1(\pi)$, i.e., $\pi(V_r(\cdot/S)) < \infty$, and $u_n, v$ are dominated by $V_r(\cdot/S)$ (up to a constant), $\pi(u_n) \to \pi(v)$ by dominated convergence so $\pi(v) = 0$. Let $\mu \in \mathscr{W}_1$ be a probability measure. By dominated convergence $\langle f_n, \mu \rangle \to \langle f, \mu \rangle$. On the other hand, for any such measure $(-\sL)^* \mu$ is a finite signed measure, with $\abs{(-\sL)^* \mu}(V_r(\cdot/S)) = \mu(V_r(\cdot/S)) + K^* \mu(V_r(\cdot/S)) < \infty$. So by dominated convergence
    \[
        \langle -\sL u_n, \mu \rangle = \langle u_n, (-\sL)^* \mu \rangle \to \langle v, (-\sL)^* \mu \rangle
    \]
    so that $\langle -\sL v, \mu \rangle = \langle f, \mu \rangle$. Taking $\mu = \delta_x$ yields $-\sL v = f$. By uniqueness we conclude $v = u$.
\end{proof}

In the following,  let $\mathfrak{L}_{r,S}^k$ denote the space of linear functionals, $\ell \in (\cV_{r,S}^k)^*$, with the following property: if $\phi_n\to 0$ pointwise and $\sup_n \norm{\phi_n}_{\cV_{r,S}^k} < \infty$, then $\ell(\phi_n) \to 0$. Observe the following fact whose proof is elementary and left to the reader.

\begin{lemma}\label{lem:locality}
    The following hold for all $k \ge 0$:
    \begin{enumerate}
        \item $\mathfrak{L}_{r,S}^k$ is a norm-closed subspace of $(\cV_{r,S}^k)^*$.
        \item If $(\ell_n)$ converges in $\mathfrak{L}_{r,S}^k$ and $\phi_n \to 0$ pointwise with $\sup_n \norm{\phi_n}_{\cV_{r,S}^k} < \infty$, then $\ell_n(\phi_n) \to 0$.
        \item $\mathscr{W}_1 \subset \mathfrak{L}_{r,S}^k$.
    \end{enumerate}
\end{lemma}

Writing $\phi_\theta := (I - B)\psi_\theta$, define the operators:
\begin{align*}
    \nabla_\theta K_\theta f(x) &= \EE\bigl[\nabla f(B x + \phi_\theta)^\top \nabla_\theta \phi_\theta\bigr], &
    \nabla_\theta^2 K_\theta f(x) &= \EE\bigl[\nabla^2 f(\nabla_\theta \phi_\theta, \nabla_\theta \phi_\theta) + \nabla f^\top \nabla_\theta^2 \phi_\theta\bigr], \\
    \nabla_\theta \pi_\theta f &= \EE\bigl[\nabla f(\tilde{X}_0)^\top \nabla_\theta \tilde{X}_0\bigr], &
    \nabla_\theta^2 \pi_\theta f &= \EE\bigl[\nabla^2 f(\nabla_\theta \tilde{X}_0, \nabla_\theta \tilde{X}_0) + \nabla f^\top \nabla_\theta^2 \tilde{X}_0\bigr].
\end{align*}
Observe that these operators are well-defined for $f \in \cV_{r, S}^2$. Let $D_v K_\theta = v \cdot \nabla_\theta K_\theta$ and $D_v \pi_\theta = v \cdot \nabla_\theta \pi_\theta$ denote the corresponding directional derivatives. For any $v, w \in \RR^p$, define the following finite difference quotients:
\begin{align}
    A_h^{(1),\theta,v} f &= \frac{K_{\theta+hv} - K_\theta}{h} f, &
    A_h^{(2),\theta,v,w} f &= \frac{D_v K_{\theta+hw} - D_v K_\theta}{h} f, \\
    B_h^{(1),\theta,v} &= \frac{\pi_{\theta+hv} - \pi_\theta}{h}, &
    B_h^{(2),\theta,v,w} &= \frac{D_v \pi_{\theta+hw} - D_v \pi_\theta}{h}.
\end{align}
Let $\cB(X,Y)$ denote the space of bounded linear operators from $X \to Y$. Our main technical part of this section is to prove a family of estimates on these difference operators and the corresponding directional derivative operators. In the following we write,
\[
    \norm{\psi(\theta)}_{\cR_q^2} = \sup_i ( \norm{\psi_i(\theta)}_{L^q} + \sup_{\norm{v'} \le 1} \norm{D_{v'} \psi_i(\theta)}_{L^q} + \sup_{\norm{v'}, \norm{w'} \le 1} \norm{D_{w'} D_{v'} \psi_i(\theta)}_{L^q} )
\]
and set $\norm{\psi}_{\cR_q^2(A)} = \sup_{\theta \in A} \norm{\psi(\theta)}_{\cR_q^2}$.

\begin{lemma}\label{lem:operator-estimates}
    Fix $1 \le r < 5$, $S \ge 1$, and $A \subset \RR^p$ open, and suppose that Assumption~\ref{assume:localize}--\ref{item:A-grad} holds on $A$.
    \begin{enumerate}
        \item For any $v \in \RR^p$, $\{(A_h^{(1),\theta,v})_{h>0}, D_v K_\theta\}$ is uniformly bounded in $\cB(\cV_{r,S}^3, \cV_{r,S}^2)$. Furthermore, $A_h^{(1),\theta,v} \to D_v K_\theta$ in $\cB(\cV_{r,S}^3, \cV_{r,S}^1)$.

        \item For any $v, w \in \RR^p$, $\{(A_h^{(2),\theta,v,w})_{h>0}, D_w D_v K_\theta\}$ is uniformly bounded in $\cB(\cV_{r,S}^3, \cV_{r,S}^1)$, and $A_h^{(2),\theta,v,w} \to D_w D_v K_\theta$ in $\cB(\cV_{r,S}^3, \cV_{r,S}^0)$.

        \item For any $v \in \RR^p$, the family of linear functionals $\{(B_h^{(1),\theta,v})_{h>0}, D_v \pi_\theta\}$ is uniformly bounded in $(\cV_{r,S}^1)^*$, and $B_h^{(1),\theta,v} \to D_v \pi_\theta$ in the norm topology.

        \item For any $v, w \in \RR^p$, $\{(B_h^{(2),\theta,v,w})_{h>0}, D_w D_v \pi_\theta\}$ is uniformly bounded in $(\cV_{r,S}^2)^*$, and $B_h^{(2),\theta,v,w} \to D_w D_v \pi_\theta$ in the norm topology.

        \item If $T$ is any operator from Items~(1)--(2), then $T^* \delta_x \in \mathfrak{L}_{r,S}^3$. Moreover, every functional $\ell$ in Items~(3)--(4) satisfies $\ell \in \mathfrak{L}_{r,S}^3$.

        \item If $f \in \cV_{r,S_I}^3$ depends only on coordinates $I \subseteq [p]$ and $S_I = \abs{I}^{1/2}$, then
        \begin{align*}
            \norm{D_v K_\theta f}_{\cV_{r,S_I}^2} &\lesssim_{r} \norm{f}_{\cV_{r,S_I}^3} \bigl( 1 + \norm{\psi}_{\cR_{r+1}^2(A)}^r \bigr) \norm{\psi}_{\cR_{r+1}^2(A)} \norm{v}, \\
            \norm{D_w D_v K_\theta f}_{\cV_{r,S_I}^1} &\lesssim_{r} \norm{f}_{\cV_{r,S_I}^3} \bigl( 1 + \norm{\psi}_{\cR_{r+2}^2(A)}^r \bigr) \bigl( \norm{\psi}_{\cR_{r+2}^2(A)} + \norm{\psi}_{\cR_{r+2}^2(A)}^2 \bigr) \norm{v} \norm{w}, \\
            \abs{D_v \pi_\theta f} &\lesssim_{r} \norm{f}_{\cV_{r,S_I}^1} \bigl( 1 + \norm{\psi}_{\cR_{r+1}^2(A)}^r \bigr) \norm{\psi}_{\cR_{r+1}^2(A)} \norm{v}, \\
            \abs{D_w D_v \pi_\theta f} &\lesssim_{r} \norm{f}_{\cV_{r,S_I}^2} \bigl( 1 + \norm{\psi}_{\cR_{r+2}^2(A)}^r \bigr) \bigl( \norm{\psi}_{\cR_{r+2}^2(A)} + \norm{\psi}_{\cR_{r+2}^2(A)}^2 \bigr) \norm{v} \norm{w}.
        \end{align*}
    \end{enumerate}
\end{lemma}
\begin{proof}
    \textbf{Item (1):} Fix $v$, abbreviate $A_h = A_h^{(1),\theta,v}$, and set $D_h = (\phi_{\theta + hv}(Y) - \phi_\theta(Y)) / h$. By the mean-value theorem, for $j \in \{0, 1, 2\}$ and $v_1, \dots, v_j \in \cX$,
    \begin{align*}
        \nabla_x^j (A_h f)(x)[v_1, \ldots, v_j] &= \EE\left[ \int_0^1 \nabla_x^{j+1} f(B x + \phi_\theta + t h D_h)[B v_1, \ldots, B v_j, D_h] \,\dd t \right], \\[1mm]
        \nabla_x^j (D_v K_\theta f)(x)[v_1, \ldots, v_j] &=  \EE\left[ \nabla_x^{j+1} f(B x + \phi_\theta)[B v_1, \ldots, B v_j, D_v \phi_\theta] \right].
    \end{align*}
    Using $S^{j+1}\opnorm{\nabla_x^{j+1} f(z)} \le \norm{S^{j+1}\nabla_x^{j+1} f}_{\cV_{r,S}^0} V_r(z/S)$, we obtain for $f \in \cV_{r,S}^3$ and $j \in \{0, 1, 2\}$,
    \begin{equation}\label{eq:Ah-weighted-grad}
        \frac{S^j \opnorm{\nabla_x^j (A_h f)(x)}}{V_r(x/S)}
        \lesssim_r \beta^j \norm{S^{j+1}\nabla_x^{j+1} f}_{\cV_{r,S}^0} \EE\Bigl[\bigl(V_r(\phi_\theta/S) + V_r(h D_h/S)\bigr)\frac{\norm{D_h}_2}{S}\Bigr].
    \end{equation}
    By Assumption~\ref{assume:localize}--\ref{item:A-grad}, $\norm{D_h}_{L^q} \le C_{A,p}\norm{v}$ uniformly for $q \le 8$. H\"older's inequality with exponents $(r+1)/r$ and $r+1$ therefore bounds the expectation in \eqref{eq:Ah-weighted-grad}, giving uniform boundedness of $A_h$ (and identically $D_v K_\theta$ using $\nabla^j_x D_v K$) in $\cB(\cV_{r,S}^3, \cV_{r,S}^2)$.

    For convergence, fix $\norm{f}_{\cV_{r,S}^3} \le 1$. Expanding $(A_h - D_v K_\theta) f(x)$ yields, for $j \in \{0, 1\}$,
\[
        \norm{S^j \nabla_x^j(A_h - D_v K_\theta) f}_{\cV_{r,S}^0} \lesssim_r \EE[V_r(\phi_\theta/S)\frac{\norm{D_h - D_v \phi_\theta}_2}{S}] + \abs{h} \EE[(V_r(\phi_\theta/S) + V_r(h D_h/S))\frac{\norm{D_h}_2^2}{S^2}].
\]
    Since $D_h \to D_v \phi_\theta$ almost surely and is uniformly bounded in $L^8$, Vitali's convergence theorem gives $\norm{D_h - D_v \phi_\theta}_{L^q} \to 0$ for all $q < 8$, so the right-hand side of the above
    vanishes as $h \to 0$.

    \smallskip
    \textbf{Item (2):} Fix $v, w$ and set $E_h = (D_v \phi_{\theta + hw} - D_v \phi_\theta) / h$. Applying the same argument to $A_h^{(2)} = A_h^{(2),\theta,v,w}$ gives, for $j \in \{0, 1\}$,
    \[
        \norm{S^j \nabla_x^j A_h^{(2)} f}_{\cV_{r,S}^0} \lesssim_{r, \beta} \norm{f}_{\cV_{r,S}^3} \EE[ V_r(\phi_\theta/S)\frac{\norm{E_h}_2}{S} + \bigl(V_r(\phi_\theta/S) + V_r(h D_h/S)\bigr)\frac{\norm{D_h}_2 \norm{D_v \phi_\theta}_2}{S^2} ],
    \]
    and for $\norm{f}_{\cV_{r,S}^3} \le 1$,
    \begin{align*}
        \norm{(A_h^{(2)} - D_w D_v K_\theta) f}_{\cV_{r,S}^0} &\lesssim_r \EE[V_r(\phi_\theta/S)(\frac{\norm{E_h - D_w D_v \phi_\theta}_2}{S} + \frac{\norm{D_h - D_w \phi_\theta}_2 \norm{D_v \phi_\theta}_2}{S^2})] \\
        &\qquad + \abs{h} \EE[(V_r(\phi_\theta/S) + V_r(h D_h/S))(\frac{\norm{D_h}_2\norm{E_h}_2}{S^2} + \frac{\norm{D_h}_2^2 \norm{D_v \phi_\theta}_2}{S^3})] \to 0.
    \end{align*}

    \smallskip
    \textbf{Items (3)--(5):} Items~(3) and~(4) follow identically to Items~(1) and~(2) upon  replacing $\phi_\theta$ by $\tilde{X}_0(\theta)$. Item~(5) follows from Lemma~\ref{lem:locality}.

    \smallskip
    \textbf{Item (6):} When $f(x) = \bar{f}(x_I)$ with $S_I = \abs{I}^{1/2}$, the expectations in Items~(1)--(4) with $S = S_I$ depend only on the $I$-coordinates $\phi_I$ and $\tilde{X}_{0,I}$. Since Minkowski's inequality gives
    \[
        \norm{\nabla_\theta^a \phi_I[v_1, \dots, v_a]}_{L^q} \le S_I \norm{\psi}_{\cR_q^2(A)} \prod_{m=1}^a \norm{v_m}_2, \qquad a \in \{0, 1, 2\},
    \]
    and the same bound holds for $\tilde{X}_{0,I}$ via \eqref{eq:fixed-momentum}, applying H\"older's inequality to the $D_v K_\theta$, $D_w D_v K_\theta$, $D_v \pi_\theta$, and $D_w D_v \pi_\theta$ expectations from Items~(1)--(4) yields all four estimates.
\end{proof}

We will require more regularity in our results, namely differentiable dependence of the operator on $\theta$. Let us define for $k \in \{2, 3\}$, the space $\cW_{r,S}^k(A)$ as the space of functions in $(\theta, x)$ equipped with the norm
\[
    \norm{f}_{\cW_{r,S}^k(A)} = \sum_{a=0}^{k-1} \sup_{\theta \in A} \sup_{\norm{h_1}, \ldots, \norm{h_a} \le 1} \norm{\nabla_\theta^a f_\theta[h_1, \ldots, h_a]}_{\cV_{r,S}^{k-a}},
\]

\begin{lemma} \label{lem:poisson-first-deriv-pde}
    Suppose that $f \in \cW_{r,S}^2(A)$. Then the solution $u_\theta$ of the Poisson equation \eqref{eq:general-poisson} is $C^1$ in $\theta$ on $A$ and, for any $v \in \RR^p$, the directional derivative $D_v u_\theta \in \cV_{r,S}^1$ solves
    \begin{equation}\label{eq:differentiated-PDE}
        \begin{cases}
            -\sL_\theta D_v u_\theta = (I - \pi_\theta) D_v f_\theta - (D_v \pi_\theta) f_\theta + (D_v \sL_\theta) u_\theta, \\
            \pi_\theta D_v u_\theta = -(D_v \pi_\theta) u_\theta.
        \end{cases}
    \end{equation}
    Furthermore, if $f$ depends only on coordinates $I \subseteq [p]$ and $S_I = \abs{I}^{1/2}$, then
    \begin{equation} \label{eq:poisson-1st-few-coords}
        \norm{\nabla_\theta u_\theta}_{\cV_{r,S_I}^1} \lesssim_{r} \frac{\norm{f}_{\cW_{r,S_I}^2(A)}}{(1 - \beta^r)^2} \bigl(1 + \norm{\psi}_{\cR_{r+1}^2(A)}^r\bigr)^3 \bigl(1 + \norm{\psi}_{\cR_{r+1}^2(A)}\bigr).
    \end{equation}
\end{lemma}

\begin{proof}
    The argument follows the standard differentiable dependence argument for Poisson equations: We first show that the differentiated equation admits a solution, and then show that the difference quotient converges pointwise to this solution.

    Fix $v$ and the solutions at $\theta + hv$ and $\theta$, call them $u_h$ and $u_0$ respectively. Similarly define $K_h, \pi_h,$ and $\sL_h$. Let $D_v$ denote the directional derivative in the $v$ direction. We begin by observing that \eqref{eq:differentiated-PDE} admits a unique solution. It suffices to show that the forcing is in $\cV_{r,S}^1$ and centred with respect to $\pi_\theta$. This follows by assumption on $f$ and Lemma~\ref{lem:operator-estimates} for the first two terms. For the third, note that Lemma~\ref{lem:poisson-wellposed} gives $u_\theta \in \cV_{r,S}^2$, so again by Lemma~\ref{lem:operator-estimates} the third term is in $\cV_{r,S}^1$. Thus by Lemma~\ref{lem:poisson-wellposed}, there is a solution that is uniformly in $\cV_{r,S}^1$ for each $\theta \in A$.

    Notice that each of the two solutions $u_h, u_0$ solves the Poisson equations
    \[
        -\sL_h u_h = f_h - \pi_h f_h, \qquad -\sL_0 u_0 = f_0 - \pi_0 f_0,
    \]
    from which their difference $w_h = u_h - u_0$ solves (by subtracting the equations and rearranging)
    \[
        -\sL_0 w_h = (I - \pi_0)(f_h - f_0) - (\pi_h - \pi_0) f_h + (\sL_h - \sL_0) u_h =: \xi_h.
    \]
    Observe that $\pi_0 \xi_h = 0$ (since $\pi_0 \sL_0 = 0$). Thus we may write
    \[
        u_h - u_0 = (-\sL_0)^{-1}(\xi_h) - (\pi_h - \pi_0) u_h.
    \]
    Let us consider the difference quotient $w_h / h$ and prove convergence. We will do this by looking at the terms in the above, term by term.

    We first show that $\xi_h / h$ and $B_h u_h$ are uniformly in $\cV_{r,S}^1$. By definition of $\cW_{r,S}^2(A)$, $(f_h - f_0)/h$ is uniformly bounded in $\cV_{r,S}^1$. Furthermore, since $f_h$ is uniformly in $\cV_{r,S}^2$, Lemma~\ref{lem:operator-estimates} implies that $B_h f_h$ is uniformly bounded as well. Finally, by Lemma~\ref{lem:poisson-wellposed}, $u_h$ is uniformly in $\cV_{r,S}^2$, so by Lemma~\ref{lem:operator-estimates}, both $A_h u_h$ and $B_h u_h$ are uniformly bounded as well.

    We now show pointwise convergence. Let us look at the $(-\sL_0)^{-1} \xi_h / h$ term. The first term is clear. For the second term write
    \[
        B_h f_h - B f_0 = (B_h - B) f_h + B(f_h - f_0).
    \]
    The first term vanishes since $B_h \to B$ in norm and $f_h$ are bounded. The second vanishes since $B \in \mathfrak{L}_{r,S}^2$ and $f_h \to f_0$ pointwise and are uniformly bounded in $\cV_{r,S}^2$, by Lemmas~\ref{lem:operator-estimates} and~\ref{lem:locality}. The third term follows the same logic after noting that by Lemmas~\ref{lem:poisson-wellposed}--\ref{lem:weak-compactness}, $u_h \to u_0$ locally uniformly and they are uniformly in $\cV_{r,S}^2$, and that $(A_h)^* \delta_x \in \mathfrak{L}_{r,S}^2$ by Lemma~\ref{lem:operator-estimates}. Thus
    \[
        \xi_h / h \to (I - \pi_0) D_v f - D_v \pi_\theta f + D_v \sL_\theta u_0 =: \xi'
    \]
    pointwise and are uniformly in $\cV_{r,S}^1$. Thus by Lemma~\ref{lem:weak-compactness}, $(-\sL_0)^{-1} \xi_h / h \to (-\sL_\theta)^{-1} \xi'$.

    By the same reasoning, $B_h u_h \to B u_0$ pointwise. Putting this together yields that $w_h / h$ converges pointwise to $w'$ solving
    \[
        w' = (-\sL_\theta)^{-1} \xi' - D_v \pi_\theta u_\theta,
    \]
    which, by the above, is the unique solution of the differentiated PDE. Continuity of $w'_\theta$ in $\theta$ follows by the bounded convergence theorem as before.

    We now turn to the desired estimates. Hitting \eqref{eq:differentiated-PDE} by $(I - \pi_\theta)$, inverting $-\sL_\theta$, and subtracting the centering $(D_v \pi_\theta) u_\theta$ gives the explicit form for $D_v u_\theta$,
    \begin{equation}\label{eq:poisson-diff-1st-solution-form}
        D_v u_\theta = (-\sL_\theta)^{-1}(I - \pi_\theta)\left[ D_v f_\theta + (D_v \sL_\theta) u_\theta \right] - (D_v \pi_\theta) u_\theta.
    \end{equation}
    For any $\norm{v} \le 1$, the estimate \eqref{eq:poisson-1st-few-coords} for $D_v u_\theta$ follows from \eqref{eq:poisson-few-coords} and Lemma~\ref{lem:operator-estimates}--(6), followed by taking the supremum over $v$.
\end{proof}

\begin{lemma} \label{lem:poisson-second-deriv-pde}
    Suppose that $f \in \cW_{r,S}^3(A)$. Then the solution $u_\theta$ of \eqref{eq:general-poisson} is $C^2$ in $\theta$ on $A$ and, for any $v, w \in \RR^p$, the second directional derivative $D_w D_v u_\theta \in \cV_{r,S}^1$ solves
    \begin{equation}\label{eq:differentiated-2nd-PDE}
        \begin{cases}
            -\sL_\theta D_w D_v u_\theta = (I - \pi_\theta)\left[ D_w D_v f_\theta + (D_w D_v \sL_\theta) u_\theta + (D_v \sL_\theta)(D_w u_\theta) + (D_w \sL_\theta)(D_v u_\theta) \right], \\
            \pi_\theta D_w D_v u_\theta = -(D_w \pi_\theta)(D_v u_\theta) - (D_w D_v \pi_\theta) u_\theta - (D_v \pi_\theta)(D_w u_\theta).
        \end{cases}
    \end{equation}
    Furthermore, if $f$ depends only on coordinates $I \subseteq [p]$ and $S_I = \abs{I}^{1/2}$, then
    \begin{equation} \label{eq:poisson-2nd-few-coords}
        \norm{\nabla_\theta^2 u_\theta}_{\cV_{r,S_I}^1} \lesssim_{r} \frac{\norm{f}_{\cW_{r,S_I}^3(A)}}{(1 - \beta^r)^3} \bigl(1 + \norm{\psi}_{\cR_{r+2}^2(A)}^r\bigr)^5 \bigl(1 + \norm{\psi}_{\cR_{r+2}^2(A)}^2\bigr).
    \end{equation}
\end{lemma}
\begin{proof}
    Since $f \in \cW_{r,S}^3(A)$, by Lemma~\ref{lem:poisson-first-deriv-pde}, $D_v u_\theta \in \cV_{r,S}^2$ and solves \eqref{eq:differentiated-PDE}. From \eqref{eq:poisson-diff-1st-solution-form}, observe that the right-hand side falls in $\cV_{r,S}^2$ by Lemmas~\ref{lem:poisson-wellposed} and~\ref{lem:operator-estimates}. Apply Lemma~\ref{lem:poisson-first-deriv-pde} to \eqref{eq:poisson-diff-1st-solution-form}, hit both sides again by $(I - \pi_\theta)$, and use the fact that quantities independent of $x$ are annihilated by $\sL_\theta$, $D_w \sL_\theta$, $D_w \pi_\theta$, and $(I - \pi_\theta)$ to obtain \eqref{eq:differentiated-2nd-PDE}. Finally, the estimate \eqref{eq:poisson-2nd-few-coords} follows by inverting $-\sL_\theta$ on the $\pi_\theta$-centred version of \eqref{eq:differentiated-2nd-PDE},
    \begin{align*}
        D_w D_v u_\theta &= (-\sL_\theta)^{-1}(I - \pi_\theta)\left[ D_w D_v f_\theta + (D_w D_v \sL_\theta) u_\theta + (D_v \sL_\theta)(D_w u_\theta) + (D_w \sL_\theta)(D_v u_\theta) \right] \\
        &\qquad - (D_w \pi_\theta)(D_v u_\theta) - (D_v \pi_\theta)(D_w u_\theta) - (D_w D_v \pi_\theta) u_\theta,
    \end{align*}
    and applying \eqref{eq:poisson-few-coords} and Lemma~\ref{lem:operator-estimates}--(6).
\end{proof}

\begin{corollary} \label{cor:poisson-w3-scale-uniform}
    Suppose that $f \in \cW_{r,S_I}^3(A)$, where $f$ depends only on coordinates $I \subseteq [p]$ and $S_I = \abs{I}^{1/2}$. Then the solution $u$ of \eqref{eq:general-poisson} is $C^2$ in $\theta$ and satisfies
    \[
        \norm{u}_{\cW_{r,S_I}^3(A)} \lesssim_r \frac{\bigl(1 + \norm{\psi}_{\cR_{r+2}^2(A)}^r\bigr)^5}{(1 - \beta^r)^3} \bigl(1 + \norm{\psi}_{\cR_{r+2}^2(A)}^2\bigr) \norm{f}_{\cW_{r,S_I}^3(A)}.
    \]
\end{corollary}
\begin{proof}
    Combine \eqref{eq:poisson-few-coords}, \eqref{eq:poisson-1st-few-coords}, and \eqref{eq:poisson-2nd-few-coords}.
\end{proof}

\subsection{Applying to correctors}
We now apply the preceding results to control the correctors in the proof of Theorem~\ref{thm:main-ballistic}. Define
\begin{align*}
    F_\theta(x) := -\nabla u(\theta)[\rho(x)], \qquad Q_\theta(x) := \frac{1}{2} \nabla^2 u(\theta)[\rho(x), \rho(x)], \qquad
    B_{1,\theta}(x) := \nabla_\theta \chi_1(\theta, x)[\rho(x)],
\end{align*}
so that with $\sP_\theta := (-\sL_\theta)^{-1}(I - \pi_\theta)$, the correctors are $\chi_1 := \sP_\theta K_\theta F_\theta$, $\chi_2 := \sP_\theta K_\theta Q_\theta$, and $\xi := \sP_\theta K_\theta B_{1,\theta}$.

\begin{lemma} \label{lem:adam-correctors}
    Suppose Assumption~\ref{assume:localize} hold on $A \subset \RR^p$. Then $\chi_1 \in \cW_1^3(A)$ and $\chi_2, \xi \in \cW_2^2(A)$ are well-defined and satisfy, uniformly for $\theta \in A$ and $x \in \cX$,
    \begin{align}
        \sum\nolimits_{0 \le a \le 2} \opnorm{\nabla_\theta^a \chi_1(\theta, x)}
        &\lesssim_A \sqrt{p} + V_1(x), \label{eq:chi1-sharp-growth} \\
        \sum\nolimits_{a \in \{0, 1\}} \left( \opnorm{\nabla_\theta^a \chi_2(\theta, x)} + \opnorm{\nabla_\theta^a \xi(\theta, x)} \right)
        &\lesssim_A p + V_2(x). \label{eq:quadratic-sharp-growth}
    \end{align}
\end{lemma}
\begin{proof}
    Assumption~\ref{assume:localize}--\ref{item:A-grad} and Assumption~\ref{def:rho}--(i) imply $\norm{\psi}_{\cR_8^2(A)} \lesssim_A 1$. For each block $\alpha \in [N]$, $\rho_{I_\alpha}(x_\alpha)$ depends only on $x_\alpha \in \cX_{I_\alpha}$ and satisfies $S_\alpha^{-1}\norm{\rho_{I_\alpha}}_{\cW_{1, S_\alpha}^3(A)} = S_\alpha^{-1}\norm{\rho_{I_\alpha}}_{\cV_{1, S_\alpha}^3} \le C_\rho$. By the proof of Lemma~\ref{lem:poisson-wellposed} and Lemma~\ref{lem:operator-estimates}(6), $S_\alpha^{-1}\norm{K_\theta \rho_{I_\alpha}}_{\cW_{1, S_\alpha}^3(A)} \lesssim_A 1$. Applying Corollary~\ref{cor:poisson-w3-scale-uniform} to $a_\alpha(\theta, x_\alpha) := \sP_\theta K_\theta \rho_{I_\alpha}(x_\alpha)$ gives
    \begin{equation} \label{eq:a-alpha-w3-bound}
        \sup\nolimits_{\alpha \in [N]} S_\alpha^{-1}\norm{a_\alpha}_{\cW_{1, S_\alpha}^3(A)} \le C_A.
    \end{equation}
    Writing $\chi_1(\theta, x) = -\sum_{\alpha=1}^N \nabla_{I_\alpha} u(\theta)[a_\alpha(\theta, x_\alpha)]$ and using $S_\alpha V_1(x_\alpha / S_\alpha) = S_\alpha + \norm{x_\alpha}_2$, $\sum_{\alpha=1}^N S_\alpha^2 = p$, and $\sum_{\alpha=1}^N \norm{x_\alpha}_2^2 = \norm{x}_2^2$, we have
    \begin{equation}\label{eq:coordinate-growth}
        (\sum\nolimits_{\alpha=1}^N S_\alpha^2 V_1(x_\alpha / S_\alpha)^2)^{1/2} \lesssim \sqrt{p} + \norm{x}_2, \qquad
        \sum\nolimits_{\alpha=1}^N S_\alpha^2 V_1(x_\alpha / S_\alpha)^2 \lesssim p + V_2(x).
    \end{equation}
    Since \eqref{eq:a-alpha-w3-bound} implies $S_\alpha^b \opnorm{\nabla_\theta^a \nabla_{x_\alpha}^b a_\alpha(\theta, x_\alpha)} \le C_A S_\alpha V_1(x_\alpha / S_\alpha)$, combining \eqref{eq:a-alpha-w3-bound}, \eqref{eq:coordinate-growth}, and Assumption~\ref{assume:localize}--\ref{item:A-stat} via Cauchy--Schwarz bounds every mixed derivative $\nabla_\theta^a \nabla_x^b \chi_1$ ($0 \le a \le 2$, $0 \le b \le 3 - a$) by $C_A(\sqrt{p} + V_1(x))$, proving \eqref{eq:chi1-sharp-growth}. Since $\rho \in \cW_1^3(A)$, $\nabla_\theta \chi_1 \in \cW_1^2(A)$, and $V_1(x)^2 \le 2 V_2(x)$, we have $Q_\theta, B_{1,\theta} \in \cW_2^2(A)$, so Lemmas~\ref{lem:poisson-wellposed} and~\ref{lem:poisson-first-deriv-pde} ensure that $U_\theta := \sP_\theta K_\theta f_\theta \in \cW_2^2(A)$ for $f \in \{Q, B_1\}$ is $C^1$ in $\theta$ and satisfies \eqref{eq:differentiated-PDE}.

    Next, let $b_\alpha \in \bigl\{\rho_{I_\alpha}, a_\alpha, \nabla_\theta a_\alpha, \nabla_\theta^2 a_\alpha, S_\alpha \nabla_{x_\alpha}\rho_{I_\alpha}, S_\alpha \nabla_{x_\alpha}a_\alpha, S_\alpha \nabla_{x_\alpha}\nabla_\theta a_\alpha\bigr\}$ on the block chain $X_{n,\alpha}^x = B_\alpha^n x_\alpha + \phi_{n,\alpha}$. By \eqref{eq:a-alpha-w3-bound}, $S_\alpha^{-1}\norm{b_\alpha}_{\cV_{1, S_\alpha}^1} \lesssim_A 1$, so
    \begin{equation}\label{eq:b-alpha-lip}
        \opnorm{b_\alpha(z_\alpha) - b_\alpha(y_\alpha)} \lesssim_A V_1(y_\alpha / S_\alpha) \norm{z_\alpha - y_\alpha}_2.
    \end{equation}
    Since $\norm{\tilde{X}_{n,\alpha}}_{L^8} \lesssim_A S_\alpha$ and $\norm{X_{n,\alpha}^x - \tilde{X}_{n,\alpha}}_{L^8} \lesssim_A \beta^n S_\alpha V_1(x_\alpha / S_\alpha)$, the coupling argument of Lemma~\ref{lem:poisson-wellposed} and H\"older's inequality yield, for $1 \le q \le 4$,
    \[
        \norm{b_\alpha(X_{n,\alpha}^x) - b_\alpha(\tilde{X}_{n,\alpha})}_{L^q} \le C_A \beta^n S_\alpha V_1(x_\alpha / S_\alpha), \quad
        \norm{b_\alpha(X_{n,\alpha}^x)}_{L^q} \le C_A S_\alpha V_1(x_\alpha / S_\alpha), \quad
        \norm{b_\alpha(\tilde{X}_{n,\alpha})}_{L^q} \le C_A S_\alpha,
    \]
    with the last two bounds holding for $q \le 8$. By Minkowski's inequality and \eqref{eq:coordinate-growth},
    \[
        \norm{\rho(X_n^x) - \rho(\tilde{X}_n)}_{L^q} + \sum\nolimits_{j=1}^2 \norm{\nabla_\theta^j \chi_1(X_n^x) - \nabla_\theta^j \chi_1(\tilde{X}_n)}_{L^q}
        \le C_A \beta^n ( \sum\nolimits_{\alpha=1}^N S_\alpha^2 V_1(x_\alpha / S_\alpha)^2 )^{1/2}
        \le C_A \beta^n\bigl(\sqrt{p} + V_1(x)\bigr).
    \]
    Controlling $U \in \{\chi_2, \xi\}$ via $ab - a'b' = (a - a')b + a'(b - b')$, Cauchy--Schwarz, and \eqref{eq:coordinate-growth} gives
    \[
        \norm{B_1(X_n^x) - B_1(\tilde{X}_n)}_{L^2}
        \le C_A \beta^n \sum\nolimits_{\alpha=1}^N S_\alpha^2 V_1(x_\alpha / S_\alpha)^2 \lesssim_A \beta^n\bigl(p + V_2(x)\bigr),
    \]
    and identically for $Q$. Summing over $n \ge 0$ proves the $a = 0$ bound in \eqref{eq:quadratic-sharp-growth}.

    For $\nabla_\theta U_\theta$, \eqref{eq:poisson-diff-1st-solution-form} gives, for $\norm{v}_2 \le 1$,
    \[
        D_v U_\theta = (-\sL_\theta)^{-1}(I - \pi_\theta)\left[ K_\theta D_v f_\theta + (D_v \sL_\theta)(f_\theta + U_\theta) \right] - (D_v \pi_\theta) U_\theta.
    \]
    Since $D_v Q_\theta = \frac{1}{2}\nabla^3 u[v, \rho, \rho]$ and $D_v B_{1,\theta} = \nabla_\theta^2 \chi_1[v, \rho]$, the same bilinear bound gives $\abs{(K_\theta^n - \pi_\theta) D_v f_\theta(x)} \le C_A \beta^n(p + V_2(x))$, so $\abs{(-\sL_\theta)^{-1}(I - \pi_\theta) K_\theta D_v f_\theta(x)} \le C_A(p + V_2(x))$. For the second term, expand $\nabla_x(f_\theta + U_\theta) = \sum_{m \ge 0} \nabla_x K_\theta^m f_\theta$ and write, with $t = n + m + 1$ and $\zeta = D_v \phi_\theta(Y_{n+1})$,
    \[
        (K_\theta^n - \pi_\theta)(D_v \sL_\theta)(K_\theta^m f_\theta)(x)
        = \EE[\nabla_x f_\theta(X_t^x)[B^m \zeta] - \nabla_x f_\theta(\tilde{X}_t)[B^m \zeta]].
    \]
    On each block $\alpha \in [N]$, $\nabla_{x_\alpha}\rho_{I_\alpha}(X_{t,\alpha}^x)[B_\alpha^m \zeta_\alpha] = (S_\alpha \nabla_{x_\alpha}\rho_{I_\alpha}(X_{t,\alpha}^x))[B_\alpha^m \zeta_\alpha / S_\alpha]$ with $\norm{\zeta_\alpha / S_\alpha}_{L^4} \lesssim_A \norm{v}_2 \le 1$. H\"older's inequality, \eqref{eq:coordinate-growth}, and \eqref{eq:b-alpha-lip} then yield
    \[
        \norm{\nabla_x \rho(X_t^x)[B^m \zeta]}_{L^2} + \norm{\nabla_x \nabla_\theta \chi_1(X_t^x)[B^m \zeta]}_{L^2} \le C_A \beta^m\bigl(\sqrt{p} + V_1(x)\bigr),
    \]
    together with
    \[
        \abs{\EE\left[\nabla_x f_\theta(X_t^x)[B^m \zeta] - \nabla_x f_\theta(\tilde{X}_t)[B^m \zeta]\right]} \le C_A \beta^{m+t}\bigl(p + V_2(x)\bigr), \qquad
        \abs{\EE\left[\nabla_x f_\theta(\tilde{X}_t)[B^m \zeta]\right]} \le C_A \beta^m p.
    \]
    Summing over $n, m \ge 0$ and noting $\abs{(D_v \pi_\theta) U_\theta} \le C_A p$ proves \eqref{eq:quadratic-sharp-growth}.
\end{proof}

\section{Controlling the remainder term}

We now turn to controlling the remainder terms in \eqref{eq:error-def}, namely proving Theorem~\ref{thm:main}. The analysis in Section~\ref{sec:poisson} was carried out for a fixed $\theta \in E_{R}^{*}\cap A$. To apply those estimates along the trajectory, where $\theta_\ell$ is updated at each step and depends on the accumulated momentum, we record the following trajectory moment bounds.

\begin{lemma} \label{lem:traj-moments}
    Let Assumption~\ref{assume:localize} hold. For all $\ell \le \lfloor T / \delta \rfloor \land (\tau_R / \delta )$ and $1 \le i \le p$:
    \begin{enumerate}[label=\textup{(\alph*)},leftmargin=*]
        \item $\norm{X_{\ell,i}}_{L^{16}}\le C_R$ and $\norm{V_r(X_{\ell,i})}_{L^{8/r}}\le C_{R,r}$ for $1\le r\le8$.
        \item $\norm{D_\ell}_{L^8}\le C_R\sqrt p$, and $\norm{V_r(X_\ell)}_{L^{8/r}}\le C_{R,r}p^{r/2}$ for $1\le r\le8$.
        \item $\norm{\max_{\ell\le \lfloor T / \delta \rfloor\land\lfloor\tau_R/\delta\rfloor}V_r(X_\ell)}_{L^{8/r}}\le C_{R,T,r}p^{r/2}(\lfloor T / \delta \rfloor+1)^{r/8}$ for $1\le r\le8$.
    \end{enumerate}
\end{lemma}

\begin{proof}
    \textit{(a)} This is immediate from \eqref{eq:P} and Minkowski's inequality.

    \textit{(b)} By \eqref{eq:rho-scale-bound} and \eqref{eq:coordinate-growth}, $\norm{\rho(x)}_2^2 \le C_\rho^2 \sum_{\alpha=1}^N V_1(x_\alpha/S_\alpha)^2 \lesssim p + \norm{x}_2^2$. Combining with (a) and Minkowski's inequality, $\bigl\|\norm{X_\ell}_2^2\bigr\|_{L^4} \le \sum_{i=1}^p \bigl\|\norm{X_{\ell,i}}_2^2\bigr\|_{L^4} \le C_R p$, which gives both bounds in (b).

    \textit{(c)} This follows from the inequality $\max_{\ell \le N} | x_\ell|^{8 / r} \le \sum_{\ell \le N} | x_\ell |^{8 / r}$. Take expectations and apply (b) to obtain (c).
\end{proof}

\subsection{Proof of Theorem~\ref{thm:main}} \label{sec:proof-bounds}
From \eqref{eq:Ru-bound} and Lemma~\ref{lem:traj-moments},
\[
    |R_\ell^u|\lesssim_R\delta^3\norm{D_\ell}^3\lesssim_R\delta^3V_3(X_\ell).
\]
Furthermore, since $\chi=\chi_1+\delta\chi_2$, write $\cR^{(1)}_{\chi,\ell}=R_1+\delta R_2$, where $R_1=\Delta_\theta\chi_{1,\ell}+\delta B_u(\theta_{\ell-1},X_\ell)$ and $R_2=\Delta_\theta\chi_{2,\ell}$. By Lemmas~\ref{lem:adam-correctors} and~\ref{lem:traj-moments}, retaining the additive bounds \eqref{eq:chi1-sharp-growth}--\eqref{eq:quadratic-sharp-growth},
\begin{align*}
    |R_1|&\lesssim_R\delta^2\bigl(\sqrt p+V_1(X_\ell)\bigr)\norm{D_\ell}^2
    \lesssim_R\delta^2\bigl(p^{3/2}+V_3(X_\ell)\bigr), \\
    |R_2|&\lesssim_R\delta\bigl(p+V_2(X_\ell)\bigr)\norm{D_\ell}
    \lesssim_R\delta\bigl(p^{3/2}+V_3(X_\ell)\bigr).
\end{align*}
Here we used Young's inequality $ab \le a^3/3 + 2b^3 / 3$ to absorb $\sqrt p$ and $V_1(X_\ell)$. Similarly,
\[
    |\Delta_\theta\xi_\ell|
    \lesssim_R\delta\bigl(p+V_2(X_\ell)\bigr)\norm{D_\ell}
    \lesssim_R\delta\bigl(p^{3/2}+V_3(X_\ell)\bigr).
\]
Since $\cR^{(1)}_{\chi,\ell}=R_1+\delta R_2$, summing over at most $\lfloor T / \delta \rfloor$ steps gives
\begin{align*}
    \EE\sup\nolimits_{r\le \lfloor T / \delta \rfloor}|\sum\nolimits_{\ell=1}^{r}
    \bigl(R_\ell^u+\delta\cR^{(1)}_{\chi,\ell}-\delta^2\Delta_\theta\xi_\ell\bigr)|
    &\le C_R\delta^3\sum\nolimits_{\ell\le \lfloor T / \delta \rfloor}
    \EE\bigl[p^{3/2}+V_3(X_\ell)\bigr] \le C_{R,T}\delta^{1/2}.
\end{align*}
Finally, for the endpoints, $\chi=\chi_1+\delta\chi_2$ and \eqref{eq:chi1-sharp-growth}--\eqref{eq:quadratic-sharp-growth} give
\[
    \delta|\chi_\ell|+\delta^2|\xi_\ell|
    \lesssim_R\delta\bigl(\sqrt p+V_1(X_\ell)\bigr)
    +\delta^2\bigl(p+V_2(X_\ell)\bigr).
\]
Taking the maximum over the stopped trajectory and applying Lemma~\ref{lem:traj-moments}(c),
\[
    \EE\max_{\ell\le \lfloor T / \delta \rfloor\wedge\lfloor\tau_R/\delta\rfloor}
    \bigl(\delta|\chi_\ell|+\delta^2|\xi_\ell|\bigr)
    \lesssim_{R,T}\delta\sqrt p\,(\lfloor T / \delta \rfloor+1)^{1/8}
    +\delta^2p\,(\lfloor T / \delta \rfloor+1)^{1/4}
    \lesssim_{R,T}\delta^{3/8}+\delta^{3/4}. \qed
\]

\section{Controlling the martingale} \label{sec:martingale}

We now turn to controlling the martingale term. To propagate our fixed-$\theta$ Poisson estimates along the trajectory, we exploit the fast mixing of the exponential moving average by coupling $X_\ell$ over a logarithmic window of length $\Lambda = \max\{1, \lceil C\log(1/\delta)\rceil\}$ (with $C = -2/\log\beta$ when $0 < \beta < 1$, so $\beta^\Lambda \le \delta^2$, and $\Lambda = 1$ if $\beta = 0$) to a stationary chain fixed-parameter at $\ell_0 = (\ell - \Lambda) \vee 0$. Conditional on $\cF_{\ell_0}$, sample $X_{\ell_0}^{\mathrm{fr}} \sim \pi_{\theta_{\ell_0}}$ independently of $(Y_j)_{j>\ell_0}$, and evolve $(X_j^{\mathrm{fr}})_{j=\ell_0}^\ell$ with fixed parameter $\theta_{\ell_0}$ using the \textit{same} data stream $Y_{\ell_0+1}, \ldots, Y_\ell$.

\begin{lemma} \label{lem:window}
    Let Assumption~\ref{assume:localize} hold and $\delta=O(1/p)$. For any $\ell\le\lfloor\tau_R/\delta\rfloor\wedge \lfloor T / \delta \rfloor$:
    \begin{enumerate}[label=\textup{(\alph*)},leftmargin=*]
        \item For every $\ell_0+1\le \ell'\le\ell$ and coordinate $i\in[p]$,
        \[
            \norm{\theta_{\ell'-1}-\theta_{\ell_0}}_{L^8}\le C_R\delta\Lambda\sqrt p,
            \qquad
            \norm{X_{\ell,i}-X_{\ell,i}^{\mathrm{fr}}}_{L^8}
            \le C_R\bigl(\beta^{\ell-\ell_0}+\delta\Lambda\sqrt p\bigr).
        \]
        \item Writing $F_\theta(x)=-\nabla u(\theta)\cdot\rho(x)$,
        \[
            \norm{F_{\theta_{\ell-1}}(X_\ell)-F_{\theta_{\ell_0}}(X_\ell^{\mathrm{fr}})}_{L^4}
            +\norm{\chi_1(\theta_{\ell-1},X_\ell)-\chi_1(\theta_{\ell_0},X_\ell^{\mathrm{fr}})}_{L^4} \le C_R\bigl(\Lambda+\sqrt p\,\beta^\ell\bigr).
        \]
        \item $\norm{F_{\theta_{\ell-1}}(X_\ell)}_{L^4}+\norm{\chi_1(\theta_{\ell-1},X_\ell)}_{L^4}\le C_R\bigl(\Lambda+\sqrt p\,\beta^\ell\bigr)$.
    \end{enumerate}
\end{lemma}

\begin{proof}
    \textit{(a)} Since $\theta_{\ell'-1}-\theta_{\ell_0}=-\delta\sum_{s=\ell_0+1}^{\ell'-1}D_s$, the first bound follows from Minkowski's inequality and Lemma~\ref{lem:traj-moments}(b). Subtracting the momentum recursions gives
    \[
        X_{\ell,i}-X_{\ell,i}^{\mathrm{fr}}
        = B_i^{\ell-\ell_0}(X_{\ell_0,i}-X_{\ell_0,i}^{\mathrm{fr}})
        +(I-B_i)\sum_{\ell'=\ell_0+1}^{\ell} B_i^{\ell-\ell'}
        \bigl(\psi_i(\theta_{\ell'-1},Y_{\ell'})-\psi_i(\theta_{\ell_0},Y_{\ell'})\bigr).
    \]
    By Lemma~\ref{lem:traj-moments}(a), the initial term has $L^8$ norm bounded by $C_R\beta^{\ell-\ell_0}$. By \eqref{eq:P} and the mean-value theorem, $\norm{\psi_i(\theta_{\ell'-1},Y_{\ell'})-\psi_i(\theta_{\ell_0},Y_{\ell'})}_{L^8} \le C_R\norm{\theta_{\ell'-1}-\theta_{\ell_0}}_{L^8}$, and summing the geometric series yields (a).

    \textit{(b)} For $b \in \{\rho, a\}$, write $\Phi(\theta,x) = -\sum_{\alpha=1}^N \nabla_{I_\alpha} u(\theta)[b_\alpha(\theta,x_\alpha)]$ and decompose
    \[
        \Phi(\theta_{\ell-1},X_\ell)-\Phi(\theta_{\ell_0},X_\ell^{\mathrm{fr}})
        = \bigl[\Phi(\theta_{\ell-1},X_\ell)-\Phi(\theta_{\ell_0},X_\ell)\bigr] + \bigl[\Phi(\theta_{\ell_0},X_\ell)-\Phi(\theta_{\ell_0},X_\ell^{\mathrm{fr}})\bigr].
    \]
    By (a), Cauchy--Schwarz, Lemma~\ref{lem:traj-moments}(b), and \eqref{eq:chi1-sharp-growth}, the first term is bounded in $L^4$ by $C_R\delta\Lambda p \le C_R\Lambda$. For the second term, summing (a) over $i \in I_\alpha$ gives $\norm{X_{\ell,\alpha}-X_{\ell,\alpha}^{\mathrm{fr}}}_{L^8} \le C_R S_\alpha(\beta^{\ell-\ell_0}+\delta\Lambda\sqrt{p})$. Since $\norm{V_1(X_{\ell,\alpha}^{\mathrm{fr}} / S_\alpha)}_{L^8} \le C_R$, \eqref{eq:b-alpha-lip} and H\"older's inequality yield $\norm{b_\alpha(\theta_{\ell_0},X_{\ell,\alpha})-b_\alpha(\theta_{\ell_0},X_{\ell,\alpha}^{\mathrm{fr}})}_{L^4} \le C_R S_\alpha(\beta^{\ell-\ell_0}+\delta\Lambda\sqrt{p})$. Applying Cauchy--Schwarz across blocks $\alpha \in [N]$ with $\sum_{\alpha=1}^N S_\alpha^2 = p$ and Assumption~\ref{assume:localize}--\ref{item:A-stat} gives
    \[
        \Bigl\|\sum_{\alpha=1}^N \nabla_{I_\alpha} u(\theta_{\ell_0})
        \bigl[b_\alpha(\theta_{\ell_0},X_{\ell,\alpha})-b_\alpha(\theta_{\ell_0},X_{\ell,\alpha}^{\mathrm{fr}})\bigr]\Bigr\|_{L^4} \le C_R\sqrt{p}\bigl(\beta^{\ell-\ell_0}+\delta\Lambda\sqrt p\bigr).
    \]
    Since $\ell-\ell_0=\ell$ for $\ell<\Lambda$ and $\sqrt p\,\beta^{\ell-\ell_0}=\sqrt p\,\beta^\Lambda\le\sqrt p\,\delta^2\lesssim\delta^{3/2}$ for $\ell\ge\Lambda$, (b) follows.

    \textit{(c)} Assumption~\ref{assume:localize}--\ref{assume:A-proj} gives $\norm{F_{\theta_{\ell_0}}(X_\ell^{\mathrm{fr}})}_{L^4}\le C_R$. Expanding $\chi_{1,\theta}=\sum_{j=1}^\infty(K_\theta^j F_\theta-\pi_\theta F_\theta)$, contraction of $K_\theta$ on $L^4(\pi_\theta)$ gives $\norm{K_\theta^j F_\theta-\pi_\theta F_\theta}_{L^4(\pi_\theta)}\le 2C_R$ for $1\le j\le\Lambda$, while for $j>\Lambda$ the coupling bound in the proof of Lemma~\ref{lem:adam-correctors} gives $\norm{K_\theta^j F_\theta-\pi_\theta F_\theta}_{L^4(\pi_\theta)}\le C_R\sqrt p\,\beta^j$. Summing over $j\ge 1$ bounds $\norm{\chi_1(\theta_{\ell_0},X_\ell^{\mathrm{fr}})}_{L^4} \le C_R(\Lambda + \sqrt{p}\beta^\Lambda) \le C_R\Lambda$, and combining with (b) yields (c).
\end{proof}

\subsection{Proof of Lemma~\ref{lem:mart-increment}}

Fix $\ell\le \lfloor T / \delta \rfloor\land(\tau_R/\delta)$ and set $r_*=8/3$. With $F_\theta, Q_\theta$ as in Section~\ref{sec:poisson}, define $Z_\ell^u := \bigl(F_\theta + \delta Q_\theta + \chi - \delta B_u - \delta\xi\bigr)(\theta_{\ell-1},X_\ell)$, so that $\Delta\cM_\ell^u = Z_\ell^u - \EE[Z_\ell^u\mid\cF_{\ell-1}]$ by \eqref{eq:total-mart}. Conditional Jensen and Minkowski's inequality give
\[
    \norm{\Delta\cM_\ell^u}_{L^{r_*}}\le 2\norm{Z_\ell^u}_{L^{r_*}}\le 2\Bigl(\norm{F_{\theta_{\ell-1}}(X_\ell)}_{L^{r_*}}+\delta\norm{Q_{\theta_{\ell-1}}(X_\ell)}_{L^{r_*}}+\norm{\chi}_{L^{r_*}}+\delta\norm{B_u}_{L^{r_*}}+\delta\norm{\xi}_{L^{r_*}}\Bigr).
\]
By Lemma~\ref{lem:window}(c), Assumption~\ref{assume:localize}\ref{item:A-stat}, and Lemma~\ref{lem:traj-moments}(b),
\[
    \norm{F_{\theta_{\ell-1}}(X_\ell)}_{L^{r_*}}+\delta\norm{Q_{\theta_{\ell-1}}(X_\ell)}_{L^{r_*}}
    \le\norm{F_{\theta_{\ell-1}}(X_\ell)}_{L^4}+C_R\delta\,\norm{D_\ell}_{L^{16/3}}^2\le C_R\bigl(\Lambda+\sqrt p\,\beta^\ell\bigr).
\]
Similarly, by Lemma~\ref{lem:window}(c), Lemma~\ref{lem:adam-correctors}, and Lemma~\ref{lem:traj-moments}(b),
\[
    \norm{\chi}_{L^{r_*}} \le \norm{\chi_1}_{L^4} + \delta \norm{\chi_2}_{L^{8 / 3}} \le C_R\bigl(\Lambda+\sqrt p\,\beta^\ell\bigr)+C_R\delta\bigl(p+\norm{V_2(X_\ell)}_{L^{8/3}}\bigr)\le C_R\bigl(\Lambda+\sqrt p\,\beta^\ell\bigr).
\]
For the remaining two terms, \eqref{eq:chi1-sharp-growth}--\eqref{eq:quadratic-sharp-growth} and Lemma~\ref{lem:traj-moments}(b) give
\[
    \delta\norm{B_u}_{L^{8/3}}+\delta\norm{\xi}_{L^{8/3}}
    \le C_R\delta\bigl(p+\norm{V_2(X_\ell)}_{L^{8/3}}\bigr)\le C_R.
\]
Summing these bounds and noting $\Lambda \lesssim \log(1/\delta)$ proves \eqref{eq:mart-increment}.\qed

\section{Spiked Tensor and Matrix PCA}
\label{sec:ex-tensor-pca}

In this section, we provide the effective dynamics of the Spiked Tensor PCA model presented in Example~\ref{ex:app-tensor-pca}. In the following, fix $k \ge 2$, $\lambda > 0$, and step size $\delta = c_\delta / d$ where we track the statistics $m = \langle \theta , \theta_{*} \rangle$ and $R^2 = \| \theta \|^2$.

For a two-sided i.i.d.\ standard Gaussian sequence $(Z_a)_{a\in\ZZ}$, set $h \in \ZZ$ and define,
\[
    P_h=(1-\beta_1)\sum\nolimits_{a=0}^\infty\beta_1^aZ_{h-a},\qquad
    Q_h=(1-\beta_2)\sum\nolimits_{a=0}^\infty\beta_2^aZ_{h-a}^2.
\]
The effective coefficients are
\begin{align}
    \alpha(R^2)
    &=(1-\beta_1)\sum\nolimits_{j=0}^\infty\beta_1^j
    \EE[Z_{-j}^2/\sqrt{4kR^{2k-2}Q_0+\epsilon}],
    \label{eq:pca-alpha-def}\\
    \cD(R^2)
    &=\Gamma_0(R^2)+2\sum\nolimits_{h=1}^\infty\Gamma_h(R^2),\qquad
    \Gamma_h(R^2)=\EE[\widetilde D_0(R^2)\widetilde D_h(R^2)],
    \label{eq:pca-D-def}
\end{align}
where $\widetilde D_h(R^2)=\sqrt{4kR^{2k-2}}P_h/\sqrt{4kR^{2k-2}Q_h+\epsilon}$. Furthermore, for $\beta_2 \in (0, 1)$, define the constants $\kappa_1, \kappa_2$ by,
\begin{align}
\kappa_1=\kappa_1(\beta_1,\beta_2)
    &:=(1-\beta_1)\sum\nolimits_{j=0}^\infty\beta_1^j
    \EE[Z_{-j}^2/\sqrt{Q_0}],\label{eq:pca-kappa-one}\\
    \kappa_2=\kappa_2(\beta_1,\beta_2)
    &:=\EE[P_0^2/Q_0]
    +2\sum\nolimits_{h=1}^\infty
    \EE[P_0P_h/\sqrt{Q_0Q_h}].
    \label{eq:pca-kappa-two}
\end{align}

\begin{proposition}[ADAM dynamics for Spiked Tensor PCA]\label{prop:adam-pca-dynamics}
    Fix $k\ge2$, $\lambda,c_\delta>0$, $\beta_1,\beta_2\in[0,1)$, and $\epsilon>0$. Suppose that, for some $\zeta\in[0,1/2)$, $\theta_0,\theta_*\in\sD_\zeta$, and that $\uv(\theta_0)\Rightarrow\nu$ as $d\to\infty$. Then, for every $T>0$, the statistics $\uv_n$ converge to the solution $\uv_t=(m_t,R_t^2)$ of
    \begin{equation}\label{eq:adam-ode-pca}
            \dd m=2k\alpha(R^2)\bigl(\lambda m^{k-1}-R^{2k-2}m\bigr)\,\dd t,\qquad
            \dd R^2=\bigl[4k\alpha(R^2)(\lambda m^k-R^{2k})
            +c_\delta\cD(R^2)\bigr]\,\dd t,
    \end{equation}
    initialized from $\uv_0\sim\nu$.
\end{proposition}

The lag-zero variance in \eqref{eq:pca-D-def} arises from $\delta\widetilde\cL_\theta R^2$, whereas the autocovariance series is obtained from $-\delta\widetilde\cA_\theta\chi_{1,\theta}^{R^2}$. Sending $\epsilon\downarrow0$ provides a cleaner reduction of the limiting dynamics. In particular, we first take the $d \to \infty$ limit for fixed $\epsilon > 0$, followed by $\epsilon \downarrow 0$ in the limiting ODE.

\begin{corollary} \label{cor:adam-pca-eps-zero}
    As $\epsilon\downarrow0$, $\alpha(R^2)\to\kappa_1/(2\sqrt{k}R^{k-1})$ and $\cD(R^2)\to\kappa_2$, from which $\uv_n$ converge to the solution of \eqref{eq:adam-ode-pca} with $\alpha (R^2) = \kappa_1 / ( 2 \sqrt{k} R^{k-1} )$ and $\cD (R^2 ) = \kappa_2$.
\end{corollary}

\subsection{Comparison with other first-order algorithms}
\label{subsec:pca-algo-comparison}

All three algorithms considered below have the form \eqref{eq:adam-ode-pca}, with different choices of the two scalar coefficients.

\begin{proposition} \label{prop:pca-sgdm-sgdu}
    \label{cor:pca-ra-sgdm-sgdu}
    Under the setting off Proposition~\ref{prop:adam-pca-dynamics}, the limiting dynamics of SGD and SGD-M are given by \eqref{eq:adam-ode-pca} with the coefficients in Table~\ref{tab:pca-effective-coefficients}. The ADAM row has the order of limits and momentum assumptions of Corollary~\ref{cor:adam-pca-eps-zero}.
\end{proposition}

\begin{table}[t]
    \centering
    \begin{tabular}{lcc}
        \toprule
        Algorithm & $\alpha(R^2)$ & $\cD( R^2 )$ \\
        \midrule
        SGD & $1$ & $4kR^{2k-2}$\\[1mm]
        SGD-M & $(1-\beta)^{-1}$ & $4kR^{2k-2}(1-\beta)^{-2}$\\[1mm]
        ADAM ($\epsilon\downarrow0$)
        & $\displaystyle\frac{\kappa_1}{2\sqrt{k}R^{k-1}}$ & $\kappa_2$\\[2mm]
        \bottomrule
    \end{tabular}
    \caption{Effective coefficients at the critical scaling $\delta=c_\delta/d$.}
    \label{tab:pca-effective-coefficients}
\end{table}

To isolate whether an algorithm is learning the direction of $\theta_*$, write $a = m / R$ and $y = R^2 - m^2$. One change of variables suffices for all the rows of Table~\ref{tab:pca-effective-coefficients}:
\begin{equation}\label{eq:pca-common-coordinate-transform}
    \begin{aligned}
        \dot R=2k\alpha(R^2)
        (\lambda R^{k-1}a^k-R^{2k-1})
        +\frac{c_\delta\cD(R^2)}{2R},\qquad
        \dot a=2k\lambda\alpha(R^2)R^{k-2}a^{k-1}(1-a^2)
        -\frac{c_\delta\cD(R^2)}{2R^2}a.
    \end{aligned}
\end{equation}
In particular, for zero-$\epsilon$ ADAM, rescaling time by $\sqrt{k} \kappa_1$ leaves the momentum dependence only through $\zeta_{c_\delta} = c_\delta\kappa_2 / (2\sqrt{k}\kappa_1)$.

The following identifies the $t \to \infty$ behaviour of the limiting ODEs for Matrix PCA ($k = 2$).

\begin{proposition} \label{prop:adam-pca-global}
    \label{prop:adam-pca-equilibria}
    Fix $k=2$, $\lambda,c_\delta>0$, $\beta_1\in[0,1)$, $\beta_2\in(0,1)$, and let $\zeta_{c_\delta}$ be defined as above. The fixed points for the the zero-$\epsilon$ ADAM dynamics are as follows.
    \begin{enumerate}[label=\textup{(\roman*)},leftmargin=*]
        \item If $\zeta_{c_\delta}<\lambda^{3/2}$ and $m_0\ne0$, then $(R_{\mathrm{eq}}, a_{\mathrm{eq}}) = \bigl(\sqrt{\lambda},\,  \pm \sqrt{1 - \zeta_{c_\delta}/ \lambda^{3/2}} \bigr)$ are two stable fixed points of the system. If $m_0=0$, then $(R_{\mathrm{eq}}, a_{\mathrm{eq}}) = \bigl(\zeta_{c_\delta}^{1/3},\, 0\bigr)$ is an unstable fixed point.
        \item If $\zeta_{c_\delta}\ge\lambda^{3/2}$, then $(R_{\mathrm{eq}}, a_{\mathrm{eq}}) = \bigl(\zeta_{c_\delta}^{1/3},\, 0\bigr)$ is an unstable fixed point.
        \item Under online SGD, if $m_0\ne0$, then the stable fixed points are given by $(R_{\mathrm{eq}}, a_{\mathrm{eq}}) = (\sqrt{\lambda}, \pm \sqrt{1-c_\delta/\lambda})$ for $c_\delta<\lambda$ and $(c_\delta,0)$ otherwise. Under unnormalized SGD-M, the same statements hold with $c_\delta/(1-\beta)$ in place of $c_\delta$.
    \end{enumerate}
\end{proposition}

Next, observe that the non-constant part of the population risk is given by,
\begin{equation}\label{eq:pca-population-risk}
    \cR(\theta):=R^{2k}-2\lambda m^k
    =\EE_Y\bigl[\Fnorm{Y-\theta^{\otimes k}}^2\bigr]
    -\EE_Y\bigl[\Fnorm{Y}^2\bigr].
\end{equation}

\begin{corollary}[Population risk dynamics]\label{cor:pca-population-risk}
    Along \eqref{eq:adam-ode-pca}, with any row of Table~\ref{tab:pca-effective-coefficients},
    \begin{equation}\label{eq:pca-exact-risk-ode-adam}
        \dot\cR(m_t,R_t^2)
        =-4k^2\alpha(R_t^2)\cB_k(m_t,R_t)
        +c_\delta kR_t^{2k-2}\cD(R_t^2),
    \end{equation}
    where $\cB_k(m,R) = \norm{R^{2k-2}\theta-\lambda m^{k-1}\theta_*}^2$.
\end{corollary}

\subsection{Proof of the effective dynamics}
\label{sec:proofs-tensor-pca}

Fix $\zeta<\zeta_+<1/2$ and $A_d=\sD_{\zeta_+}$. We verify the assumptions of Theorem~\ref{thm:main-ballistic} on $E_L = \{ \theta : \| \theta \| < L\}$. Throughout this proof, constants may depend on $L, k,\lambda,\beta_1,\beta_2$, and $\epsilon$, but not on $d$.

\subsubsection{Law of the gradient}

Set $F(\theta)=\langle W,\theta^{\otimes k}\rangle$. Expanding the loss gives $L(\theta,Y)=R^{2k}-2\lambda m^k-2F(\theta)+\Fnorm Y^2$, and hence
\begin{equation}\label{eq:gradient}
    g(\theta,Y)=b(\theta)-2\nabla F(\theta),\qquad
    b(\theta)=A(R^2)\theta-H(m)\theta_*,
\end{equation}
where the scalar coefficients are given by $A(R^2)=2kR^{2k-2}$, $H(m)=2k\lambda m^{k-1}$, $\mathsf s(R^2)=4kR^{2k-2}$, and $\mathsf u(R^2)=4k(k-1)R^{2k-4}$. Differentiating the covariance $\EE[F(\theta)F(\vartheta)]=(\theta\cdot\vartheta)^k$ shows that
\begin{equation}\label{eq:gradient-law}
    g(\theta,Y)\sim\cN(b(\theta),\Sigma(\theta)),\qquad
    \Sigma(\theta)=\mathsf s(R^2)I_d+\mathsf u(R^2)\theta\theta^\top,
    \qquad q_i=\mathsf s(R^2)+\mathsf u(R^2)\theta_i^2.
\end{equation}
Thus the fixed-parameter gradients can be represented as
\begin{equation}\label{eq:indep-common}
    G_{\ell,i}=b_i+\sqrt{\mathsf s(R^2)}Z_{\ell,i}
    +\sqrt{\mathsf u(R^2)}\theta_iZ_{\ell,0},
\end{equation}
where $(Z_{\ell,i})_{\ell\in\ZZ,\,0\le i\le d}$ are independent standard Gaussian variables.

\subsubsection{Moment bounds and delocalization} \label{sec:pca-moment-deloc}

The statistic derivatives are $\nabla m=\theta_*$, $\nabla^2m=0$, $\nabla R^2=2\theta$, $\nabla^2R^2=2I_d$, and $\nabla^3R^2=0$n and satisfy Assumption~\ref{assume:localize}, Item~\ref{item:A-stat}. For every finite moment order $r\ge2$ and $j=1,2,3$, the tensor derivative bound
\[
    \Fnorm{\nabla^j(\theta^{\otimes k})[e_i,h_1,\ldots,h_{j-1}]}
    \le C_k\norm\theta^{k-j}\prod\nolimits_{a=1}^{j-1}\norm{h_a}
    \qquad (j\le k)
\]
and the Gaussian law of $\langle W,T\rangle$ give uniform $L^r$ bounds for $\partial_i\nabla^{j-1}F[h_1,\ldots,h_{j-1}]$ on $E_R^{*}$. The deterministic polynomial $b$ has the same local derivative bounds. In particular,
\begin{equation}\label{eq:pca-all-gradient-moments}
    \sup\nolimits_{\theta \in E_{L}}\max\nolimits_{i\le d}
    (\norm{g_i}_{L^r}
    +\sup\nolimits_{\norm h\le1}\norm{\nabla g_i[h]}_{L^r}
    +\sup\nolimits_{\norm h,\norm v\le1}\norm{\nabla^2g_i[h,v]}_{L^r})
    \le C_{L,r}.
\end{equation}
Taking $r=16$ verifies Item~\ref{item:A-grad}. We next establish the stronger stationary projection bound
\begin{equation}\label{eq:pca-strong-proj}
    \sup\nolimits_{E_L}\norm{v\cdot\rho}_{L^r(\pi_\theta)}
    \le C_{L,r}\norm v,\qquad r\ge2,
\end{equation}
which implies Item~\ref{assume:A-proj}. Couple the stationary coordinate $D_i=\rho_i(\widetilde X_{0,i})$ to $\widehat D_i=\rho_i(\widehat X_{0,i})$ using the comparison gradients $\widehat G_{\ell,i}=\sqrt{\mathsf s(R^2)}Z_{\ell,i}$ in \eqref{eq:indep-common}. Then $\norm{G_{\ell,i}-\widehat G_{\ell,i}}_{L^{2r}} \le C_{L,r}(|b_i|+|\theta_i|)$. Unrolling the two moving averages gives the same bound for the difference of the first moments and, by H\"older's inequality, for the difference of the second moments. Furthermore,
\begin{equation}\label{eq:Di-Dhat-diff}
        \norm{D_i-\widehat D_i}_{L^r}
        \le\epsilon^{-1/2}\norm{\widetilde M_{0,i}-\widehat M_{0,i}}_{L^r}
        +\norm{\widehat M_{0,i}}_{L^{2r}}/(2\epsilon^{3/2})
        \norm{\widetilde V_{0,i}-\widehat V_{0,i}}_{L^{2r}} \lesssim_{R, r} |b_i|+|\theta_i|.
\end{equation}
The variables $(\widehat D_i)_{i\le d}$ are independent, centered, and have uniformly bounded moments of every finite order. The moment inequality for sums of independent centered variables and Cauchy--Schwarz therefore yield
\[
    \norm{v\cdot\rho}_{L^r(\pi_\theta)}
    \le C_{L,r}\norm v+C_{L,r}\sum|v_i|(|b_i|+|\theta_i|)
    \le C_{L,r}\norm v\,(1+\norm b+\norm\theta)
    \le C_{L,r}\norm v.
\]
To verify that the parameter coordinates remain delocalized over training, we follow the proof of Theorem~\ref{thm:deloc} and provide the model-specific estimates that change. The first estimate needed is \eqref{eq:drift-U}, which follows from the same coupling argument for \ref{eq:Di-Dhat-diff}, from which we have the coordinate drift estimate $|\pi_\theta\rho_i| \le C_L(|b_i|+|\theta_i|) \le C_L (\norm\theta_\infty+\norm{\theta_*}_\infty )$. The remaining estimates used in the proof of Theorem~\ref{thm:deloc} follow from \eqref{eq:pca-all-gradient-moments}, in particular we require the moment bounds,
\[
    \sup\nolimits_{\ell,i}\norm{1+\norm{X_{\ell,i}}}_{L^r}\le C_{L,r},\qquad
    \sup\nolimits_{\ell,i}\norm{D_{\ell,i}}_{L^r}\le C_{L,r},\qquad
    \sup\nolimits_\ell\norm{\norm{D_\ell}}_{L^r}\le C_{L,r}\sqrt d
\]
for every finite $r$. The decomposition~\eqref{eq:theta-decomp} remains the same, and again following the same arguments in Theorem~\ref{thm:deloc} using the PCA moment bounds in place of those used from Lemma~\ref{lem:moments-gauss} proves the same upper bounds in \eqref{eq:three-error-bounds}. The Gr\"onwall argument given in \eqref{eq:gronwall} remains the same, after which the remainder of the proof establishes the desired high-probability event \eqref{eq:thmC-prop} for the Tensor PCA data model. Consequently, since $\uv$ includes $R^2$, Corollary~\ref{cor:deloc-interface} applies, and $\PP\bigl(\tau_{A_d}\le\tau_{L}\wedge\lfloor T/\delta\rfloor\bigr) \to 0$ for every fixed $L$ and $T$, thereby verifying Assumption~\ref{assume:A-subset-timescale} for the spiked Tensor PCA data model.

\subsubsection{The stationary mean and effective drift}

By \eqref{eq:gradient-law}, the marginal fixed-parameter law of $g_i$ depends on $\theta$ only through $(b_i,q_i)$. With $P_0,Q_0$ as in Proposition~\ref{prop:adam-pca-dynamics} and $T_0=(1-\beta_2)\sum_{a\ge0}\beta_2^aZ_{-a}$, write
\begin{equation}\label{eq:f-bq-def}
    \pi_\theta\rho_i=f(b_i,q_i),\qquad
    f(b,q)=\EE\bigl[(b+\sqrt qP_0)V(b,q)^{-1/2}\bigr],\qquad
    V(b,q)=b^2+2b\sqrt qT_0+qQ_0+\epsilon.
\end{equation}

\begin{lemma}\label{lem:f-bq-expansion}
    Uniformly on $E_{L} \cap \sD_{\zeta_+}$,
    \begin{equation}\label{eq:pca-mean-expansion}
        \pi_\theta\rho=\alpha(R^2)b(\theta)+\cE(\theta),\qquad
        \norm{\cE(\theta)}\le C_Ld^{-1+2\zeta_+}.
    \end{equation}
\end{lemma}

\begin{proof}
    The map $b\mapsto f(b,q)$ is odd, so $f(0,q)=\partial_b^2f(0,q)=0$. Differentiating \eqref{eq:f-bq-def} at $b=0$ gives
    \[
        \partial_bf(0,q)=\EE[(qQ_0+\epsilon)^{-1/2}]
        -q\EE[P_0T_0(qQ_0+\epsilon)^{-3/2}].
    \]
    In the expansion of $P_0T_0$, off-diagonal terms vanish because the denominator is even in every Gaussian coordinate. Gaussian integration by parts then gives
    \begin{align*}
        \partial_bf(0,q)
        &=(1-\beta_1)\sum_{a\ge0}\beta_1^a
        \EE\left[\frac1{\sqrt{qQ_0+\epsilon}}
        -\frac{q(1-\beta_2)\beta_2^aZ_{-a}^2}
        {(qQ_0+\epsilon)^{3/2}}\right] =(1-\beta_1)\sum_{a\ge0}\beta_1^a
        \EE\left[\frac{Z_{-a}^2}{\sqrt{qQ_0+\epsilon}}\right].
    \end{align*}
    At $q=\mathsf s(R^2)$ this is \eqref{eq:pca-alpha-def}. Since $\EE[Z_{-a}^2Q_0]\le3$, $|\partial_q\partial_bf(0,q)|\le3/(2\epsilon^{3/2})$. Three derivatives in $b$ of \eqref{eq:f-bq-def} are bounded on $E_L$, because $V\ge\epsilon$ and the Gaussian sums have all moments. Taylor's theorem therefore yields the coordinate estimate
    \[
        |f(b_i,q_i)-\alpha(R^2)b_i|
        \le C_L\bigl(|b_i|\theta_i^2+|b_i|^3\bigr).
    \]
    Taking the Euclidean norm, proves $\norm{\cE(\theta)} \le C_L(\norm\theta_\infty^2+\norm b_\infty^2)\norm b \le C_Ld^{-1+2\zeta_+}$.
\end{proof}

We now assemble the effective generator in Assumption~\ref{assume:closable}. By Lemma~\ref{lem:adam-correctors}, write $\chi_{1,\theta}^u(x)=-\nabla u(\theta)\cdot\av(\theta,x)$, where $\av_i=\sP_\theta K_\theta\rho_i$ is the coordinate Poisson corrector. The product rule gives
\begin{equation}\label{eq:psi-prelimit-pca}
    \Psi_n^u
    =-\nabla u\cdot\pi_\theta\rho
    +\delta\pi_\theta\left[\frac12\nabla^2u[\rho,\rho]
    +\nabla^2u[\rho,\av]\right]
    +\delta T_{\nabla u},\qquad
    T_v=\pi_\theta[v\cdot\nabla_\theta\av[\rho]].
\end{equation}
For the derivative term, write its corrector as $\mathfrak a(b_i,\sigma_i;x_i)$ with $\sigma_i=\sqrt{q_i}$. The Poisson estimates of Corollary~\ref{cor:poisson-w3-scale-uniform}, together with \eqref{eq:pca-all-gradient-moments}, bound its first $(b,\sigma)$ derivatives in $L^4$ uniformly on bounded parameter sets. The polynomial formula for $b_i$ and \eqref{eq:pca-strong-proj} imply $\norm{\nabla b_i[\rho]}_{L^4(\pi_\theta)}\le C_L$. Next, putting $H_i=(R^2+(k-1)\theta_i^2)^{1/2}$ gives,
\[
    \sigma_i=2\sqrt{k}R^{k-2}H_i,\qquad
    \nabla\sigma_i[\rho]
    =2\sqrt{k}[(k-2)R^{k-4}H_i\langle\theta,\rho\rangle
    +(R^{k-2}/H_i)
    (\langle\theta,\rho\rangle+(k-1)\theta_i\rho_i)].
\]
The first term vanishes for $k=2$. Since $R\le H_i\le\sqrt{k}R$, $\norm{\langle\theta,\rho\rangle}_{L^4}\le C_LR$, and $|\theta_i|\le R$, this also gives $\norm{\nabla\sigma_i[\rho]}_{L^4}\le C_L$. H\"older's inequality and the chain rule applied to $\nabla_\theta a [\rho ] = \partial_b \mathfrak{a} ( \nabla_\theta b_i [\rho ] ) + \partial_\sigma \mathfrak{a} ( \nabla_\theta \sigma_i [\rho ] )$ yield,
\begin{equation}\label{eq:Tv-pca-bound}
    |T_v|\le C_L\sum\nolimits_{i=1}^d|v_i|\le C_L\sqrt d\norm v,
    \qquad
    \delta\bigl(|T_{\theta_*}|+|T_{2\theta}|\bigr)\le C_Ld^{-1/2}.
\end{equation}

For the quadratic term, set $J(b,q)=\pi_{b,q}(\rho_i^2+2\rho_i\mathfrak a(b,\sqrt q;\cdot))$. Then
\[
    \delta\pi_\theta[\norm\rho^2+2\rho\cdot\av]
    =(c_\delta/d)\sum\nolimits_{i=1}^d J(b_i,q_i),\qquad
    J(0,\mathsf s(R^2))=\cD(R^2).
\]
The second identity follows by expanding $\mathfrak a(0,\sqrt{\mathsf s};\cdot) =\sum_{h\ge1}K_{0,\mathsf s}^{h}\rho_i$ under its stationary law. Corollary~\ref{cor:poisson-w3-scale-uniform} and Lemma~\ref{lem:operator-estimates}(6) give a locally bounded first derivative of $(b,\sigma)\mapsto J(b,\sigma^2)$. Since $|\sqrt{q_i}-\sqrt{\mathsf s(R^2)}| \le\sqrt{\mathsf u(R^2)}|\theta_i|$, we obtain
\begin{equation}\label{eq:J-avg-pca}
    |d^{-1}\sum\nolimits_{i=1}^dJ(b_i,q_i)-\cD(R^2)|
    \le(C_L/d)\sum\nolimits_{i=1}^d(|b_i|+|\theta_i|)
    \le C_Ld^{-1/2}.
\end{equation}
Substituting \eqref{eq:pca-mean-expansion}, \eqref{eq:Tv-pca-bound}, and \eqref{eq:J-avg-pca} into \eqref{eq:psi-prelimit-pca} gives the desired drifts,
\begin{align*}
    \Psi_n^m
    &=\alpha(R^2)(H(m)-A(R^2)m)
    +O(d^{-1+2\zeta_+}+d^{-1/2}),\\
    \Psi_n^{R^2}
    &=2\alpha(R^2)(H(m)m-A(R^2)R^2)+c_\delta\cD(R^2)
    +O(d^{-1+2\zeta_+}+d^{-1/2}),
\end{align*}
which proves the required closure on $E_L \cap A_d$ and identifies \eqref{eq:adam-ode-pca}.

\begin{proof}[Proof of Proposition~\ref{prop:pca-sgdm-sgdu}]

    Differentiating $R=\sqrt{R^2}$, $a=m/R$, and $y=R^2-m^2$ in yields \eqref{eq:pca-common-coordinate-transform}. The coefficients follow from the dynamics for SGD and SGD-M from \cite{BenArousGheissariJagannath2024Effective, jagannath2026high}.
\end{proof}

\begin{proof}[Proof of Corollary~\ref{cor:adam-pca-eps-zero}]
    Fix $\beta_1\in[0,1)$ and $\beta_2\in(0,1)$. On $E_L$, the inequality $0\le x^{-1/2}-(x+\epsilon)^{-1/2}\le\epsilon/(2x^{3/2})$ gives,
    \[
        \sup\nolimits_{R \le L}
        (|\alpha_\epsilon(R^2)
        -\kappa_1/(2\sqrt{k}R^{k-1})|
        +|\cD_\epsilon(R^2)-\kappa_2|)
        \le C_{L} \epsilon,
    \]
    which follows from finiteness of the two series $\kappa_1$ and $\kappa_2$.
\end{proof}

\begin{proof}[Proof of Proposition~\ref{prop:adam-pca-global}]

    Part (iii) follows from \cite{BenArousGheissariJagannath2024Effective, jagannath2026high}. For part (i) and (ii), write the dynamics as $\sqrt{k}\,\kappa_1 (F_R(R, a), F_a(R, a))$, where
    \begin{equation}\label{eq:pca-FR-Fa-defs}
        F_R(R, a) := \lambda a^k - R^k + \zeta_{c_\delta}R^{-1}, \qquad F_a(R, a) := \lambda R^{-1} a^{k-1}(1 - a^2) - \zeta_{c_\delta}R^{-2} a.
    \end{equation}
    Setting $a = 0$ gives $F_a(R, 0) = 0$ for all $R > 0$ since $k \ge 2$, while $F_R(R, 0) = -R^k + \zeta_{c_\delta} R^{-1} = 0$ has the root $R_{\mathrm{eq}} = \zeta_{c_\delta}^{1/(k+1)}$.

    On the other hand, consider $a \neq 0$. Solving $F_R(R, a) = 0$ for $\zeta_{c_\delta} R^{-1} = R^k - \lambda a^k$ and substituting into $R F_a(R, a) = 0$ gives
    \[
        0 = \lambda a^{k-1}(1 - a^2) - \bigl(R^k - \lambda a^k\bigr)a = a\bigl(\lambda a^{k-2} - R^k\bigr).
    \]
    Since $a \neq 0$, we obtain the system of equations, $R^k = \lambda a^{k-2}$ and $R^{k+1}(1 - a^2) = \zeta_{c_\delta}$. When $k = 2$, the first equation gives $R_* = \sqrt{\lambda}$, and the second becomes $\lambda^{3/2}(1 - a_*^2) = \zeta_{c_\delta}$, which admits real solutions $a_* \neq 0$ if and only if $\lambda > \zeta_{c_\delta}^{2/3}$.
\end{proof}

\subsection{Proofs of the cross-over and risk-threshold conclusions}

\begin{proof}[Proof of Corollary~\ref{cor:pca-population-risk}]
    Differentiating \eqref{eq:pca-population-risk} along the common dynamics,
    \begin{align*}
        \dot\cR
        =kR^{2k-2}\dot{R^2}-2k\lambda m^{k-1}\dot m
        =-4k^2\alpha(R^2)\cB_k(m,R)
        +c_\delta kR^{2k-2}\cD(R^2).
    \end{align*}
    The squared-norm representation of $\cB_k$ follows by expansion using $\norm{\theta_*}=1$ and $\langle\theta,\theta_*\rangle=m$.
\end{proof}

We may also obtain the separate limits for each of the respective generators. Note that the population Hessian is given by,
\[
    \nabla^2\cR_d
    =2kR^{2k-2}I_d+4k(k-1)R^{2k-4}\theta\theta^\top
    -2k(k-1)\lambda m^{k-2}\theta_*\theta_*^\top.
\]
By \eqref{eq:pca-strong-proj}, the last two terms are finite rank and do not contribute to the second-order generator $\delta \tilde \cL$. Similarly, by the finite-dimensional projection argument in the proof of Lemma~\ref{lem:risk-operator-split}, their contraction with $\{ \rho , \av \}$ also vanishes in $\tilde \cA \chi_1$. Thus the first term of the Hessian and \eqref{eq:pca-D-def} give on $E_L \cap A_d$,
\[
    \widetilde\cA_\theta\cR_d\rightarrow4k^2\alpha(R^2)\cB_k(m,R),
    \quad d^{-1}\widetilde\cL_\theta\cR_d
    \rightarrow kR^{2k-2}\Gamma_0(R^2),
\quad
    -d^{-1}\widetilde\cA_\theta\chi_{1,\theta}^{\cR_d}
    \rightarrow2kR^{2k-2}\sum\Gamma_h(R^2).
\]
For SGD, $\Gamma_0(R^2)=4kR^{2k-2}$ from which we may compute,
$$
    d^{-1}\EE\bigl[g^\top\nabla^2\cR_d\,g\bigr]
    =8k^2R^{4k-4}+O(d^{-1}),\qquad
    \norm{\nabla\cR_d}^2=4k^2\cB_k(m,R).
$$
Dividing the two quantities proves the Tensor PCA part of Proposition~\ref{prop:app-eff-inf-lap-window}.

\begin{proof}[Proof of Proposition~\ref{prop:app-pca-crossovers}]
    Write $\zeta=\kappa_2/(2\sqrt2\kappa_1)$, so $\zeta_{c_\delta}=c_\delta\zeta$ when $k=2$. If $\lambda\le c_\delta<\lambda^{3/2}/\zeta$, then SGD and SGD-M fall in the zero-alignment regime and ADAM is in its learned regime for $m_0 \neq 0$, thus proving the first assertion. Next, initialize all algorithms at $R_0=1$ and $a_0=m_0>0$. The common transformation gives the exact difference
    \[
        a_0^{-1}(\dot a_{\mathrm{SGD}}(0)-\dot a_{\mathrm A}(0))
        =(4-\sqrt2\kappa_1)\lambda(1-a_0^2)
        - ( 4 - \kappa/2 ) c_\delta.
    \]
    For $\kappa_1<2\sqrt2$ and $\kappa_2<8$, let $c_*=(8-2\sqrt2\kappa_1)/(8-\kappa_2)>0$. If $c_\delta<c_*\lambda$, the difference is positive for sufficiently small $a_0>0$. Hence SGD has strictly greater alignment on a nonempty initial time interval. If also $\lambda>\zeta^2$ and $c_\delta<\lambda$, both trajectories have positive limiting alignment, and
    \[
        (a_{\mathrm A,\infty})^2
        =1-\lambda^{-3/2}\zeta c_\delta
        >1-\lambda^{-1}c_\delta
        =(a_{\mathrm{SGD},\infty})^2,
    \]
    implying that ADAM attains a strictly larger equilibrium alignment as both have their positive initial signs preserved from $a_0$. Continuity therefore gives a crossing, after which ADAM has greater alignment.

    For the risk comparison, put $\cB(m,R)=R^4(R^2-m^2)+m^2(R^2-\lambda)^2$. At $R_0=1$, \eqref{eq:pca-exact-risk-ode-adam} gives
    \[
        \dot\cR_{\mathrm{SGD}}(0)=-16\cB(m_0,1)+16c_\delta,\qquad
        \dot\cR_{\mathrm A}(0)=-4\sqrt2\kappa_1\cB(m_0,1)+2\kappa_2c_\delta,
    \]
    where $\cB(m_0,1)=1+(\lambda^2-2\lambda)m_0^2\to1$ as $m_0\downarrow0$. Hence $c_\delta<c_*$ implies $\dot\cR_{\mathrm{SGD}}(0)<\dot\cR_{\mathrm A}(0)$ for all sufficiently small positive $m_0$ (which can be observed by looking at the expression $\cR_{\mathrm A}(0) - \dot\cR_{\mathrm{SGD}}(0)$). At $t \to \infty$, Proposition~\ref{prop:adam-pca-global} gives,
    \[
        \cR_{\mathrm A,\infty}
        =-\lambda^2+2\zeta c_\delta\sqrt\lambda
        <-\lambda^2+2\lambda c_\delta
        =\cR_{\mathrm{SGD},\infty}
        \qquad\text{if }\lambda>\zeta^2 \text{ and }c_\delta < \lambda.
    \]
    Thus $c_\delta<\min\{\lambda,c_*\}$ gives a risk cross-over by the same continuity argument.
\end{proof}

\begin{proof}[Matrix PCA assertion of
    Proposition~\ref{prop:app-pca-eoss-contraction}] For $k=2$, $\cB_2(m,R)=\cB(m,R)=R^4(R^2-m^2)+m^2(R^2-\lambda)^2$. The risk equations are
    \[
        \dot\cR_{\mathrm{SGD}}=-16\cB+16c_\delta R^4,\qquad
        \dot\cR_{\mathrm A}
        =-(4\sqrt2\kappa_1/R) \cB+2c_\delta\kappa_2R^2.
    \]
    In the notation $\dot\cR=-\cF+c_\delta\cD$ and $\Gamma^{\mathrm{eff}}=2\cD/\cF$ of Section~\ref{sec:applications}, these identify
    \[
        \Gamma_{\mathrm{SGD}}^{\mathrm{eff}}(m,R)
        =2 R^4/\cB(m,R),
        \qquad
        \Gamma_{\mathrm A}^{\mathrm{eff}}(m,R)
        =\kappa_2R^3/(\sqrt2\kappa_1\cB(m,R)).
    \]
    By Proposition~\ref{prop:adam-pca-global}, at equilibrium for ADAM $R_\infty=\sqrt\lambda$ and $y_\infty=z/\sqrt\lambda$, so $\cB(m_\infty,R_\infty)=zR_\infty^3$. Otherwise, $m_\infty=0$ and $R_\infty^3=z$, giving the same identity. Consequently
    \[
        \Gamma_{\mathrm A}^{\mathrm{eff}}(m_t,R_t)
        \rightarrow \kappa_2/(\sqrt2\kappa_1z)
        =2/c_\delta.
    \]
    For SGD, both regimes give $\cB(m_\infty,R_\infty)=c_\delta R_\infty^4$, and hence $\Gamma_{\mathrm{SGD}}^{\mathrm{eff}}(m_t,R_t)\to2/c_\delta$.
\end{proof}

\section{Gaussian Projective Models}
\label{sec:ex-gpm}

We present the limiting dynamics for a general family of data models which we call Gaussian projective models. The later sections regarding the multi-class logistic regression and Gaussian XOR classification models are treated as specific examples from this family.

Suppose the trainable parameter decomposes as $\theta=(\theta^1,\ldots,\theta^q,w)$ with $\theta^a\in\RR^d$, $w\in\RR^s$, and $p=qd+s$ where where $q,k\ge1$ and $s,q_*\ge0$ are fixed integers. Each sample is a pair $Y=(J,X)$ from an isotropic Gaussian mixture,
\[
J\sim\operatorname{Cat}(\mathsf p_1,\ldots,\mathsf p_k),
    \qquad X=\mu^J+\sigma Z,\qquad Z\sim\cN(0,I_d),
\]
with $J$ independent of $Z$ and $\sigma>0$. The teacher directions $\theta_*^1,\ldots,\theta_*^{q_*}$ and mixture means $\mu^1,\ldots,\mu^k$ are deterministic and have uniformly bounded norms. Write $P_\theta(X)= (\langle\theta^a,X\rangle_{a\le q}, \langle\theta_*^b,X\rangle_{b\le q_*} )$. The loss is $L(\theta,Y)=\Upsilon(P_\theta(X),w;y(J))$. For a joint coordinate $\xi=(a,i)$ with $a\le q$ and $i\le d$, its gradient is
\begin{equation}\label{eq:g-coord}
    g_{ai}(\theta,Y)=\alpha_a(\theta,Y)X_i,
    \qquad \alpha_a(\theta,Y)
    :=\partial_{z_a}\Upsilon(P_\theta(X),w;y(J)),
\end{equation}
while the second-layer gradient coordinates are $\partial_{w_\nu}\Upsilon( P_\theta(X),w;y(J) )$.

\begin{assumption}\label{assume:gpm-loss}
    For each label $y\in\{y(1),\ldots,y(k)\}$, the map $(z,z_*,\mathsf w)\mapsto\Upsilon(z,z_*,\mathsf w;y)$ is $C^3$. For every $S\ge1$, there are constants $A_S<\infty$ and $m_S\ge1$, independent of $d$, such that, for $j=1,2,3$,
    \begin{equation}\label{eq:gpm-R}
        \sup\nolimits_{\norm{\mathsf w}\le S}
        \opnorm{\Dd^j\Upsilon(z,z_*,\mathsf w;y)}
        \le A_S(1+\norm z+\norm{z_*})^{m_S}.
    \end{equation}
    Here $\Dd^j$ denotes the full derivative with respect to $(z,z_*,\mathsf w)$.
\end{assumption}

Collect all student directions, teacher directions, and mixture means into
\begin{equation}\label{eq:gpm-tracked-vectors}
    v(\theta)=(v^1,\ldots,v^r)
    :=(\theta^1,\ldots,\theta^q,
    \theta_*^1,\ldots,\theta_*^{q_*},\mu^1,\ldots,\mu^k),
    \qquad r=q+q_*+k,
\end{equation}
and track the statistics $\uv(\theta)=(G(\theta),w)$ with $G(\theta)=v(\theta)^\top v(\theta)$. Throughout this section, we consider the critical scaling regime $\delta_d = c_\delta / d$ for $c_\delta > 0$ so that $\delta_dp\to qc_\delta$.

We now state the general ballistic limits for models that fall into this family. Let $(J_t,P_t,\xi_t)_{t\in\ZZ}$ be independent over time, with $J_t\sim\operatorname{Cat}(\mathsf p_1,\ldots,\mathsf p_k)$ and $\xi_t\sim\cN(0,1)$ where $\xi_t$ is independent of $(J_t,P_t)$, and
\begin{equation}\label{eq:gpm-rep-proj}
    P_t\mid\{J_t=c\}\sim
    \cN(G_{\cdot,q+q_*+c},\sigma^2G).
\end{equation}
Write $P_t^{\mathrm{st}}=P_{t,1:q+q_*}$ and $\alpha_{t,a}=\partial_{z_a}\Upsilon(P_t^{\mathrm{st}},w;y(J_t))$. The representative coordinate gradients are
\begin{equation}\label{eq:gpm-rep-grad}
    g_t^a=\sigma\alpha_{t,a}\xi_t,\qquad
    g_t^{w_\nu}=\partial_{w_\nu}\Upsilon(P_t^{\mathrm{st}},w;y(J_t)).
\end{equation}
Fix $\beta_1,\beta_2\in[0,1)$ for ADAM and for each block define,
\begin{equation}\label{eq:gpm-rep-adam}
        M_t^\diamond
        =\sum\nolimits_{j\ge0}(1-\beta_{1,\diamond})
        \beta_{1,\diamond}^j g_{t-j}^\diamond,\quad
        V_t^\diamond
        =\sum\nolimits_{j\ge0}(1-\beta_{2,\diamond})
        \beta_{2,\diamond}^j(g_{t-j}^\diamond)^2, \quad
        \tilde D_t^\diamond
        =(M_t^\diamond)/\sqrt{V_t^\diamond+\epsilon_\diamond}.
\end{equation}
where $\diamond\in\{1,\ldots,q,w_1,\ldots,w_s\}$. All blocks use the same sample sequence and, in the first layer, the same $\xi_t$. For $a\le q$ and $b\le r$, set
\begin{align}
    F_{ab}(G,w)
    &:=\sum\nolimits_{j\ge0}(1-\beta_1)\beta_1^j
    \EE[
    \alpha_{-j,a}P_{-j,b}\xi_{-j}^2/
    \sqrt{V_0^a+\epsilon}],
    \label{eq:gpm-F-def}\\
    \cD_{ab}(G,w)
    &:=\EE[\tilde D_0^a\tilde D_0^b]
    +\sum\nolimits_{h\ge1}\EE[
    \tilde D_0^a\tilde D_h^b+\tilde D_h^a\tilde D_0^b],
    \qquad a,b\le q.
    \label{eq:gpm-D-def}
\end{align}

\begin{proposition}[ADAM dynamics]\label{prop:adam-gpm-dynamics}
    Suppose Assumption~\ref{assume:gpm-loss} holds. Fix the first-layer parameters $\beta_1,\beta_2\in[0,1)$ and $\epsilon>0$, and assume that, for some $0\le\zeta<1/2$,
    \[
        \max_{a\le q}\norm{\theta_0^a}_\infty,
        \quad\max_{b\le q_*}\norm{\theta_*^b}_\infty,
        \quad\max_{c\le k}\norm{\mu^c}_\infty
        <d^{-1/2+\zeta}.
    \]
    If $\uv(\theta_0)\Rightarrow\nu$, the interpolated statistics converge to the solution of
    \begin{equation}\label{eq:gpm-ode}
        \begin{aligned}
            \dot G_{ab}&=-F_{ab}(G,w)-F_{ba}(G,w)
            +c_\delta\cD_{ab}(G,w),&&a,b\le q,\\
            \dot G_{ab}&=-F_{ab}(G,w),&&a\le q<b\le r,\\
            \dot w_\nu&=-\EE[\tilde D_0^{w_\nu}],&&\nu\le s,
        \end{aligned}
    \end{equation}
    initialized from $\uv_0\sim\nu$. Note that $\dot G_{ab}=0$ for $a,b>q$.
\end{proposition}

For online SGD, whose dynamics were obtained in \cite{GheissariJagannath2025Universality}, the same system holds with
\begin{equation}\label{eq:gpm-sgd-FD}
    F_{\cdot,b}^{\mathrm{SGD}}=\EE[\alpha_0P_{0,b}],\qquad
    \cD^{\mathrm{SGD}}=\sigma^2\EE[\alpha_0\alpha_0^\top],\qquad
    \dot w=-\EE[\nabla_w\Upsilon(P_0^{\mathrm{st}},w;y(J_0))].
\end{equation}

\subsubsection{Matrix-normalized first layers}

For Exact MUON, write $f_{\epsilon_M}(S)=(S+\epsilon_MI_q)^{-1/2}$, with $\epsilon_M>0$. For a fixed number $K\ge1$ of Newton--Schulz steps with fixed coefficients $(\mathsf a,\mathsf b,\mathsf c)$ and $\mathsf a>0$, define the recursion
\begin{equation}\label{eq:ns5-scalar-poly}
    p_0(t)=1,\qquad
    p_{j+1}(t)=p_j(t)\bigl(\mathsf a+
    \mathsf b\,t\,p_j(t)^2+\mathsf c\,t^2p_j(t)^4\bigr),
    \qquad 0\le j<K,
\end{equation}
and
\begin{equation}\label{eq:ns5-R5-map}
    R_K(S)=\tau(S)^{-1/2}p_K\!\left(\frac{S}{\tau(S)}\right),
    \qquad \tau(S)=\tr(S)+\epsilon_F,\qquad S\in\Sd_+^q,
\end{equation}
where $\epsilon_F>0$. In particular, $R_5$ is obtained by setting $K=5$. The corresponding direction is
\[
    \rho_\theta(M)=M\mathsf R(M^\top M/d),\qquad
    \mathsf R=f_{\epsilon_M}\ \text{or}\ R_K.
\]
The exact factorization for Newton--Schulz follows by the polynomial induction in Lemma~\ref{lem:ns5-factorization-bounds}. Let $\omega_j^M=(1-\beta_M)\beta_M^j$, where $\beta_M\in[0,1)$, and define
\begin{equation}\label{eq:muon-C-R-def}
    C_t=\sigma^2\sum\nolimits_{j\ge0}(\omega_j^M)^2
    \alpha_{t-j}\alpha_{t-j}^\top,
    \qquad R_t=\mathsf R(C_t).
\end{equation}
For Exact MUON, $R_t=(C_t+\epsilon_MI_q)^{-1/2}$; for MUON--NS$K$, $R_t=R_K(C_t)$. Set
\begin{align}
    F_{\cdot,b}^{\mathsf R}(G,w)
    &:=\sum\nolimits_{j\ge0}\omega_j^M
    \EE[R_0\alpha_{-j}P_{-j,b}],\qquad b\le r,
    \label{eq:gpm-matrix-F}\\
    \cD^{\mathsf R}(G,w)
    &:=\EE[R_0C_0R_0]
    +\sum\nolimits_{h\geq1}\beta_M^h
    \EE[R_0C_0R_h+R_hC_0R_0].
    \label{eq:gpm-matrix-D}
\end{align}

\begin{proposition}[Matrix-normalized dynamics]\label{prop:gpm-matrix-dynamics}
    Suppose Assumption~\ref{assume:gpm-loss} holds. Apply Exact MUON or fixed-$K$ MUON--NS$K$ to the first layer, with $\beta_M\in[0,1)$. If $s>0$, apply either SGD or coordinatewise ADAM with its own fixed parameters $\beta_{1,w},\beta_{2,w}\in[0,1)$ and $\epsilon_w>0$ to $w$. If $\uv(\theta_0)\Rightarrow\nu$, the statistics converge to the system \ref{eq:gpm-ode} with $F = F^{\mathsf R}$, $\cD = \cD^{\mathsf R}$, and the corresponding dynamics for the second layer $\dot{w}_\nu$.
\end{proposition}

\subsection{Population risk}

Put $\ell=q+q_*$, $Q=G_{1:\ell,1:\ell}$, and $m^c=G_{1:\ell,\ell+c}$. The population risk is
\begin{equation}\label{eq:gpm-pop-risk}
    \cR(G,w)=\sum\mathsf p_c
    \EE[\Upsilon(Z_c,w;y(c))],\qquad
    Z_c\sim\cN(m^c,\sigma^2Q).
\end{equation}
Thus convergence of the summary statistics implies convergence of the risk by continuity.

\begin{lemma} \label{lem:gaussian-risk-chain-rule}
    Let $t\mapsto(Q_t,m_t^1,\ldots,m_t^k,w_t)$ be a $C^1$ path on $[0, T]$. Under Assumption~\ref{assume:gpm-loss},
    \begin{equation}\label{eq:gpm-risk-chain-rule}
        \begin{aligned}
            \frac{\dd}{\dd t}\cR(Q_t,m_t,w_t)
            &=\sum_{c=1}^k\mathsf p_c\Bigl\{
            \EE[\nabla_z\Upsilon(Z_c,w_t;y(c))\cdot\dot m_t^c] +\EE[\nabla_w\Upsilon(Z_c,w_t;y(c))\cdot\dot w_t]\\
            &\qquad+\frac{\sigma^2}{2}
            \EE[\tr\bigl(\nabla_z^2\Upsilon(Z_c,w_t;y(c))\dot Q_t\bigr)]
            \Bigr\},
        \end{aligned}
    \end{equation}
    where $Z_c\sim\cN(m_t^c,\sigma^2Q_t)$.
\end{lemma}

\begin{proof}
    The proof follows from Gaussian integration by parts, see for example Lemma 2.8 of \cite{GheissariJagannath2025Universality}.
\end{proof}

Write the averages $\bar\alpha^c=\EE[\nabla_{z_{1:q}}\Upsilon(Z_c,w;y(c))]$ along with,
\[
    \bar H=\sum_c\mathsf p_c\EE[\nabla_z^2\Upsilon(Z_c,w;y(c))],
    \quad \bar b=\sum_c\mathsf p_c\EE[\nabla_w\Upsilon(Z_c,w;y(c))],
\]
and let $\bar d_w=\EE[\tilde D_0^w]$. For either dynamics, denote its first-layer coefficients by $(F,\cD)$. Lemma~\ref{lem:gaussian-risk-chain-rule} gives the decomposition of the risk $\dot\cR=-\cF_{\mathrm{alg}}+c_\delta\cD_{\mathrm{alg}}$ where,
\begin{equation}\label{eq:gpm-risk-F+D}
    \cF_{\mathrm{alg}}
    :=\sum_{c=1}^k\mathsf p_c\,
    \bar\alpha^c\cdot F_{\cdot,\ell+c}
    +\sigma^2\sum_{a=1}^q\sum_{b=1}^{\ell}\bar H_{ab}F_{ab}
    +\bar b\cdot\bar d_w,
    \qquad
    \cD_{\mathrm{alg}}
    :=\frac{\sigma^2}{2}\sum_{a,b=1}^q\bar H_{ab}\cD_{ab}.
\end{equation}
Combining our proof of Assumption~\ref{assume:closable} with Lemma~\ref{lem:gaussian-risk-chain-rule} immediately yields the following.

\begin{lemma} \label{lem:risk-operator-split}
    For the Gaussian projective model and algorithms above, the following limits hold uniformly on  $E_L \cap A_d$ for ADAM for $E_L = \{ \theta : \| \theta < L \}$
    \begin{equation}\label{eq:lem-risk-op-limits}
        \tilde\cA_\theta\cR_d\rightarrow\cF_{\mathrm{alg}},
        \qquad d^{-1}\tilde\cL_\theta\cR_d
        \rightarrow\cD_{\mathrm{alg}}^{(0)},
        \qquad -d^{-1}\tilde\cA_\theta\chi_{1,\theta}^{\cR_d}
        \rightarrow\cD_{\mathrm{alg}}^{(\mathrm{mem})},
    \end{equation}
    where $\cR_d(\theta)=\cR(G(\theta),w)$ and
    \begin{equation}\label{eq:gpm-D0-Dmem-split}
            \cD_{\mathrm{alg}}^{(0)}
            =\frac{\sigma^2}{2}\sum_{a,b\le q}
            \bar H_{ab}\Gamma_{0,ab}^{\mathrm{alg}}, \qquad \cD_{\mathrm{alg}}^{(\mathrm{mem})}
            =\frac{\sigma^2}{2}\sum_{a,b\le q}\bar H_{ab}
            \sum_{h\ge1}
            (\Gamma_{h,ab}^{\mathrm{alg}}+
            \Gamma_{h,ba}^{\mathrm{alg}}).
    \end{equation}
\end{lemma}

When both layers use online SGD, $\tilde\cA_\theta\cR_d=\norm{\nabla\cR_d}^2$. Hence, when $\cF_{\mathrm{SGD}} > 0$, the effective infinity Laplacian has the limit,
$$
    \frac1d\,
    \frac{\EE[\nabla L(\theta,Y)^\top
    \nabla^2\cR_d(\theta)\nabla L(\theta,Y)]}
    {\norm{\nabla\cR_d(\theta)}^2}
    \rightarrow
    \frac{2\cD_{\mathrm{SGD}}}{\cF_{\mathrm{SGD}}}.
$$
This proves the Gaussian projective-model part of Proposition~\ref{prop:app-eff-inf-lap-window}.

\subsection{Coordinate delocalization of the parameter}
\label{sec:deloc}

Let $R_0$ bound all teacher and mean norms. In this subsection the first layer is updated by coordinatewise ADAM, and this is the algorithm for which coordinate delocalization is needed. For $0\le\zeta<\zeta_+<1/2$, put $\sD_\zeta=\{v\in\RR^d:\norm v_\infty<d^{-1/2+\zeta}\}$, and define $B_S=\{\theta:\norm\theta<S\}$ with $A_d=\{\theta:\max_{a\le q}\norm{\theta^a}_\infty <d^{-1/2+\zeta_+}\}$. We also define the stopping times $\tau_S^\theta=\inf\{n\ge0:\theta_n\notin B_S\}$ and $\tau_A=\inf\{n\ge0:\theta_n\notin A_d\}$.

\begin{lemma}\label{lem:moments-gauss}
    Fix $S\ge1$ and $m\ge2$. The following estimates hold with constants independent of $d$.
    \begin{enumerate}[label=\textup{(\alph*)},leftmargin=*]
        \item For every deterministic $v\in\RR^d$,
        \begin{equation}\label{eq:proj-Upsilon-moment}
            \norm{v\cdot X}_{L^m}\le(R_0+\sigma c_m)\norm v,
            \qquad
            \sup_{\theta\in B_S}\max_{1\le j\le3}
            \left\|\opnorm{\Dd^j\Upsilon(P_\theta(X),w;y(J))}
            \right\|_{L^m}\le C_{S,m},
        \end{equation}
        where $c_m=\norm{\cN(0,1)}_{L^m}$. Moreover,
        \begin{equation}\label{eq:g-moments}
            \sup_{\theta\in B_S}\max_\xi
            \left(\norm{g_\xi}_{L^m}
            +\sup_{\norm h\le1}\norm{D_hg_\xi}_{L^m}
            +\sup_{\norm h,\norm k\le1}\norm{D_kD_hg_\xi}_{L^m}
            \right)\le C_{S,m}.
        \end{equation}
        \item Up to $\tau_S^\theta$, the ADAM momentum coordinates and all search directions have bounded moments of every fixed order. In particular, they satisfy,
        \begin{equation}\label{eq:traj-all-r}
            \sup_{\ell,\xi}\left(
            \norm{1+\norm{X_{\ell,\xi}}_2}_{L^m}
            +\norm{D_{\ell,\xi}}_{L^m}\right)\le C_{S,m},
            \qquad
            \sup_{\ell,\xi}
            \norm{(1+\norm{X_{\ell,\xi}}_2)\norm{D_\ell}}_{L^m}
            \le C_{S,m}\sqrt p.
        \end{equation}
        The same estimates hold under the stationary fixed-parameter law for $\theta\in B_S$.
    \end{enumerate}
\end{lemma}

\begin{proof}
    Since $v\cdot X=v\cdot\mu^J+\sigma v\cdot Z$, the first bound in \eqref{eq:proj-Upsilon-moment} is immediate. There are only finitely many tracked projections, all with bounded variance on $B_S$. Assumption~\ref{assume:gpm-loss} therefore proves the second bound at every moment order. For $h=(h^1,\ldots,h^q,h^w)$, the linear map $T_Xh=(\langle h^1,X\rangle,\ldots,\langle h^q,X\rangle,h^w)$ satisfies $\norm{\norm{T_Xh}}_{L^m}\le C_m\norm h$. Differentiating \eqref{eq:g-coord} twice and applying H\"older's inequality proves \eqref{eq:g-moments}.

    Next, unrolling each momentum recursion and applying Minkowski's inequality bounds its coordinate moments. In each ADAM block, using  $|D_{\ell,\xi}|\le\epsilon_\xi^{-1/2}|M_{\ell,\xi}|$, summing squared coordinate norms yields $\norm{\norm{D_\ell}}_{L^{2m}}\le C_{S,m}\sqrt p$. H\"older's inequality then gives the last bound in \eqref{eq:traj-all-r}. The stationary proof uses the same convergent geometric sums.
\end{proof}

For a coordinate $\xi$, write $\bar D_\xi(\theta)=\pi_\theta\rho_\xi$ and recall the coordinate corrector $a_\xi(\theta,x_\xi) =\sum_{h\ge1}\bigl(K_\theta^h\rho_\xi(x_\xi) -\bar D_\xi(\theta)\bigr)$. The corrector estimates in Corollary~\ref{cor:poisson-w3-scale-uniform}, with coordinate scale one, apply on $B_S$ by \eqref{eq:g-moments}. In particular, $a_\xi$ and its first parameter derivative are bounded by $C_S(1+\norm{x_\xi}_2)$, and
\begin{equation}\label{eq:coord-poisson-id}
    (I-K_\theta)a_\xi=K_\theta\rho_\xi-\bar D_\xi(\theta),
    \qquad
    K_\theta(\rho_\xi+a_\xi)=a_\xi+\bar D_\xi(\theta).
\end{equation}

\begin{lemma}\label{lem:stationary-drift}
    Uniformly over $\theta\in B_S$, $a\le q$, and $i\le d$,
    \begin{equation}\label{eq:drift-U}
        |\bar D_{ai}(\theta)|\le L_S\left(
        \max_{b\le q}|\theta_i^b|
        +\max_{b\le q_*}|\theta_{*,i}^b|
        +\max_{c\le k}|\mu_i^c|\right).
    \end{equation}
\end{lemma}

\begin{proof}
    Put $h_i=(\theta_i^1,\ldots,\theta_i^q, \theta_{*,i}^1,\ldots,\theta_{*,i}^{q_*})$ and $b_i=\norm{h_i}+\max_c|\mu_i^c|$. Decompose $P_\theta(X)=P_\theta^{(-i)}(X)+h_iX_i$, where $P_\theta^{(-i)}(X)$ is independent of $Z_i$. Compare the gradients
    \(g=\alpha_a(P_\theta(X),w;y(J))X_i\) and \(
        \widehat g=\sigma\alpha_a(P_\theta^{(-i})(X),w;y(J))Z_i\).
    The fundamental theorem of calculus and the Gaussian moment bounds yield, for every fixed $m$,
   \(\norm{g-\widehat g}_{L^m}\le C_{S,m}b_i.\)
    Unrolling the stationary momentum recursions gives the same bound for the difference of their first moments and, using $g^2-\widehat g^2=(g-\widehat g)(g+\widehat g)$, for their second moments. Since $\gamma(v)=(v+\epsilon)^{-1/2}$ satisfies $|\gamma|\le\epsilon^{-1/2}$ and $|\gamma'|\le\epsilon^{-3/2}/2$ on $[0,\infty)$,
    \begin{equation}\label{eq:direction-diff}
        \norm{M\gamma(V)-\widehat M\gamma(\widehat V)}_{L^m}
        \le\epsilon^{-1/2}\norm{M-\widehat M}_{L^m}
        +\norm{\widehat M}_{L^{2m}}
        \norm{V-\widehat V}_{L^{2m}}/(2\epsilon^{3/2})
        \le C_{S,m}b_i.
    \end{equation}
    Conditional on the labels and all Gaussian coordinates except coordinate $i$, simultaneous sign reversal of $(Z_{t,i})_{t\in\ZZ}$ negates $\widehat M$ and leaves $\widehat V$ unchanged. The comparison direction therefore has mean zero, i.e. $\EE [\widehat{M} \gamma (\widehat{V} )]$. Taking expectations in \eqref{eq:direction-diff} with $m = 1$ proves \eqref{eq:drift-U}.
\end{proof}

\begin{theorem}[Coordinate delocalization for ADAM]\label{thm:deloc}
    Suppose Assumption~\ref{assume:gpm-loss} holds, the ADAM parameters are fixed, and $\delta_d=c_\delta/d$. Assume $\theta_0^a,\theta_*^b,\mu^c\in\sD_\zeta$ for some $0\le\zeta<1/2$. For every $\zeta<\zeta_+<1/2$ and $S\ge1$, there is $c_S>0$ such that
    \begin{equation}\label{eq:thmC-prop}
        \PP\left(\exists\,n\le\tau_S^\theta\wedge
        \lfloor c_S\delta_d^{-1}\log d\rfloor:
        \max_{a\le q}\norm{\theta_n^a}_\infty
        \ge d^{-1/2+\zeta_+}\right)\rightarrow0.
    \end{equation}
\end{theorem}

\begin{proof}\label{subsec:proof-deloc}
    Fix $S$, $t\ge0$, and $N=\lfloor t/\delta\rfloor$. For each coordinate $\xi$, let $H_{\ell,\xi}=\rho_\xi(X_\ell)+a_\xi(\theta_{\ell-1},X_{\ell,\xi})$, and define the martingale differences on $\ell \le \tau_S^\theta$,
    \begin{equation}\label{eq:deloc-mart-diff}
        \Delta\cM_{\ell,\xi}
        =-
        \left(H_{\ell,\xi}
        -\EE[H_{\ell,\xi}\mid\cF_{\ell-1}]\right),
        \qquad \cM_{n,\xi}=\sum\nolimits_{\ell=1}^n\Delta\cM_{\ell,\xi}.
    \end{equation}
    By \eqref{eq:coord-poisson-id}, for $h\le\tau_S^\theta$, we have the coordinate expansion of the parameter,
    \begin{equation}\label{eq:theta-decomp}
        \theta_{h,\xi}=\theta_{0,\xi}
        -\delta\sum\nolimits_{\ell=1}^h\bar D_\xi(\theta_{\ell-1})
        +\delta\cM_{h,\xi}+\cB_{h,\xi}+\cR_{h,\xi},
    \end{equation}
    where the remainder terms are given by,
    \begin{equation}\label{eq:B-R-def}
            \cB_{h,\xi}
            =\delta[a_\xi(\theta_{h-1},X_{h,\xi})
            -a_\xi(\theta_0,0)],\qquad \cR_{h,\xi}
            =-\delta\sum\nolimits_{\ell=1}^{h-1}
            [a_\xi(\theta_\ell,X_{\ell,\xi})
            -a_\xi(\theta_{\ell-1},X_{\ell,\xi})].
    \end{equation}
    For each fixed $m\ge2$, Lemma~\ref{lem:moments-gauss} and the coordinate corrector bounds give $\norm{\Delta\cM_{\ell,\xi}}_{L^m}\le C_{S,m}$. Burkholder's inequality and Minkowski's inequality in $L^{m/2}$ give the first of the following estimates:
    \begin{equation}\label{eq:three-error-bounds}
        \begin{aligned}
            \EE\max_{n\le N}|\delta\cM_{n,\xi}|^m
            \le C_{S,m}t^{m/2}d^{-m/2},&\qquad
            \EE\max_{h\le\tau_S^\theta\wedge N}|\cB_{h,\xi}|^m
            \le C_{S,m}(1+t)d^{-m/2},\\
            \EE\max_{h\le\tau_S^\theta\wedge N}|\cR_{h,\xi}|^m
            &\le C_{S,m}t^m d^{-m/2}.
        \end{aligned}
    \end{equation}
    For the second estimate, bound the maximum by the inequality $\max_{i \le n} |x_i|^m \le \sum_{i \le n} |x_i|^m$. This gives $C_{S,m}\delta^m(N+1)\le C_{S,m}(1+t)d^{1-m}$ for $m \ge 2$. For the third, the mean-value theorem and Corollary~\ref{cor:poisson-w3-scale-uniform} gives,
    \[
        |a_\xi(\theta_\ell,X_{\ell,\xi})
        -a_\xi(\theta_{\ell-1},X_{\ell,\xi})|
        \le C_S\delta(1+\norm{X_{\ell,\xi}}_2)\norm{D_\ell},
    \]
    whose $L^m$ norm is at most $C_{S,m}\delta\sqrt d$. Summing in \eqref{eq:B-R-def} proves the third estimate.

    Let $E_N$ be the maximum over first-layer coordinates of the three error suprema in \eqref{eq:three-error-bounds}. That is, $E_N = \max_{h, \xi} (|\delta\cM_{n,\xi}| + |\cB_{h,\xi}| + |\cR_{h,\xi}|)$. A union bound gives
    \begin{equation}\label{eq:union-d}
        \PP(E_N>x)\le C_{S,m}(1+t)^m d^{1-m/2}x^{-m}.
    \end{equation}
    On $\{E_N\le x\}$, Lemma~\ref{lem:stationary-drift} and discrete Gr\"onwall yield,
    \begin{equation}\label{eq:gronwall}
        \max_{n\le\tau_S^\theta\wedge N}\max_{a\le q}
        \norm{\theta_n^a}_\infty
        \le e^{L_St}(
        \max_a\norm{\theta_0^a}_\infty
        +\max_b\norm{\theta_*^b}_\infty
        +\max_c\norm{\mu^c}_\infty+x).
    \end{equation}
    Put $\Delta=\zeta_+-\zeta$, $\eta=(\zeta+\zeta_+)/2$, and $c_S=\Delta/(4\max\{1,L_S\})$. Taking $t=c_S\log d$ and $x=d^{-1/2+\eta}$ makes the right-hand side at most $d^{-1/2+\zeta_+}$ for large $d$. Since $\eta>0$, choosing $m>1/\eta$ in \eqref{eq:union-d} makes the failure probability $O((1+\log d)^m d^{1-m\eta})=o(1)$, proving \eqref{eq:thmC-prop}.
\end{proof}

\begin{corollary}\label{cor:deloc-interface}
    Suppose $\uv_d^{-1}(E_R)\subset B_{S(R)}$ for each $R$, where $S(R)$ is independent of $d$. Under the hypotheses of Theorem~\ref{thm:deloc}, with $A=A_d$ and $\tau_R=\inf\{n:\uv_d(\theta_n)\notin E_R\}$, for every fixed $R$ and $T$
    \[
        \PP(\tau_A\le\tau_R\wedge\lfloor T/\delta_d\rfloor)
        \rightarrow0\,.
    \]
\end{corollary}

\begin{proof}
    Before $\tau_R$ the parameter lies in $B_{S(R)}$, so $\tau_R\le\tau_{S(R)}^\theta$. For sufficiently large $d$, $T\le c_{S(R)}\log d$. The required probability is therefore bounded by the probability in \eqref{eq:thmC-prop}.
\end{proof}

\subsection{Verification of the ADAM dynamics}
\label{sec:proofs-gpm}

Put $E_R^*:=\uv_d^{-1}(E_R)$. We work on bounded sets $E_R$ of statistics and write $C_R$ for constants uniform on $E_R^*$. The statistics have bounded first and second derivatives, with zero third derivatives. In particular, $\nabla_{\theta^c}G_{ab} =\Id_{\{c=a\}}v^b+\Id_{\{c=b,\,b\le q\}}\theta^a$ with $\opnorm{\nabla^2G_{ab}}\le2$. The $w$ coordinates are linear and thus have derivative one. Lemma~\ref{lem:moments-gauss} verifies the gradient assumptions. We give the remaining projection and closure estimates explicitly.

\subsubsection{Projected direction}

We prove the stronger projection bound: for every deterministic $h=(h^1,\ldots,h^q,h^w)$ and every fixed $m\ge2$,
\begin{equation}\label{eq:gpm-strong-proj}
    \sup\nolimits_{\theta\in E_R^*}
    \norm{\langle h,\rho\rangle}_{L^m(\pi_\theta)}
    \le C_{R,m}\norm h.
\end{equation}
To prove this, choose an orthonormal basis $U\in\RR^{d\times\tilde r}$ for the span of the tracked vectors, where $\widetilde r\le r$. By rotational invariance, decompose independent samples as
\begin{equation}\label{eq:gpm-orth-decomp}
    X_t=\mu^{J_t}+\sigma U\gamma_t
    +\sigma(I-UU^\top)\eta_t,
\end{equation}
with independent standard Gaussian $\gamma_t\in\RR^{\widetilde r}$ and $\eta_t\in\RR^d$. The coefficients $\alpha_{t,a}$ depend only on $\mathcal H_\alpha=\sigma(J_t,\gamma_t:t\in\ZZ)$ and are independent of $(\eta_t)$. Compare the coordinate gradients with $\widehat g_{t,ai}=\sigma\alpha_{t,a}\eta_{t,i}$, and let $D_{ai}$ and $\widehat D_{ai}$ be their stationary ADAM directions. With $u_i=U^\top e_i$,
\[
    X_{t,i}-\sigma\eta_{t,i}
    =u_i^\top(U^\top\mu^{J_t}+\sigma\gamma_t-\sigma U^\top\eta_t).
\]
Gaussian moments and \eqref{eq:direction-diff} imply $\norm{D_{ai}-\widehat D_{ai}}_{L^m}\le C_{R,m}\norm{u_i}$. Hence
\begin{equation}\label{eq:gpm-proj-diff}
    \|\sum\nolimits_{a,i}h_i^a(D_{ai}-\widehat D_{ai})\|_{L^m}
    \le C_{R,m}\sum\nolimits_i\norm{h_i}\norm{u_i}
    \le C_{R,m}\sqrt r\,\norm h.
\end{equation}
Conditional on $\mathcal H_\alpha$, the quantities $H_i=\sum_a h_i^a\hat D_{ai}$ are independent across $i$ and centered by sign symmetry. Conditional Burkholder and then Minkowski give
\[
    \norm{\sum H_i}_{L^m}^2
    \le C_m\norm{\sum H_i^2}_{L^{m/2}}
    \le C_m\sum\norm{H_i}_{L^m}^2
    \le C_{R,m}\sum\norm{h_i}^2.
\]
The second layer has fixed dimension and bounded  moments. Combining with \eqref{eq:gpm-proj-diff} proves \eqref{eq:gpm-strong-proj}.

\subsubsection{The stationary mean drift}

For a physical coordinate $i$, put $v_i=(v_i^1,\ldots,v_i^r)$ and $b_i=\norm{v_i}$. On $E_R^*\cap A_d$, observe that $\sum_i b_i^2=\tr G\le C_R$ and $\max_i b_i\le C_Rd^{-1/2+\zeta_+}$.

\begin{lemma}\label{lem:gpm-mean-drift}
    For $a\le q$ and $b\le r$, $|\langle v^b,\pi_\theta\rho^a\rangle -F_{ab}(G(\theta),w)| \le C_R \| b_i \|$.
\end{lemma}

\begin{proof}
    Set $h_i=v_{i,1:\ell}$ and $\mu_i=(\mu_i^1,\ldots,\mu_i^k)$. Conditional on each label $J_t$, let $P_t^-\sim\cN(G_{\cdot,\ell+J_t}, \sigma^2(G-v_iv_i^\top))$ be independent of a standard scalar Gaussian $\xi_t$. Then the exact coordinate-gradient has the representation
    \[
        g_{t,ai}
        =\alpha_a(P_{t,1:\ell}^-+\sigma h_i\xi_t,w;y(J_t))
        (\mu_i^{J_t}+\sigma\xi_t).
    \]
    Next, define $P_t^{-} (s)$ by interpolating the covariance along $G-v_iv_i^\top+s v_iv_i^\top$, $0\le s\le1$ so integration by-parts yields for the truncated momentum series,
    $$
        \EE [ \tilde D_0^{a, (N)} (1 ) ] - \EE [ \tilde D_0^{a, (N)} ( 0 ) ] = \int^1_0 \frac{\sigma^2}{2}\sum\nolimits_{j=0}^N
        \EE\bigl[D^2_{P_{-j}}\tilde D_0^a[v_i,v_i]]  \dd s.
    $$
    Expanding the derivative yields the bound $\EE |D^2_{P_{-j}}\tilde D_0^a[v_i,v_i] | \lesssim_R  (\beta_1^j + \beta_2^j) \| v_i \|^2$, from which summing over $j$ and letting $N \to \infty$ proves that replacing covariance of each $P_t^-$ by $\sigma^2G$ changes $\pi_\theta \rho_{ai}$ by at most $O_R(b_i^2)$.

    With the full projection law now independent of $\xi$, parameterize the gradient as,
    \begin{equation}\label{eq:gpm-param-innov}
        g_a(h,\mu)=\alpha_a(P^{\mathrm{st}}+\sigma h\xi,w;y(J))
        (\mu^J+\sigma\xi),
    \end{equation}
    so that we may write the ADAM search direction as $f(h, \mu ) := \EE [\tilde D_0^a (h, \mu )]$. Taylor expanding at $(h,\mu)=0$ to first order yields an $O_R((\norm h+\norm\mu)^2)$ remainder, with $f(0, 0) = 0$ by sign symmetry under the expectation. Further multiplying by $v_i^b$ and summing over the index $i$ yields,
    \begin{equation}\label{eq:gpm-contracted-sum}
            \langle v^b,\pi_\theta\rho^a\rangle
            =\sum_{j\ge0}\EE[
            \partial_{g_{-j}^a}\tilde D_0^a
            \{\alpha_{-j,a}G_{b,\ell+J_{-j}}
            +\sigma^2\xi_{-j}^2\sum_{c\le\ell}
            G_{bc}\partial_{z_c}\alpha_{-j,a}\}
            ]
            +O ( \| b_i \|).
    \end{equation}
    We may further reduce the expectation using two integration by parts identities. First, observe that $\partial_{\xi_{-j}}\tilde D_0^a =\sigma\alpha_{-j,a}\partial_{g_{-j}^a}\tilde D_0^a$. Second, conditional on $J_{-j}$ and all other times, we may also write $\partial_{P_{-j,c}}\tilde D_0^a =\sigma\xi_{-j}\partial_{z_c}\alpha_{-j,a} \partial_{g_{-j}^a}\tilde D_0^a$. Thus Gaussian integration by parts in $\xi_{-j}$ and $P_{-j}$ respectively (e.g. Equation (5) of \cite{Mamis03042026}) therefore gives, for the sum in \eqref{eq:gpm-contracted-sum},
    \begin{equation}\label{eq:gpm-stein-id}
            \frac1\sigma\sum_{j\ge0}
            \EE[\xi_{-j}\tilde D_0^a
            \{G_{b,\ell+J_{-j}}
            +P_{-j,b}-G_{b,\ell+J_{-j}}\}] = \frac1\sigma\sum_{j\ge0}
            \EE[\xi_{-j}P_{-j,b}\tilde D_0^a]
            =F_{ab}(G,w),
    \end{equation}
    where the last equality holds by expanding $M_0^a$ in the numerator and observing that off-diagonal terms vanish by sign symmetry.
\end{proof}

\subsubsection{The remaining closure terms}

Write $a=\sP_\theta K_\theta\rho$ for corrector. For each statistic $u$, its pre-limit drift is
\begin{equation}\label{eq:mixed-psi-prelimit}
    \Psi_d^u(\theta)
    =-\nabla u\cdot\pi_\theta\rho
    +\delta\pi_\theta\left[
    \frac12\nabla^2u[\rho,\rho]+\nabla^2u[\rho,a]\right]
    +\delta T_{\nabla u}(\theta),
    \qquad
    T_V(\theta)=\pi_\theta[\langle V,D_\theta a[\rho]\rangle].
\end{equation}
For coordinate $(a,i)$, let $\Pi_i$ be the orthogonal projection that projects every student perturbation $h$ onto $\operatorname{span}\{v^1,\ldots,v^r,e_i\}$ and acts as the identity on $h^w$. If $h = (h^1, \ldots , h^q , h^w ) \in \ker (\Pi_i )$, every $\langle h^b,X\rangle$ is centered and independent of $(P_\theta (X), X_i , J )$. Therefore, by towering we have that $D_h K_\theta \varphi (x_{ai} ) = 0$ and similarly $D_h \pi_\theta = 0$. By Lemma~\ref{lem:poisson-first-deriv-pde}, we have $D_h a_{ai}=0$, from which we may write (using $(I - \Pi_i )\rho \in \ker (\Pi_i )$),
\[
    |D_\theta a_{ai}(\theta,x_{ai})[\rho]| = |D_\theta a_{ai}(\theta,x_{ai})[\Pi_i \rho]|
    \le C_R(1+\norm{x_{ai}}_2)\norm{\Pi_i\rho}.
\]
The rank of $\Pi_i$ is bounded, so \eqref{eq:gpm-strong-proj} and H\"older give a uniform $L^1(\pi_\theta)$ bound. The second-layer corrector has the same property without $e_i$ in the subspace. Consequently, multiplying by $|V_{ai}|$ with bounded norm and summing over $(a, i)$ yield,
\begin{equation}\label{eq:gpm-Tv-bound}
    \delta|T_V(\theta)|
    \le C_R\delta(\sum\nolimits_{a,i}|V_{ai}|+\norm{V_w})
    \le C_Rd^{-1/2}\norm V \to 0.
\end{equation}

For the quadratic correction, let $J_{ab}^{(i)} = \pi_\theta [\rho_{ai}\rho_{bi}+\rho_{ai}a_{bi}+\rho_{bi}a_{ai} ]$. Its representative version at zero coordinate parameters equals $\cD_{ab}(G,w)$ by the centered Neumann series and sign symmetry.From the coupling in the proof of Lemma~\ref{lem:gpm-mean-drift}, replace $P_t^-+\sigma v_i\xi_t$ by $P_t^-+\sigma v_i\xi_t'$ with $\xi_t'$ an independent copy of $\xi_t$. Interpolate between the two using $P_t^{-1} (s) := P_t^- + s \sigma v_i \xi_t + (1 - s) \sigma v_i\xi_t'$ and repeat the same arguments (now using Corollary~\ref{cor:poisson-w3-scale-uniform} to bound the derivatives), to obtain the estimate along the integrated path $| J_{ab}^{(i)} (1) - J_{ab}^{(i)} (0) | = | J_{ab}^{(i)}-\cD_{ab}(G,w) |\le C_R b_i$. Therefore, summing over $i$ and dividing by $d$ allows us to conclude that,
\begin{equation}\label{eq:gpm-J-avg}
    |(1/d)\sum\nolimits_iJ_{ab}^{(i)}-\cD_{ab}(G,w)|
    \le(C_R/d)\sum\nolimits_i b_i\le C_Rd^{-1/2}.
\end{equation}

\subsection{Verification of the matrix-normalized dynamics}

We use the two maps $\mathsf R$ specified above. Their compositions $B(N)=\mathsf R(N^\top N)$ have bounded first two derivatives, which following from Lemma~\ref{lem:muon-derivatives}--Lemma~\ref{lem:muon-ns5-derivatives} for Exact MUON and the NS5 algorithm respectively. We establish all of the conditions for Theorem~\ref{thm:main-ballistic} with $A_d = \RR^d$ (i.e. no coordinate delocalization is required). The stationary first-layer momentum is
\begin{equation}\label{eq:muon-stat-momentum}
    \tilde M_t=\sum\nolimits_{j\ge0}\omega_j^M X_{t-j}\alpha_{t-j}^\top,
    \qquad
    \rho_\theta(\tilde M_t)=\tilde M_t\widehat R_t,
    \qquad \widehat R_t=\mathsf R(\tilde M_t^\top\tilde M_t/d).
\end{equation}
For a deterministic matrix $H$ with a fixed number of columns, Gaussian projection moments give
\begin{equation}\label{eq:muon-proj-moment-reuse}
    \norm{\Fnorm{H^\top\tilde M_t}}_{L^m}
    \le C_{R,m}\Fnorm H,
    \qquad
    \norm{\langle H,\rho_\theta(\tilde M_t)\rangle_F}_{L^m}
    \le C_{R,m}\Fnorm H
\end{equation}
when $H$ has $q$ columns in the second inequality. Indeed the $(ab)$-th entry of $H^\top\tilde M_t$ is given by the sum $(H^\top \tilde M_t )_{ab} = \sum_{j \ge 0}\omega_j^M\alpha_{t-j,a}\langle h^b,X_{t-j}\rangle$, from which Holder and Minkowski proves the first inequality when summing over $(a, b)$. The second inequality follows from $|\tr(H^\top\tilde M_t\widehat R_t)|\le\sqrt q\,\opnorm{\widehat R_t}\Fnorm{H^\top\tilde M_t}$ and the operator norm bound on $\mathsf R$. The second layer has fixed dimension and bounded ADAM moments. These observations verify localizability on $E_L$.

Note that the mixed corrector has the block decomposition,
\begin{equation}\label{eq:mixed-corrector-decomp}
    \chi_{1,\theta}^u(x)
    =-\langle\nabla_{\theta^{1:q}}u,a_\theta(\theta,M)\rangle_F
    -\langle\nabla_wu,a_w(\theta,x_w)\rangle,
\end{equation}
where the first block has scale $\sqrt{qd}$ and the scalar second-layer blocks have unit scale. The bounds from Lemma~\ref{lem:adam-correctors} are then,
\begin{equation}\label{eq:mixed-a-w3-bound}
    (qd)^{-1/2}\norm{a_\theta}_{\cW_{1,\sqrt{qd}}^3(E_R^*)}
    +\max_{\nu\le s}\norm{a_{w_\nu}}_{\cW_{1,1}^3(E_R^*)}
    \le C_R.
\end{equation}
The pre-limit drift again has the form \eqref{eq:mixed-psi-prelimit}.

\begin{lemma} \label{lem:muon-gram-conc}
    For every fixed $m\ge2$, uniformly on $E_L$ and in $t\in\ZZ$,
    \begin{equation}\label{eq:muon-gram-Lr}
        \norm{\opnorm{\tilde M_t^\top\tilde M_t/d-C_t}}_{L^m}
        \le C_{R,m}d^{-1/2},\qquad
        \norm{\opnorm{\widehat R_t-R_t}}_{L^m}
        \le C_{R,m}d^{-1/2}.
    \end{equation}
    Consequently, for every $b\le r$, $\|\pi_\theta[\rho_\theta(\tilde M_0)^\top v^b] -F_{\cdot,b}^{\mathsf R}(G,w)\|_2 \le C_Rd^{-1/2}$.
\end{lemma}

\begin{proof}
    Recall the Gaussian decomposition \eqref{eq:gpm-orth-decomp} and write $\tilde M_t=M_t^\parallel+M_t^\perp$ corresponding to the decomposition of $X^{\parallel}$ and $X^{\perp}$ according to the range of $U$ and its orthogonal complement respectively. The parallel term is measurable with respect to $\mathcal H_\alpha$, and \eqref{eq:muon-proj-moment-reuse} with $H=U$ gives
    \begin{equation}\label{eq:muon-parallel-small}
        \norm{\opnorm{(M_t^\parallel)^\top M_t^\parallel/d}}_{L^m} \le
        d^{-1}\norm{\Fnorm{M_t^\parallel}^2}_{L^{m}} \le d^{-1}\norm{\Fnorm{M_t^\parallel}}_{L^{2m}}^2
        \le C_{R,m}d^{-1}.
    \end{equation}
    In an orthonormal basis of $\text{span} (U )^\perp$, the rows of $M_t^\perp$ are conditionally independent $\cN(0,C_t)$ vectors, given $\mathcal H_\alpha$. Writing these rows as the vectors $\zeta_1,\ldots,\zeta_{d-\widetilde r}$, then
    \[
        (M_t^\perp)^\top(M_t^\perp)/d-C_t
        =(1/d)\sum
        (\zeta_j\zeta_j^\top-C_t)
        -(\tilde r/d)C_t.
    \]
    For the first term, fix any matrix entry $(a, b) \in [q] \times [q]$ and apply conditional Burkholder with $\cH_\alpha$ with $W_{\ell, ab} = \zeta_{\ell , a} \zeta_{\ell , b} - C_{\ell , ab}$ to write,
    $$
        \EE [ (1/d) \sum\nolimits^{d - \tilde{r}}_{\ell = 1} W_{\ell, ab} \mid \cH_\alpha] ^{2 / m} \le Cd^{-2}\sum\nolimits^{d - \tilde{r}}_{\ell = 1} \| W_{\ell, ab} \|^2_{L^m (\cdot \mid \cH_\alpha )} \le (C_m'/d) \| C_t \|^2_{\text{op}},
    $$
    after which taking the power $m / 2$ and unconditional expectations gives an $L^m$ bound $C_m d^{-1/2}\norm{\opnorm{C_t}}_{L^m}$. Together with \eqref{eq:muon-parallel-small}, this proves the first assertion. The Lipschitz property of $\mathsf R$ proves the second. Finally, $\rho_\theta(\tilde M_0)^\top v^b=\widehat R_0\tilde M_0^\top v^b$. Replacing $\widehat R_0$ by $R_0$ costs $O_R(d^{-1/2})$ by Cauchy--Schwarz and \eqref{eq:muon-proj-moment-reuse} applied to $v^b$. The remaining expectation equals \eqref{eq:gpm-matrix-F} exactly.
\end{proof}

\begin{lemma} \label{lem:muon-autocovariance}
    For the matrix-normalized rules, uniformly on $E_L$,
    \begin{equation}\label{eq:muon-D-closure}
        \|(c_\delta/d)\pi_\theta[
        \rho_\theta^\top\rho_\theta
        +\rho_\theta^\top a_\theta+a_\theta^\top\rho_\theta]
        -c_\delta\cD^{\mathsf R}(G,w)\|_{\mathrm{op}}
        \le C_Rd^{-1/2}.
    \end{equation}
\end{lemma}

\begin{proof}
    Let $\bar\rho_\theta=\pi_\theta\rho_\theta$, and set $\Gamma_h^{(d)} =\EE[\rho_\theta(\tilde M_0)^\top\rho_\theta(\tilde M_h)] -\bar\rho_\theta^\top\bar\rho_\theta$ for $h \ge 1$. Expanding the corrector, we may write
    \[
        \pi_\theta[
        \rho_\theta^\top\rho_\theta
        +\rho_\theta^\top a_\theta+a_\theta^\top\rho_\theta] = \EE[\rho_\theta(\tilde M_0)^\top
        \rho_\theta(\tilde M_0)]
        +\sum
        (\Gamma_h^{(d)}+(\Gamma_h^{(d)})^\top).
    \]
    First, observe that for the orthogonal Gaussian noises $(\eta_t )_{t \in \ZZ}$ that coincide with $M_0^\perp$, we have that $M_0^{\parallel}$ is invariant to sign flips (as it does not depend on $\eta_t$) so that $\widehat{R}_0 = R (\tilde M_0^\top \tilde M_0 / d )$ is invariant  under $\eta \mapsto - \eta$. Consequently, $\EE [M_0^\perp \widehat{R}_0] = 0$ implying that $\bar \rho_\theta = UU^\top \EE [ \rho_\theta (\tilde M_0 )]$ lies in the range of $U$. Therefore \eqref{eq:muon-proj-moment-reuse} implies $d^{-1}\Fnorm{\bar\rho_\theta}^2\le C_Rd^{-1}$.

    The lag-zero estimate follows directly from \eqref{eq:muon-gram-Lr}. For $h\ge1$, unroll $\tilde M_h=\beta_M^h\tilde M_0+M_{0,h}$, where $M_{0,h}=\sum_{j=0}^{h-1}\omega_j^M X_{h-j}\alpha_{h-j}^\top$. Furthermore, we write the estimate,
    $$
        d^{-1} \EE [ \rho_\theta (\tilde M_0 )^\top \rho_\theta (\tilde M_h ) ] = \beta^h_M \EE [ \widehat{R}_0 ( \tilde M_0^\top \tilde M_0/d ) \widehat{R}_h] + \EE [ \widehat{R}_0 ( \tilde M_0^\top M_{0, h}/d )  \widehat{R}_h ].
    $$
    For the first term, exactly as in the previous step apply \eqref{eq:muon-gram-Lr} with Cauchy-Schwarz to obtain $\beta_M^h\EE[R_0C_0R_h] + O (\beta_M^h d^{-1/2} )$. For the cross term, use the conditional decomposition given $\mathcal H_\alpha$ to write $\tilde M_0^\top M_{0, h} = (M_0^{\parallel} )^\top M_{0, h}^{\parallel} + (M_0^{\perp} )^\top M_{0, h}^{\perp}$. For the parallel term, observe that,
    $$
        \| \EE [ \widehat{R}_0 ( (  M_0^\parallel )^\top ( M_{0, h}^\parallel )/d )  \widehat{R}_h ] \|_{\text{op}} \le (C/d) \| \| M_0^\parallel \|_F \|_{L^2} \, \| \| M_{0, h}^\parallel \|_F \|_{L^2} \le C/d\,.
    $$
    For the orthogonal term, we similarly decompose it as,
    $$
        \EE [ \widehat{R}_0 ( (  M_0^\perp )^\top ( M_{0, h}^\perp )/d )  \widehat{R}_h ] = \EE [ \widehat{R}_0 ( (  M_0^\perp )^\top ( M_{0, h}^\perp )/d )  R_h ] + \EE [ \widehat{R}_0 \left( (  M_0^\perp )^\top ( M_{0, h}^\perp )/d \right)  (\widehat{R}_h - R_h ) ].
    $$
    The first term vanishes by sign symmetry of the orthogonal Gaussian variables $(\eta_t )_{1 \le t \le h}$. For the second term, observe that the rows of $M_0^\perp$ and $M_{0, h}^\perp$ are conditionally independent mean-zero Gaussians, thus writing $(M_0^\perp)^\top M_{0,h}^\perp = \sum_{\ell = 1}^{d - \tilde r} \zeta_{\ell, 0} \zeta_{\ell, h}^\top$ we have that $\norm{\opnorm{(M_0^\perp)^\top M_{0,h}^\perp/d}}_{L^2} \le C_Rd^{-1/2}$. Thus by Cauchy-Schwarz the second expectation is also bounded by $O (d^{-1} )$ by \eqref{eq:muon-proj-moment-reuse}. Combining the two terms yields the bound,
    \[
        \|d^{-1}\Gamma_h^{(d)}
        -\beta_M^h\EE[R_0C_0R_h]\|_{\mathrm{op}}
        \le C_R(\beta_M^hd^{-1/2}+d^{-1}).
    \]
    Finally, to sum the lag errors, first use the coupling argument with a second stationary chain $\tilde M_h '$ initialized from $\tilde M_0 '$ to obtain the apriori estimate for $\Gamma_h^{(d)} = \EE [\rho_\theta (\tilde M_0 )^\top ( \rho_\theta (\tilde M_h ) - \rho_\theta (\tilde M_h ' ) )]$ which is $d^{-1}\opnorm{\Gamma_h^{(d)}}\le C_R\beta_M^h$. Truncate at $L_d=\lceil(1-\beta_M)^{-1}\log d\rceil$ so that the total error is bounded by,
    \[
        \sum\nolimits_{h \ge 1} \|d^{-1}\Gamma_h^{(d)}
        -\beta_M^h\EE[R_0C_0R_h]\|_{\op} \le \sum\nolimits^{L_d}_{h = 1} ( \beta_M^h d^{-1/2} + d^{-1} ) + \sum\nolimits_{h > L_d} ( \frac{1}{d} \| \Gamma_h^{(d)} \|_{\op} + \beta_M^h \| \EE[R_0C_0R_h] \|_{\op} ),
    \]
    with the right hand side of order $O(d^{-1/2} )$, completing the proof.
\end{proof}

\begin{lemma} \label{lem:muon-Tv-bound}
    For every deterministic $V$ with $\norm V\le C_R$, the mixed matrix--ADAM rule satisfies $\delta_d\left|\pi_\theta[ \langle V,D_\theta a(\theta,x)[\rho(x)]\rangle] \right|\le C_Rd^{-1/2}$.
\end{lemma}

\begin{proof}
    It suffices to prove the bound for the first layer. Put $B(N)=\mathsf R(N^\top N)$, so $\rho_\theta(M)=MB(M/\sqrt d)$. Differentiating the first two derivatives of $B$ and subsequently bounding the quantities term-wise yields,
    \begin{equation}\label{eq:muon-V-grad-bounds}
        \begin{aligned}
            |\langle V,D\rho_\theta(M)[H]\rangle_F|
            &\le C\left(\Fnorm{V^\top H}
            +\Fnorm{V^\top M}\Fnorm{H}/\sqrt d\right),\\
            |\langle V,(D\rho_\theta(M)-D\rho_\theta(M'))[H]\rangle_F|
            &\le C\biggl(
            \Fnorm{V^\top H}\frac{\Fnorm{M-M'}}{\sqrt d} +\Fnorm{V^\top(M-M')}\frac{\Fnorm H}{\sqrt d}\\
            &\qquad+\Fnorm{V^\top M'}
            \frac{\Fnorm{M-M'}}{\sqrt d}
            \frac{\Fnorm H}{\sqrt d}\biggr).
        \end{aligned}
    \end{equation}
    Fix a state $x$ and direction $h$. Couple the chain $M_n^x$ to a stationary chain $\tilde M_n$ so that $M_n^x-\tilde M_n=\beta_M^n(x-\tilde M_0)$ and furthermore,
    \[
        H_n:=D_hM_n^x
        =\sum\nolimits_{j=1}^n
        (1-\beta_M)\beta_M^{n-j}
        D_hg_\theta(\theta,Y_j),
    \]
    while $D_h\tilde M_n=H_n+\beta_M^nD_h\tilde M_0$. Differentiate the Neumann series at fixed $x$ to obtain
    \begin{equation}\label{eq:muon-Dh-atheta-series}
            \langle V,D_ha_\theta(\theta,x)\rangle_F
            =\sum\nolimits_j
            \EE[
            \langle V,(D\rho_\theta(M_n^x)
            -D\rho_\theta(\tilde M_n))[H_n]\rangle_F -\beta_M^n
            \langle V,D\rho_\theta(\tilde M_n)
            [D_h\tilde M_0]\rangle_F].
    \end{equation}
    Since $D_hg_\theta(\theta,Y)=X(D_h\alpha(\theta,Y))^\top$, Gaussian moment bounds give, for every fixed $m$, the projection bounds $\norm{\Fnorm{V^\top H_n}}_{L^m} \le C_{R,m}\norm h$ and $\norm{\Fnorm{H_n}/\sqrt d}_{L^m} \le C_{R,m}\norm h$ respectively, as well as the same bounds for $D_h\tilde M_0$ in place of $H_n$. Furthermore $\norm{\Fnorm{V^\top\tilde M_n}}_{L^m} +\norm{\Fnorm{\tilde M_n}/\sqrt d}_{L^m}\le C_{R,m}$. Substituting these bounds into \eqref{eq:muon-V-grad-bounds}, each summand in \eqref{eq:muon-Dh-atheta-series} is at most $C_R\beta_M^n\norm h (1+\Fnorm x/\sqrt d+\Fnorm{V^\top x} )$ in expectation. Summing over the geometric series yields the pointwise bound,
    \[
        |\langle V,D_ha_\theta(\theta,x)\rangle_F|
        \le C_R\norm h
        (1+\Fnorm x/\sqrt d+\Fnorm{V^\top x}).
    \]
    Now set $h=\rho(x)$, and integrate with respect to $\pi_\theta$ and apply Cauchy-Schwarz. The first norm satisfies $\| \rho \|_{L^2 (\pi )} \le O (\sqrt d)$, while the other factor is $O(1)$ by \eqref{eq:muon-proj-moment-reuse}. Together, the first-layer projection is bounded by $C_R\sqrt d$ from which multiplying by $\delta$ completes the proof.
\end{proof}

\section{Results for Multiclass Logistic Regression}
\label{sec:multi-class-log-reg-model}

Consider Example~\ref{ex:app-mlr} with $q \ge 2$, $k \ge 1$, and critical scaling $\delta = c_\delta / d$, and optional decoupled weight decay $\kappa \theta$ for $\kappa \ge 0$. Writing $s_a(z)=e^{z_a}/\sum_{b=1}^q e^{z_b}$ and $s(z)=(s_a(z))_{a\le q}$, the gradient and Hessian with respect to the logits are
\begin{equation}\label{eq:mlr-alpha-hess}
    \alpha(z;y)=s(z)-y,\qquad
    H(z)=\diag(s(z))-s(z)s(z)^\top.
\end{equation}
This is a Gaussian projective model with $q_*=s=0$ and $\sigma=\lambda^{-1/2}$ where the Gram matrix decomposes as $Q=\theta^\top\theta$ and $M=(m^1,\ldots,m^k)$ with $m^j=\theta^\top\mu^j$ (with the other components being static). Use the representative projections of \eqref{eq:gpm-rep-proj}, with $r=q+k$, and write $z_t=P_{t,1:q}$ and $\alpha_t=\alpha(z_t;y(J_t))$. For ADAM, the representative coordinate gradient is $g_t^a=\lambda^{-1/2}\alpha_{t,a}\xi_t$; its stationary processes are those in \eqref{eq:gpm-rep-adam}, with regularizer $\epsilon_A>0$. We denote the corresponding coefficients in \eqref{eq:gpm-F-def}--\eqref{eq:gpm-D-def} by $(F^A,\cD^A)$.

\begin{corollary}[ADAM Dynamics - Logistic Regression]\label{cor:mlr-adam-sgd}
    Under the setting of Proposition~\ref{prop:adam-gpm-dynamics}, fix $c_\delta>0$, $\kappa\ge0$, and fix $\beta_1,\beta_2\in[0,1)$ and $\epsilon_A>0$. If $G(\theta_0)\Rightarrow\nu_0$, then the interpolated Gram process converges to the solution, initialized from $G_0\sim\nu_0$, of
    \begin{equation}\label{eq:mlr-adam-ode}
            \dot Q=-F_{1:q,1:q}-F_{1:q,1:q}^\top+c_\delta\cD-2\kappa Q,\qquad
            \dot m^j=-F_{\cdot,q+j}-\kappa m^j,\qquad 1\le j\le k,
    \end{equation}
    For ADAM, $(F,\cD)=(F^A,\cD^A)$. For online SGD,  $F_{\cdot,b}^{\mathrm{SGD}}=\EE[\alpha_0P_{0,b}]$ and $\cD^{\mathrm{SGD}}=\lambda^{-1}\EE[\alpha_0\alpha_0^\top]$.
\end{corollary}

\begin{proof}
    Proposition~\ref{prop:adam-gpm-dynamics} gives the ADAM coefficients, and the SGD dynamics follow from \cite{BenArousEtAl2025Spectral}.
\end{proof}

For the matrix optimizers, let $b_j=(1-\beta_M)\beta_M^j$ and define (with $R_K$ defined in \eqref{eq:ns5-R5-map}),
\begin{equation}\label{eq:mlr-muon-CR}
    C_t=\lambda^{-1}\sum\nolimits_{j\ge0}b_j^2\alpha_{t-j}\alpha_{t-j}^\top,
    \qquad R_t=(C_t+\epsilon_M I_q)^{-1/2},
    \qquad R_t^{\mathrm{NS}K}=R_K(C_t),
\end{equation}

\begin{proposition}\label{prop:mlr-muon-dynamics}
    Fix $\beta_M\in[0,1)$ and positive regularizers, and let $K$ be fixed for MUON--NS$K$. Under the initialization hypotheses of Proposition~\ref{prop:gpm-matrix-dynamics}, the Gram process for Exact MUON or MUON--NS$K$, with optional decoupled decay $\kappa\theta$, converges to \eqref{eq:mlr-adam-ode}. The coefficients are those of \eqref{eq:gpm-matrix-F}--\eqref{eq:gpm-matrix-D}, with $\alpha_t=s(z_t)-y(J_t)$, $\sigma=\lambda^{-1/2}$, and respectively $R_t$ or $R_t^{\mathrm{NS}K}$ from \eqref{eq:mlr-muon-CR}.
\end{proposition}

\begin{proof}
    Apply Proposition~\ref{prop:gpm-matrix-dynamics} with $q_*=s=0$.
\end{proof}

As a projective model, recall that the population risk (with respect to the cross-entropy) is closed with respect to the Gram matrix. The population cross-entropy depends on the parameter only through its Gram matrix. Furthermore, we have as an immediate consequence of Lemma~\ref{lem:gaussian-risk-chain-rule} and \eqref{eq:mlr-alpha-hess} the following.

\begin{proposition}\label{prop:mlr-pop-risk-ode}
    For each of the dynamics above, the population risk dynamics converges to,
    \begin{equation}\label{eq:mlr-risk-ode}
       \dot\cR(G_t)
        =\sum\nolimits_{j=1}^k\mathsf p_j\,
        \EE[\alpha(Z;y(j))\cdot\dot m^j
        +(2\lambda)^{-1}\tr\bigl(H(Z)\dot Q\bigr)
        \,|\,J=j].
    \end{equation}
\end{proposition}

We now examine the limiting dynamics near initialization. Assume uniform mixture weights $\mathsf p_j=1/k$ and orthonormal means, and write for a direction $A=(a^1,\ldots,a^k)\in\RR^{q\times k}$,
\begin{equation}\label{eq:mlr-initial-gram}
    G_0(\tau,m_0)=
    \begin{pmatrix}\tau^2I_q&m_0A\\m_0A^\top&I_k\end{pmatrix}.
\end{equation}
Here $\tau\ge0$ is the initialization radius and $m_0\ge0$ is chosen so that the matrix is positive semidefinite. The warm-start expansion below takes $m_0\downarrow0$ at fixed $\tau>0$. Set the uniform probability vector $u=q^{-1}\one_q$ and projection $\Pi=I_q-q^{-1}\one_q\one_q^\top$, and define
\begin{equation}\label{eq:mlr-h-lambda-def}
    h(\tau,\lambda)
    :=(1-\EE\norm{s(Z)}_2^2)/(q-1),
    \qquad Z\sim\cN(0,\tau^2\lambda^{-1}I_q).
\end{equation}
Then $0<h(\tau,\lambda)\le1/q$, $h(0,\lambda)=1/q$, and we abbreviate $h(\lambda):=h(1,\lambda)$. In the following, $F^{\mathrm{alg}}$ and $\cD^{\mathrm{alg}}$ without an argument are evaluated at $G_0(\tau,0)$.

\begin{lemma}\label{lem:mlr-init-master}
    At $G_0(\tau,0)$, the initial population-risk derivative for SGD, ADAM, Exact MUON, or MUON--NS$K$ is
    \begin{equation}\label{eq:mlr-proof-master-slope}
            \dot\cR_{\mathrm{alg}}(0)
            =-\frac1k\sum_{j=1}^k(u-y(j))\cdot F_{\cdot,q+j}^{\mathrm{alg}}
            -\frac{h(\tau,\lambda)}{\lambda}(
            \tr(\Pi F_{1:q,1:q}^{\mathrm{alg}})
            -(c_\delta/2)
            \tr(\Pi\cD^{\mathrm{alg}})
            +\kappa\tau^2(q-1)).
    \end{equation}
    For fixed $\tau>0$, positive algorithm regularizers, and the warm initialization \eqref{eq:mlr-initial-gram}, the same right-hand side gives $\dot\cR_{\mathrm{alg}}(0;m_0) + O(m_0)$.
\end{lemma}

\begin{proof}
    At $m_0=0$, the student projection $Z$ is independent of $J$ and has the exchangeable law in \eqref{eq:mlr-h-lambda-def}. Hence $\EE[s(Z)]=u$ since $\EE [s_a (Z) ] = 1/q$ using $\sum_a s_a (Z) = 1$. Next, since the softmax function is permutation equivariant, i.e. $s (P_\sigma z ) = P_\sigma s (z)$ for a permutation $\sigma \in S_q$, we observe that $H (P_\sigma z ) = P_\sigma H (z) P_\sigma^\top$ from which we see that $\EE[H(Z)]$ commutes by exchangeability of $Z$ and thus annihilates $\one_q$, i.e. $\EE[H(Z)] \one_q = 0$. Together, this implies that $\EE[H(Z)] = h \Pi$ for some scalar $h$. Taking traces gives exactly \eqref{eq:mlr-h-lambda-def} from $1 - \EE \| s (Z) \|^2 = h (q - 1)$. Strict positivity follows from $\norm{s(Z)}_2^2<1$, and $h\le1/q$ follows from $\norm{s(Z)}_2^2\ge1/q$.

    Substitute $\EE[\alpha(Z;y(j))]=u-y(j)$, $\EE H(Z)=h\Pi$, and \eqref{eq:mlr-adam-ode} into \eqref{eq:mlr-risk-ode} and use symmetry of $\Pi$ to obtain the dynamics. For $m_0 > 0$, use the fact that the derivatives of $\alpha$ and $H$ are bounded with Taylor's theorem to write $\EE [\alpha (m_0 a^j + Z ; y(j) ] = u - y(j) + O (m_0 )$ and $\EE [ H (m_0 a^j + Z ; y(j) ] = h (\tau , \lambda ) \Pi + O (m_0 )$. Use the fact that $(F^{\text{alg}} (G) , \cD^{\text{alg}} (G)$ are locally Lipschitz, and evaluate at $G_0 (\tau , m_0 )$ instead of $G_0 (\tau , 0)$ together with the regularization coefficient $-\kappa m^j=O(m_0)$ to complete the proof.
\end{proof}

\begin{proposition}[Online SGD - Initial Descent]\label{prop:mlr-desc-master}
    Under the setup above, write $h=h(\tau,\lambda)$. Then
    \begin{equation}\label{eq:mlr-sgd-tau1-slope-proof}
        \dot\cR_{\mathrm{SGD}}(0)
        =(q-1)[-(qk)^{-1}-\lambda^{-1}\kappa\tau^2h-\lambda^{-2}(\tau^2h^2+c_\delta h(2/q-h)/2)]
    \end{equation}
    Consequently, the initial derivative is negative, zero, or positive according as $c_\delta$ is smaller than, equal to, or larger than the critical value,
    \begin{equation}\label{eq:mlr-sgd-general-desc}
        c_{\delta,\mathrm{SGD}}^{\mathrm{desc}}(\tau,\lambda)
        =2[\lambda^2/(qk)+\tau^2h^2+
        \kappa\tau^2\lambda h]/(h(2/q-h))
    \end{equation}
\end{proposition}

\begin{proof}
    Since the means are orthonormal and $M_0=0$, the projection $P_{0,q+j}$ has conditional mean $\Id_{\{J_0=j\}}$ and its centred version is independent of the student projection $Z$. Thus substituting $P_{0, q+ j} = \Id_{\{ J_0 = j \}} + W_j$ into $F_{\cdot , q + j}^{\text{SGD}}$ gives,
        \(F_{\cdot,q+j}^{\mathrm{SGD}}=(u-y(j))/k\), and 
        \(\sum_{j=1}^k(u-y(j))\cdot F_{\cdot,q+j}^{\mathrm{SGD}}/k
        =(q-1)(qk)^{-1}\)
    where the second equality follows by recalling that $y(j)$ is a standard basis vector. Next, by Gaussian integration by parts, we may write,
    $F_{1:q,1:q}^{\mathrm{SGD}}
        =\EE[(s(Z)-y(J))Z^\top]
        =\frac{\tau^2}{\lambda}\EE H(Z)
        =\frac{\tau^2h}{\lambda}\Pi.$
    Finally, $\Pi\alpha=\alpha$ (recall $\alpha_0^\top \Id_q = 0$) and $\EE s(Z)=u$, so
    $\tr(\Pi\cD^{\mathrm{SGD}})
        =\lambda^{-1}\EE\norm{s(Z)-y(J)}_2^2
        =(q-1)(2/q-h)/\lambda.$
    Substituting these three identities into \eqref{eq:mlr-proof-master-slope} proves \eqref{eq:mlr-sgd-tau1-slope-proof}. Since $0<h\le1/q$, its coefficient with the $c_\delta$ term is positive. Thus solving for $c_\delta$ in $\dot{\cR}_{\text{SGD}} = 0$ yields \eqref{eq:mlr-sgd-general-desc}.
\end{proof}

For SGD-M, the SGD coefficients become $F/(1-\beta)$ and $\cD/(1-\beta)^2$, which can be applied above to obtain $c_{\delta,\mathrm M}^{\mathrm{desc}}$ for SGD-M.

\begin{proposition}[Binary zero-initialization slopes]\label{prop:mlr-binary-initial-slopes}
    Suppose $q=k=2$, $y(j)=e_j$, $Q_0=M_0=0$, and $\kappa=0$. For ADAM take $\beta_1\in[0,1)$, $\beta_2\in(0,1)$ and send $\epsilon_A\downarrow0$ after $d \to \infty$. With the constants $\kappa_1,\kappa_2$ of Corollary~\ref{cor:adam-pca-eps-zero},
    \begin{equation}\label{eq:mlr-binary-initial-slopes}
            \dot\cR_{\mathrm{SGD}}(0)=-\frac14+\frac{c_\delta}{8\lambda^2}, \qquad \dot\cR_A(0)=-\frac{\sqrt\lambda}{2}\kappa_1
            +\frac{c_\delta}{2\lambda}\kappa_2, \qquad 
            \dot\cR_M(0)=-\frac{r_M}{4}
            +\frac{c_\delta r_M^2}{8\lambda^2},
    \end{equation}
    with $r_M=((1 - \beta_M ) [2\lambda  (1 + \beta_M ) ]^{-1}
            +\epsilon_M )^{-1/2}$. Letting $\epsilon_M\downarrow0$ gives the thresholds in Proposition~\ref{prop:app-mlr-eos}.
\end{proposition}

\begin{proof}
    The SGD formula is \eqref{eq:mlr-sgd-tau1-slope-proof} with $\tau=0$ and $h=1/2$. For ADAM, $\alpha_{t,1}=-\alpha_{t,2}=(1-2\Id_{\{J_t=1\}})/2$. Hence
            \(g_t^1=-g_t^2=-\frac{1}{2\sqrt\lambda}Z_t\), and 
        \(Z_t=(2\Id_{\{J_t=1\}}-1)\xi_t
        \sim\cN(0,1)\) i.i.d.
    Denote the scalar Gaussian sums of Proposition~\ref{prop:adam-pca-dynamics} by $\Psf_t,\Qsf_t$. The stationary first and second moments are $M_t^1=-M_t^2=-\Psf_t/(2\sqrt\lambda)$ and $V_t^1=V_t^2=\Qsf_t/(4\lambda)$. Taking $\epsilon_A \downarrow 0$, equation~\eqref{eq:gpm-F-def} gives $F_{\cdot,2+j}^{A}=\sqrt\lambda\,\kappa_1(u-e_j)$, $F_{1:2,1:2}^{A}=0$, and $\cD^{A}=2\kappa_2\Pi$. Substituting these expressions into \eqref{eq:mlr-proof-master-slope} proves the ADAM formula.

    For Exact MUON, $\alpha_t\alpha_t^\top=\Pi/2$, so that we may write $C_t=C_*\Pi$ with $C_*:= (1 - \beta_M ) / (2 \lambda (1 + \beta_M ))$ and $\Pi R_t\Pi=r_M\Pi$. The coefficients in \eqref{eq:gpm-matrix-F}--\eqref{eq:gpm-matrix-D} therefore reduce to
    \[
        F_{\cdot,2+j}^{M}=r_M(u-e_j)/2,\qquad
        F_{1:2,1:2}^{M}=0,\qquad
        \cD^{M}=r_M^2C_*(1+2\sum\beta_M^h)\Pi
        =r_M^2\Pi/(2\lambda).
    \]
    A final substitution into \eqref{eq:mlr-proof-master-slope} gives the MUON slope.
\end{proof}

\begin{proof}[Proof of Proposition~\ref{prop:app-mlr-eos}]
    The general SGD threshold follows from \eqref{eq:mlr-sgd-general-desc}, with the specialization obtained by setting $\tau=1$. At binary zero initialization, the zeros of \eqref{eq:mlr-binary-initial-slopes} are respectively $2\lambda^2$, $\kappa_1 / \kappa_2 \lambda^{3/2}$, and $2\lambda^2 /r_M$.
    The coefficients of $c_\delta$ are positive, so these zeros separate initial descent from initial ascent.
\end{proof}

\section{Gaussian XOR Classification using Two-Layer Neural Networks}
\label{sec:gaus-xor-two-layer}

We now provide the effective dynamics description for the Gaussian XOR classification problem given in Example~\ref{ex:app-mlr} with $q\ge4$ and $p = q (d + 1)$. This is a Gaussian projective model with noise level $\lambda^{-1/2}$, $q_*=0$, $s=q$, and $k=4$. Because the negative mixture means are redundant, we instead track
\begin{equation}\label{eq:xor-tracked-vectors}
    v(\Theta)=(\theta^1,\ldots,\theta^q,\mu,\nu),\qquad
    \uv(\Theta)=(G(\Theta),w),\qquad G(\Theta)=v(\Theta)^\top v(\Theta).
\end{equation}
There are $\binom{q+1}{2}+2q+q$ statistics, which is $22$ when $q=4$. Throughout this section, we consider the critical scaling $\delta_d=c_\delta/d$ with $c_\delta>0$.

For an activation $\varphi:\RR\to\RR$, the derivatives of the cross-entropy loss are given by,
\begin{equation}\label{eq:xor-alpha-b}
    \begin{aligned}
        \alpha_a(z,w;y)&:=\partial_{z_a}\Upsilon(z,w;y)
        =\bigl(\sigma(f(z,w))-y\bigr)w_a\varphi'(z_a),\\
        b_a(z,w;y)&:=\partial_{w_a}\Upsilon(z,w;y)
        =\bigl(\sigma(f(z,w))-y\bigr)\varphi(z_a),\qquad 1\le a\le q.
    \end{aligned}
\end{equation}
In particular, the sample gradients are $\nabla_{\theta^a}L=X\alpha_a$ and $\partial_{w_a}L=b_a$. Given $(G,w)$, set $m^1(G)=G_{\cdot,q+1}$, $m^2(G)=-G_{\cdot,q+1}$, $m^3(G)=G_{\cdot,q+2}$, and $m^4(G)=-G_{\cdot,q+2}$.

Let $(J_t,P_t,\xi_t)_{t\in\ZZ}$ be i.i.d., with $J_t$ uniform on the four mixture components, $\xi_t\sim\cN(0,1)$ independent of $(J_t,P_t)$, and
    \(P_t\mid\{J_t=c\}\sim\cN\bigl(m^c(G),\lambda^{-1}G\bigr)\).
We abbreviate $z_t=P_{t,1:q}$, $\alpha_t=\alpha(z_t,w;y(J_t))$, and $b_t=b(z_t,w;y(J_t))$. The representative gradients are $g_t^a=\lambda^{-1/2}\alpha_{t,a}\xi_t$ for the first layer and $g_t^{w_a}=b_{t,a}$ for the second. All algorithms considered below have the same form of limiting equations:
\begin{equation}\label{eq:xor-adam-ode}
    \begin{aligned}
        \dot G_{ab}&=-F_{ab}(G,w)-F_{ba}(G,w)+c_\delta\cD_{ab}(G,w),
        &&1\le a,b\le q,\\
        \dot G_{ab}&=-F_{ab}(G,w),
        &&1\le a\le q<b\le q+2,\\
        \dot w_a&=-B_a(G,w),&&1\le a\le q.
    \end{aligned}
\end{equation}
The effective dynamics provided below are all immediate consequences of Proposition~\ref{prop:adam-gpm-dynamics} and \ref{prop:gpm-matrix-dynamics}. The dynamics for online SGD are provided in \cite{BenArousEtAl2025Spectral}.

\begin{corollary}[XOR dynamics for smooth activations]\label{cor:xor-adam-smooth}
    Consider the setting of Proposition~\ref{prop:adam-gpm-dynamics}. Suppose $\varphi\in C^3(\RR)$ and its derivatives through order three have at most polynomial growth. For first-layer ADAM, fix $\beta_{1,\theta},\beta_{2,\theta}\in[0,1)$ and $\epsilon_{A,\theta}>0$, and assume $\theta_0^a,\mu,\nu\in\sD_\zeta$ for some $\zeta\in[0,1/2)$ and all $a\le q$. Suppose $\uv(\Theta_0)\Rightarrow\nu_0$. Then the interpolated summary statistics converge to \eqref{eq:xor-adam-ode} initialized from $\uv_0\sim\nu_0$. The first-layer coefficients $(F^A,\cD^A)$ are those of \eqref{eq:gpm-F-def}--\eqref{eq:gpm-D-def} evaluated with \eqref{eq:xor-alpha-b}.
    For online SGD in the first layer, the same conclusion holds without the coordinate delocalization hypothesis and with $F_{ab}^{\mathrm{SGD}}=\EE[\alpha_{0,a}P_{0,b}]$ and $\cD_{ab}^{\mathrm{SGD}}=\lambda^{-1}\EE[\alpha_{0,a}\alpha_{0,b}]$.
\end{corollary}

For matrix first-layer updates, let $b_j^M=(1-\beta_M)\beta_M^j$ and use the stationary matrices
\begin{equation}\label{eq:xor-muon-CR}
    C_t=\lambda^{-1}\sum(b_j^M)^2\alpha_{t-j}\alpha_{t-j}^\top,
    \qquad R_t=(C_t+\epsilon_M I_q)^{-1/2},
    \qquad R_t^{\mathrm{NS5}}=R_5(C_t),
\end{equation}
where $R_5$ is the polynomial matrix factor in \eqref{eq:ns5-R5-map}.

\begin{proposition}[Mixed MUON--ADAM dynamics on Gaussian XOR]\label{prop:xor-muon-adam-dynamics}
    Suppose $\varphi$ satisfies the growth hypothesis of Corollary~\ref{cor:xor-adam-smooth}. Fix $\beta_M\in[0,1)$ and $\epsilon_M>0$, and choose either the SGD direction or the ADAM direction. Under the setting of Proposition~\ref{prop:gpm-matrix-dynamics}, the summary statistics converge to \eqref{eq:xor-adam-ode}, with the same localization convention as above. The first-layer coefficients are those of \eqref{eq:gpm-matrix-F}--\eqref{eq:gpm-matrix-D} with noise standard deviation $\lambda^{-1/2}$, the vectors $\alpha_t$ in \eqref{eq:xor-alpha-b}, and $R_t$ in \eqref{eq:xor-muon-CR}.
\end{proposition}

\begin{proposition}[Mixed MUON--NS5--ADAM dynamics on Gaussian XOR]\label{prop:xor-muon-ns5-dynamics}
    Under the hypotheses of Proposition~\ref{prop:xor-muon-adam-dynamics}, replace the exact MUON first-layer update by its five-step Newton--Schulz version, with $\epsilon_F>0$. The same conclusion holds, with $R_t^{\mathrm{NS5}}$ replacing $R_t$ in the coefficients of \eqref{eq:gpm-matrix-F}--\eqref{eq:gpm-matrix-D}.
\end{proposition}

As a projective model, we again have that the population risk $\cR(\Theta)=\EE_Y[L(\Theta;Y)]$ depends on $\Theta$ only through $(G,w)$. The pre-activation Hessian is given by,
\begin{equation}\label{eq:xor-preact-hessian}
    H_{ab}(z,w;y)=\sigma'(f(z,w))w_aw_b\varphi'(z_a)\varphi'(z_b)
    +\Id_{\{a=b\}}\bigl(\sigma(f(z,w))-y\bigr)w_a\varphi''(z_a).
\end{equation}
An immediate consequence of Lemma~\ref{lem:gaussian-risk-chain-rule} is the population risk dynamics for Gaussian XOR.

\begin{corollary}\label{cor:xor-pop-risk-ode}
    Under the hypotheses of any of the preceding dynamics results, the population risk dynamics along the limiting trajectory converges to,
    \begin{equation}\label{eq:xor-risk-ode-smooth}
        \dot\cR(G_t,w_t)
        =(1/4)\sum\nolimits_{c=1}^4\EE[
        \sum\nolimits_{a=1}^q\alpha_a\dot m_a^c
        +\sum\nolimits_{a=1}^q b_a\dot w_a
        +(1/(2\lambda))\sum\nolimits_{a,b=1}^q H_{ab}\dot G_{ab}
        \,|\,J=c],
    \end{equation}
    where $\alpha,b,H$ are evaluated at $(P^{\mathrm{st}},w_t;y(c))$, and $(\dot m_{1:q}^1,\dot m_{1:q}^2,\dot m_{1:q}^3,\dot m_{1:q}^4) =(-F_{\cdot,q+1},F_{\cdot,q+1},-F_{\cdot,q+2},F_{\cdot,q+2})$.
\end{corollary}

\section{Examples of Compatible Adaptive, Momentum Algorithms}
\label{sec:compatible-algorithms}

We now verify that the claimed optimizers satisfy the algorithmic regularity conditions in Definition~\ref{def:rho}. For MUON and MUON--NS$K$, we first establish the derivative bounds needed for \eqref{eq:rho-scale-bound}. Throughout, the momentum parameters, positive regularizers, matrix widths, and number of Newton--Schulz iterations are fixed.

\subsection{Derivative bounds for matrix-normalized directions}

\begin{lemma}\label{lem:muon-derivatives}
    Fix $q\ge1$ and $\epsilon>0$. For the MUON direction defined in \eqref{eq:muon-function}, with $X\in\RR^{d\times q}$ and $p=qd$, we have $\rho(0)=0$ and $\Fnorm{\rho(X)}\le\epsilon^{-1/2}\Fnorm X$. For each $\ell\in\{1,2,3\}$, there exists a constant $C_{\epsilon,q,\ell}$, independent of $d$, such that
    \begin{equation}\label{eq:muon-deriv-scaling}
        \sup\nolimits_{X\in\RR^{d\times q}}
        \opnorm{\nabla_X^\ell\rho(X)}
        \le C_{\epsilon,q,\ell}\,p^{(1-\ell)/2}.
    \end{equation}
    The derivative norms are induced by the Frobenius norm on the matrix spaces.
\end{lemma}

\begin{proof}
    Define the normalized map $F(Z):=Z(Z^\top Z+\epsilon I_q)^{-1/2}$ and $\rho(X)=\sqrt d\,F(d^{-1/2}X)$. Clearly $\rho(0)=0$, and the bound on $\Fnorm{\rho(X)}$ follows from $\opnorm{(X^\top X/d+\epsilon I_q)^{-1/2}}\le\epsilon^{-1/2}$. By standard spectral calculus arguments (see, e.g., \cite[Chapter~V, Section~3.11, (3.43)]{kato1995perturbation},
    \begin{equation}\label{eq:F-kato-integral}
        F(Z)=\frac1\pi\int_0^\infty
        \lambda^{-1/2}Z R_\lambda(Z)\,\dd\lambda,
        \qquad
        R_\lambda(Z):=(Z^\top Z+(\epsilon+\lambda)I_q)^{-1}.
    \end{equation}
    By the singular-value decomposition and $\sigma^2+\epsilon+\lambda\ge2\sigma\sqrt{\epsilon+\lambda}$, we have the bounds $\opnorm{R_\lambda(Z)}\le(\epsilon+\lambda)^{-1}$, $\opnorm{Z R_\lambda(Z)}
            \le (\epsilon+\lambda)^{-1/2} / 2$, and $\opnorm{Z R_\lambda(Z)Z^\top}\le1$.

    To control successive derivatives, write $R=R_\lambda(Z)$, $U=ZR$, and $W=I_d-ZRZ^\top$ and observe $\opnorm W\le1$. Differentiating the resolvent gives $\Dd R[H]=-U^\top H R-RH^\top U$
    and hence $\Dd U[H]=WHR-UH^\top U$, $\Dd W[H]=-WHU^\top-UH^\top W$. By differentiating $U$ three times and bounding each of the terms, we have that each derivative $\Dd^\ell U[H_1,\ldots,H_\ell]$ is a finite sum of products bounded by $C_\ell (\epsilon + \lambda )^{-(\ell+1)/2}$ ($\ell = 1$ is immediate from the above). Here we bounded one direction in Frobenius norm and all the others in operator norm. The product rule and the above then give
    \[
        \Fnorm{\Dd^\ell(ZR_\lambda(Z))[H_1,\ldots,H_\ell]}
        \le C_\ell(\epsilon+\lambda)^{-(\ell+1)/2}
        \prod\nolimits_{j=1}^{\ell}\Fnorm{H_j},
        \qquad \ell\ge1,
    \]
    where $C_\ell$ is independent of $d$. Since
    \[
        \int_0^\infty\lambda^{-1/2}
        (\epsilon+\lambda)^{-(\ell+1)/2}\,\dd\lambda
        =\epsilon^{-\ell/2}
        \int_0^\infty u^{-1/2}(1+u)^{-(\ell+1)/2}\,\dd u<\infty
    \]
    for $\ell\ge1$, dominated convergence in \eqref{eq:F-kato-integral} yields $\sup_Z\opnorm{\nabla_Z^\ell F(Z)}\le C_{\epsilon,\ell}$. Finally, $\opnorm{\nabla_X^\ell\rho(X)} \le C_{\epsilon,\ell}q^{(\ell-1)/2}p^{(1-\ell)/2}$ which completes the proof.
\end{proof}

\begin{lemma}\label{lem:muon-ns5-derivatives}
    Fix $q,K\ge1$, $\epsilon_F>0$, and the Newton--Schulz polynomial coefficients. For the MUON--NS$K$ direction in \eqref{eq:muon-ns5-function}, with $X\in\RR^{d\times q}$ and $p=qd$, we have $\rho_{\epsilon_F}^{\mathrm{NS}K}(0)=0$ and $\Fnorm{\rho_{\epsilon_F}^{\mathrm{NS}K}(X)}
        \le C_K\epsilon_F^{-1/2}\Fnorm X$.
    Furthermore, for each $\ell\in\{1,2,3\}$, there is a constant $C_{\epsilon_F,q,K,\ell}$ such that $\opnorm{\nabla_X^\ell\rho_{\epsilon_F}^{\mathrm{NS}K}(X)}
        \le C_{\epsilon_F,q,K,\ell}\,p^{(1-\ell)/2}$ uniformly for $X\in\RR^{d\times q}$.
\end{lemma}

\begin{proof}
    Set $F_{\epsilon_F}:=\Phi_K\circ N_{\epsilon_F}$, so that $\rho_{\epsilon_F}^{\mathrm{NS}K}(X) =\sqrt d\,F_{\epsilon_F}(d^{-1/2}X)$. The function $N_{\epsilon_F}(Z)=Z/(\Fnorm Z^2+\epsilon_F)^{1/2}$ maps into the unit Frobenius ball. Writing $t=\Fnorm Z^2+\epsilon_F$ gives $\Dd N_{\epsilon_F}(Z)[H]
        =t^{-1/2}H-t^{-3/2}\langle Z,H\rangle_F Z$.
    
    Successive applications of the product rule, using $\Fnorm Z\le\sqrt t$ and the fact that $\Fnorm Z^2$ is quadratic give, $\opnorm{\Dd^\ell N_{\epsilon_F}(Z)}
        \le c_\ell\epsilon_F^{-\ell/2}$ for $\ell\in\{1,2,3\}$.
    Each $\Phi_K$ is a fixed matrix polynomial. On the unit Frobenius ball, its first three derivatives are bounded by constants depending only on $K$ and the polynomial coefficients. Expand each monomial, apply the product rule, and use $\Fnorm{AB}\le\opnorm A\Fnorm B$ and $\opnorm A\le\Fnorm A$. The chain rule therefore yields $\sup_Z\opnorm{\Dd^\ell F_{\epsilon_F}(Z)} \le C_{K,\ell}\epsilon_F^{-\ell/2}$. Also, $\Phi_K(0)=0$, so the first-derivative bound on the unit ball gives $\Fnorm{F_{\epsilon_F}(Z)}
        \le C_K\epsilon_F^{-1/2}\Fnorm Z$.
    Rescaling as in Lemma~\ref{lem:muon-derivatives} proves all the assertions.
\end{proof}

\begin{proposition}\label{prop:verify-rho}
    Online SGD, SGD-M, ADAM \eqref{eq:adam-function}, MUON \eqref{eq:muon-function}, fixed-$K$ MUON--NS$K$ \eqref{eq:muon-ns5-function}, and their layer-wise block combinations satisfy Assumption~\ref{def:rho}. Their fixed-parameter momentum maps are contractions with constant $\beta=\opnorm B<1$.
\end{proposition}

\begin{proof}
    For SGD, Polyak momentum, MUON, and MUON--NS$K$, take $\Psi(z)=z$. For ADAM, take $\Psi(z)=(z,z^2)^\top$. These maps have at most quadratic growth, their first derivatives have at most linear growth, and their second derivatives are bounded. The same two-coordinate embedding may be used for all blocks in a mixed algorithm, with the unused second component ignored by the corresponding search direction. Recall that, for a block of scale $S_\alpha=\sqrt{|I_\alpha|}$, For SGD and SGD-M, take a single block with $S=\sqrt p$. Write $\rho(x)=c_\beta x$, where $c_\beta=1$ for SGD and $c_\beta=(1-\beta)^{-1}$ for SGD-M. Since the first derivative is $c_\beta I_p$ and all higher derivatives vanish,
    \[
like         S^{-1}\norm{\rho}_{\cV_{1,S}^3}
        =c_\beta\sup_x \norm x/(S+\norm x)
        +c_\beta S \sup_x(S+\norm x)^{-1}
        =2c_\beta.
    \]
    For ADAM, take the coordinate partition $I_i=\{i\}$ and $S_i=1$. Write $x_i=(m,v)\in\RR\times\RR_{\ge0}$ and $\rho_i(m,v)=m\gamma(v)$, where $\gamma(v)=(v+\epsilon)^{-1/2}$. For $\ell\ge0$, $|\gamma^{(\ell)}(v)|\le(2\ell-1)!!\,2^{-\ell} \epsilon^{-(\ell+1/2)}$. Since $|m|\le V_1(x_i)$, bounding the derivative tensors in Frobenius norm gives $\| \nabla^\ell \rho_i(m,v) \|_{\text{op}} \lesssim_\epsilon V_1 (x_i )$ for $0 \le \ell \le 3$.
    Dividing by $V_1(x_i)$ and summing establishes $\norm{\rho_i}_{\cV_{1,1}^3}\le C_\rho$, uniformly in $i$ and $p$.

    For MUON and MUON--NS$K$, use the single matrix block of scale $S=\sqrt p$ and Lemmas~\ref{lem:muon-derivatives} and~\ref{lem:muon-ns5-derivatives}. All three cases therefore satisfy $S^\ell \sup_x \| \nabla^\ell \rho (x) \|_{\text{op}} / ( \sqrt{p} + \| x \| ) \le C_\ell$.
    Summing over $0\le\ell\le3$ yields $S^{-1}\norm{\rho}_{\cV_{1,S}^3}\le C_\rho$.

    Finally, for layer-wise combinations, take the disjoint union of the layer partitions. Multiplication by a fixed relative learning rate $c_m$ multiplies the corresponding bound by $c_m$, so \eqref{eq:rho-scale-bound} holds with $C_\rho=\max_m c_m C_{\rho^{(m)}}$. Adding unused second-moment coordinates only enlarges the weight in the denominator and thus preserves the derivative bounds.
\end{proof}

\subsection{Unified factorization for Newton--Schulz}

The search directions for Exact MUON and MUON--NS$K$ differ through the right-matrix factor. We record the polynomial factorization and its bounds, using the definitions of $p_K$, $R_K$, and $\tau$ in \eqref{eq:ns5-scalar-poly}--\eqref{eq:ns5-R5-map}.

\begin{lemma}\label{lem:ns5-factorization-bounds}
    Fix $q,K\ge1$ and $\epsilon_F>0$. Then, for every $M\in\RR^{d\times q}$, $\rho_{\epsilon_F}^{\mathrm{NS}K}(M)
        =M R_K(M^\top M/d)$.
    Furthermore, for $\ell\in\{0,1,2\}$, there are constants $C_{q,K,\ell}$ such that $\opnorm{\Dd_S^\ell R_K(S)}
        \le C_{q,K,\ell}\,\epsilon_F^{-(\ell+1/2)}$ for $S \in \Sd_+^q$.
\end{lemma}

\begin{proof}
    Set $S=M^\top M/d$ and $A(S)=S/\tau(S)$. Since $\Fnorm M^2+d\epsilon_F=d\tau(S)$, the initial normalized input $Z_0=d^{-1/2}\tau(S)^{-1/2}M$ satisfies $Z_0^\top Z_0=A(S)$. Define $Z_{j+1}=\phi(Z_j)$ for $0\le j<K$. We claim that $Z_j=Z_0p_j(A(S))$ by induction. This holds for $j=0$ because $p_0=1$, furthermore $A=A(S)$ commutes with $p_j(A)$, and $Z_j^\top Z_j=A p_j(A)^2$. Consequently,
    \[
            Z_{j+1}
            =Z_0p_j(A)\bigl(\mathsf a I_q
            +\mathsf b A p_j(A)^2
            +\mathsf c A^2p_j(A)^4\bigr) =Z_0p_{j+1}(A).
    \]
    Taking $j=K$ and multiplying by $\sqrt d$ proves the factorization. To prove the bounds, observe that $\tau(S)\ge\epsilon_F$ and $0\preceq A(S)\preceq I_q$. For directions $H_1,H_2\in\Sd^q$, $|\Dd\tau(S)[H_1]|=|\tr H_1|\le q\opnorm{H_1}$, $\Dd^2\tau(S)=0$, and
    \[
            \Dd A(S)[H_1]
            =\frac{H_1-\tr(H_1)A(S)}{\tau(S)},\qquad
            \Dd^2 A(S)[H_1,H_2]
            =-\frac{\tr(H_2)\Dd A(S)[H_1]
            +\tr(H_1)\Dd A(S)[H_2]}{\tau(S)}.
    \]
    Hence, for $0\le\ell\le2$, $\opnorm{\Dd^\ell A(S)}\lesssim_q\tau(S)^{-\ell}$ and $\opnorm{\Dd^\ell(\tau(S)^{-1/2})} \lesssim_q\tau(S)^{-(\ell+1/2)}$. The fixed polynomial $p_K$ and its first two derivatives are bounded on $\{A\in\Sd^q:\opnorm A\le1\}$. The product and chain rules applied to $R_K(S)=\tau(S)^{-1/2}p_K(A(S))$ complete the proof.
\end{proof}

\section{Miscellaneous Estimates}
\label{sec:misc-estimates}

\begin{lemma}[Decoupled drift]\label{lem:decoupled-drift}
    Suppose the hypotheses of Theorem~\ref{thm:main-ballistic} hold for the adaptive part of the update. Let $b:\RR^p\to\RR^p$ be $C^2_b$ on $E_R^*\cap A$. Consider the search direction $\rho(X_\ell)+b(\theta_{\ell-1})$, and suppose Assumption~\ref{assume:A-subset-timescale} holds. If $\|\langle b(\theta) , \nabla u^{(i)}(\theta) \rangle - b^{(i)}_{\uv}(\uv_n(\theta)) \| \to 0$ uniformly on $E_R^*\cap A$ with locally Lipschitz limits, then the limiting drift becomes $\Psiv-b_{\uv}$, where $b_{\uv}=(b^{(i)}_{\uv})_{i\le k}$.
\end{lemma}

\begin{proof}
    Denote the operators for $\rho+b$ by $\cA_b,\cL_b$. For each statistic $u$, $\tilde\cA_bu=\tilde\cA u+\nabla u[b]$ and $\tilde\cL_bu-\tilde\cL u
        =\pi_\theta[\nabla^2u[\rho,b]]+ \nabla^2u[b,b] / 2$.
    Since $\norm\rho_{L^2(\pi_\theta)}\lesssim_{R, A} \sqrt p$, $\norm{\opnorm{\nabla_\theta\chi_1}}_{L^2(\pi_\theta)}\lesssim_{R, A} \sqrt p$, the difference vanishes, i.e. 
    \[
        | ( \Psi^u_{n, b} (\theta ) + \nabla u (\theta ) [b ( \theta ) ] - \Psi^u_n (\theta ) | \le \delta|\tilde\cL_bu-\tilde\cL u|
        +\delta|\pi_\theta[\nabla_\theta\chi_1[b]]|
        \le C_{A,R}\delta(1+\sqrt p) \qedhere
    \]
\end{proof}

\bibliography{main}

@inproceedings{jagannath2026high,
  title={High-dimensional limit theorems for SGD: Momentum and adaptive step-sizes},
  author={Jagannath, Aukosh and Jones-McCormick, Taj and Sarangian, Varnan},
  booktitle={International Conference on Learning Representations},
  volume={2026},
  pages={153259--153306},
  year={2026}
}

@article{ben2022high,
  title={High-dimensional limit theorems for sgd: Effective dynamics and critical scaling},
  author={Ben Arous, G\'erard and Gheissari, Reza and Jagannath, Aukosh},
  journal={Advances in neural information processing systems},
  volume={35},
  pages={25349--25362},
  year={2022}
}

@book{goodfellow2016deep,
  title={Deep learning},
  author={Goodfellow, Ian and Bengio, Yoshua and Courville, Aaron and Bengio, Yoshua},
  volume={1},
  number={2},
  year={2016},
  publisher={MIT press, Cambridge}
}

@article{ferbach2025dimension,
  title={Dimension-adapted momentum outscales SGD},
  author={Ferbach, Damien and Everett, Katie and Gidel, Gauthier and Paquette, Elliot and Paquette, Courtney},
  journal={arXiv preprint arXiv:2505.16098},
  year={2025}
}

@book{stroock2006multidimensional,
  title        = {Multidimensional Diffusion Processes},
  author       = {Stroock, Daniel W. and Varadhan, S. R. Srinivasa},
  series       = {Classics in Mathematics},
  publisher    = {Springer-Verlag},
  address      = {Berlin},
  year         = {2006},
  note         = {Reprint of the 1997 edition}
}

@article{Mamis03042026,
author = {Konstantinos Mamis},
title = {New formulas for moments of the multivariate normal distribution extending Stein's lemma and Isserlis theorem},
journal = {Communications in Statistics - Theory and Methods},
volume = {55},
number = {7},
pages = {2008--2030},
year = {2026},
publisher = {Taylor \& Francis},
doi = {10.1080/03610926.2025.2541838},
URL = {     
        https://doi.org/10.1080/03610926.2025.2541838
    
    

},
eprint = { 
    
        https://doi.org/10.1080/03610926.2025.2541838
    
    

}

}

@book{kato1995perturbation,
  title     = {Perturbation Theory for Linear Operators},
  author    = {Kato, Tosio},
  series    = {Classics in Mathematics},
  edition   = {2},
  year      = {1995},
  publisher = {Springer-Verlag},
  address   = {Berlin}
}

@inproceedings{kingma2015adam,
  title = {{Adam}: A Method for Stochastic Optimization},
  author = {Diederik P. Kingma and Jimmy Ba},
  booktitle = {International Conference on Learning Representations},
  year = {2015},
  url = {https://arxiv.org/abs/1412.6980},
}

@misc{jordan2024muon,
  author = {Keller Jordan and Yuchen Jin and Vlado Boza and Jiacheng You and
                  Franz Cesista and Laker Newhouse and Jeremy Bernstein},
  title = {{Muon}: An optimizer for hidden layers in neural networks},
  year = {2024},
  url = {https://kellerjordan.github.io/posts/muon/},
}

@misc{liu2025muon,
  title = {{Muon} is Scalable for {LLM} Training},
  author = {Jingyuan Liu et al},
  year = {2025},
  eprint = {2502.16982},
  archiveprefix = {arXiv},
  primaryclass = {cs.LG},
  url = {https://arxiv.org/abs/2502.16982},
}

@misc{loshchilov2019decoupled,
  title = {Decoupled Weight Decay Regularization},
  author = {Ilya Loshchilov and Frank Hutter},
  year = {2019},
  eprint = {1711.05101},
  archiveprefix = {arXiv},
  primaryclass = {cs.LG},
  url = {https://arxiv.org/abs/1711.05101},
}

@article{robbins1951stochastic,
  title = {A Stochastic Approximation Method},
  volume = {22},
  issn = {0003-4851},
  url = {http://dx.doi.org/10.1214/aoms/1177729586},
  doi = {10.1214/aoms/1177729586},
  number = {3},
  journal = {The Annals of Mathematical Statistics},
  publisher = {Institute of Mathematical Statistics},
  author = {Robbins, Herbert and Monro, Sutton},
  year = {1951},
  month = sep,
  pages = {400--407},
}

@article{ljung1977analysis,
  title = {Analysis of recursive stochastic algorithms},
  volume = {22},
  issn = {0018-9286},
  url = {http://dx.doi.org/10.1109/TAC.1977.1101561},
  doi = {10.1109/tac.1977.1101561},
  number = {4},
  journal = {IEEE Transactions on Automatic Control},
  publisher = {Institute of Electrical and Electronics Engineers (IEEE)},
  author = {Ljung, L.},
  year = {1977},
  month = aug,
  pages = {551--575},
}

@book{kushner2003stochastic,
  author = {Kushner, Harold J. and Yin, G. George},
  title = {Stochastic Approximation and Recursive Algorithms and Applications},
  edition = {2},
  series = {Stochastic Modelling and Applied Probability},
  volume = {35},
  publisher = {Springer},
  year = {2003},
  isbn = {9780387008943},
  doi = {10.1007/b97441},
  url = {https://link.springer.com/book/10.1007/b97441},
}

@book{benveniste1990adaptive,
  title = {Adaptive Algorithms and Stochastic Approximations},
  isbn = {9783642758942},
  url = {http://dx.doi.org/10.1007/978-3-642-75894-2},
  doi = {10.1007/978-3-642-75894-2},
  publisher = {Springer Berlin Heidelberg},
  author = {Benveniste, Albert and M\'etivier, Michel and Priouret, Pierre},
  year = {1990},
}

@article{benaim1999dynamics,
  author = {Bena\"im, Michel},
  title = {Dynamics of stochastic approximation algorithms},
  journal = {S\'eminaire de probabilit\'es},
  pages = {1--68},
  year = {1999},
  publisher = {Springer - Lecture Notes in Mathematics},
  volume = {33},
  mrnumber = {1767993},
  zbl = {0955.62085},
  language = {en},
  url = {https://numdam.org/item/SPS_1999__33__1_0/},
}

@inproceedings{wilson2017marginal,
  title = {The Marginal Value of Adaptive Gradient Methods in Machine Learning},
  author = {Wilson, Ashia C and Roelofs, Rebecca and Stern, Mitchell and Srebro, Nati and Recht, Benjamin},
  booktitle = {Proceedings of the 31st International Conference on Neural Information Processing Systems},
  pages = {4151--4161},
  year = {2017},
  url = {https://arxiv.org/pdf/1705.08292.pdf},
}

@misc{keskar2017switching,
  title = {Improving Generalization Performance by Switching from {Adam} to {SGD}},
  author = {Nitish Shirish Keskar and Richard Socher},
  year = {2017},
  eprint = {1712.07628},
  archiveprefix = {arXiv},
  primaryclass = {cs.LG},
  url = {https://arxiv.org/abs/1712.07628},
}

@inproceedings{wu2025early,
  title = {Benefits of Early Stopping in Gradient Descent for Overparameterized Logistic Regression},
  author = {Wu, Jingfeng and Bartlett, Peter and Telgarsky, Matus and Yu, Bin},
  booktitle = {Proceedings of the 42nd International Conference on Machine Learning},
  pages = {67081--67110},
  year = {2025},
}

@inproceedings{cohen2021gradient,
  title = {Gradient Descent on Neural Networks Typically Occurs at the Edge of Stability},
  author = {Cohen, Jeremy M. and Kaur, Simran and Li, Yuanzhi and Kolter, J. Zico and Talwalkar, Ameet},
  booktitle = {International Conference on Learning Representations},
  year = {2021},
  eprint = {2103.00065},
  archiveprefix = {arXiv},
  url = {https://openreview.net/forum?id=jh-rTtvkGeM},
}

@misc{cohen2024adaptive,
  title = {Adaptive Gradient Methods at the Edge of Stability},
  author = {Jeremy M. Cohen and Behrooz Ghorbani and Shankar Krishnan and Naman Agarwal and Sourabh Medapati and Michal Badura and Daniel Suo and David Cardoze and Zachary Nado and George E. Dahl and Justin Gilmer},
  year = {2024},
  eprint = {2207.14484},
  archiveprefix = {arXiv},
  primaryclass = {cs.LG},
  url = {https://arxiv.org/abs/2207.14484},
}

@inproceedings{lee2023new,
  title = {A new characterization of the edge of stability based on a sharpness measure aware of batch gradient distribution},
  author = {Lee, Sungyoon and Jang, Cheongjae},
  booktitle = {International Conference on Learning Representations},
  year = {2023},
  url = {https://openreview.net/forum?id=bH-kCY6LdKg},
}

@misc{andreyev2025edge,
  title = {Edge of Stochastic Stability: Revisiting the Edge of Stability for {SGD}},
  author = {Arseniy Andreyev and Pierfrancesco Beneventano},
  year = {2025},
  eprint = {2412.20553},
  archiveprefix = {arXiv},
  primaryclass = {cs.LG},
  url = {https://arxiv.org/abs/2412.20553},
}

@inproceedings{wu2018how,
  author = {Wu, Lei and Ma, Chao and E, Weinan},
  booktitle = {Advances in Neural Information Processing Systems},
  editor = {S. Bengio and H. Wallach and H. Larochelle and K. Grauman and N. Cesa-Bianchi and R. Garnett},
  pages = {},
  publisher = {Curran Associates, Inc.},
  title = {How {SGD} Selects the Global Minima in Over-parameterized Learning: A Dynamical Stability Perspective},
  url = {https://proceedings.neurips.cc/paper_files/paper/2018/file/6651526b6fb8f29a00507de6a49ce30f-Paper.pdf},
  volume = {31},
  year = {2018},
}

@inproceedings{mulayoff2024exact,
  title = {Exact Mean Square Linear Stability Analysis for {SGD}},
  author = {Mulayoff, Rotem and Michaeli, Tomer},
  booktitle = {Proceedings of Thirty Seventh Conference on Learning Theory},
  pages = {3915--3969},
  year = {2024},
  editor = {Agrawal, Shipra and Roth, Aaron},
  volume = {247},
  series = {Proceedings of Machine Learning Research},
  month = {30 Jun--03 Jul},
  publisher = {PMLR},
  url = {https://proceedings.mlr.press/v247/mulayoff24a.html},
}

@misc{damian2023self,
  title = {Self-Stabilization: The Implicit Bias of Gradient Descent at the Edge of Stability},
  author = {Alex Damian and Eshaan Nichani and Jason D. Lee},
  year = {2023},
  eprint = {2209.15594},
  archiveprefix = {arXiv},
  primaryclass = {cs.LG},
  url = {https://arxiv.org/abs/2209.15594},
}

@misc{jastrzebski2020break,
  title = {The Break-Even Point on Optimization Trajectories of Deep Neural Networks},
  author = {Stanislaw Jastrzebski and Maciej Szymczak and Stanislav Fort and Devansh Arpit and Jacek Tabor and Kyunghyun Cho and Krzysztof Geras},
  year = {2020},
  eprint = {2002.09572},
  archiveprefix = {arXiv},
  primaryclass = {cs.LG},
  url = {https://arxiv.org/abs/2002.09572},
}

@inproceedings{andreyev2026momentum,
  title = {Momentum Further Constrains Sharpness at the Edge of Stochastic Stability},
  author = {Andreyev, Arseniy and Ananthkumar, Advikar and Walden, Marc and Poggio, Tomaso and Beneventano, Pierfrancesco},
  booktitle = {Proceedings of the 43rd International Conference on Machine Learning},
  pages = {2831--2871},
  year = {2026},
}

@misc{jastrzebski2019sharpest,
  title = {On the Relation Between the Sharpest Directions of {DNN} Loss and the {SGD} Step Length},
  author = {Stanis{\l}aw Jastrz{\k{e}}bski and Zachary Kenton and Nicolas Ballas and Asja Fischer and Yoshua Bengio and Amos Storkey},
  year = {2019},
  eprint = {1807.05031},
  archiveprefix = {arXiv},
  primaryclass = {stat.ML},
  url = {https://arxiv.org/abs/1807.05031},
}

@inproceedings{arora2022understanding,
  title = {Understanding Gradient Descent on the Edge of Stability in Deep Learning},
  author = {Arora, Sanjeev and Li, Zhiyuan and Panigrahi, Abhishek},
  booktitle = {Proceedings of the 39th International Conference on Machine Learning},
  pages = {948--1024},
  year = {2022},
  editor = {Chaudhuri, Kamalika and Jegelka, Stefanie and Song, Le and Szepesvari, Csaba and Niu, Gang and Sabato, Sivan},
  volume = {162},
  series = {Proceedings of Machine Learning Research},
  month = {17--23 Jul},
  publisher = {PMLR},
  url = {https://proceedings.mlr.press/v162/arora22a.html},
}

@inproceedings{lyu2022normalization,
  author = {Lyu, Kaifeng and Li, Zhiyuan and Arora, Sanjeev},
  title = {Understanding the Generalization Benefit of Normalization Layers: Sharpness Reduction},
  booktitle = {Advances in Neural Information Processing Systems 35},
  volume = {35},
  pages = {34689--34708},
  year = {2022},
  url = {https://papers.nips.cc/paper_files/paper/2022/hash/dffd1c523512e557f4e75e8309049213-Abstract-Conference.html},
}

@inproceedings{chen2023beyond,
  title = {Beyond the Edge of Stability via Two-step Gradient Updates},
  author = {Chen, Lei and Bruna, Joan},
  booktitle = {Proceedings of the 40th International Conference on Machine Learning},
  pages = {4330--4391},
  year = {2023},
  editor = {Krause, Andreas and Brunskill, Emma and Cho, Kyunghyun and Engelhardt, Barbara and Sabato, Sivan and Scarlett, Jonathan},
  volume = {202},
  series = {Proceedings of Machine Learning Research},
  month = {23--29 Jul},
  publisher = {PMLR},
  url = {https://proceedings.mlr.press/v202/chen23b.html},
}

@inproceedings{agarwala2023second,
  title = {Second-order regression models exhibit progressive sharpening to the edge of stability},
  author = {Agarwala, Atish and Pedregosa, Fabian and Pennington, Jeffrey},
  booktitle = {Proceedings of the 40th International Conference on Machine Learning},
  pages = {169--195},
  year = {2023},
}

@misc{zhu2023minimalist,
  title = {Understanding Edge-of-Stability Training Dynamics with a Minimalist Example},
  author = {Xingyu Zhu and Zixuan Wang and Xiang Wang and Mo Zhou and Rong Ge},
  year = {2023},
  eprint = {2210.03294},
  archiveprefix = {arXiv},
  primaryclass = {cs.LG},
  url = {https://arxiv.org/abs/2210.03294},
}

@misc{agarwala2025conservative,
  title = {High dimensional analysis reveals conservative sharpening and a stochastic edge of stability},
  author = {Atish Agarwala and Jeffrey Pennington},
  year = {2025},
  eprint = {2404.19261},
  archiveprefix = {arXiv},
  primaryclass = {cs.LG},
  url = {https://arxiv.org/abs/2404.19261},
}

@article{devlin2019bert,
  title = {{BERT}: Pre-training of Deep Bidirectional Transformers for Language Understanding},
  author = {Devlin, Jacob  and
      Chang, Ming-Wei  and
      Lee, Kenton  and
      Toutanova, Kristina},
  booktitle = {Proceedings of the 2019 Conference of the {N}orth {A}merican Chapter of the {A}ssociation for {C}omputational {L}inguistics: Human Language Technologies, Volume 1 (Long and Short Papers)},
  pages = {4171--4186},
}

@article{schmidt2021crowded,
  title = {Descending through a Crowded Valley - Benchmarking Deep Learning Optimizers},
  author = {Schmidt, Robin M and Schneider, Frank and Hennig, Philipp},
  booktitle = {Proceedings of the 38th International Conference on Machine Learning},
  pages = {9367--9376},
  year = {2021}
}

@misc{kimi2025k2,
  title = {Kimi {K2}: Open Agentic Intelligence},
  author = {{Kimi Team} et al},
  year = {2025},
  eprint = {2507.20534},
  archiveprefix = {arXiv},
  primaryclass = {cs.LG},
  url = {https://arxiv.org/abs/2507.20534},
}

@article{dupuis1988large,
  title = {Large Deviations Analysis of Some Recursive Algorithms with State Dependent Noise},
  volume = {16},
  issn = {2168-894X},
  url = {http://dx.doi.org/10.1214/aop/1176991581},
  doi = {10.1214/aop/1176991581},
  number = {4},
  journal = {The Annals of Probability},
  publisher = {Institute of Mathematical Statistics},
  author = {Dupuis, Paul},
  year = {1988},
  month = oct,
  pages = {1509--1536},
}

@article{SaadSolla1995Exact,
  title = {Exact Solution for On-Line Learning in Multilayer Neural Networks},
  volume = {74},
  issn = {1079-7114},
  url = {http://dx.doi.org/10.1103/PhysRevLett.74.4337},
  doi = {10.1103/physrevlett.74.4337},
  number = {21},
  journal = {Physical Review Letters},
  publisher = {American Physical Society (APS)},
  author = {Saad, David and Solla, Sara A.},
  year = {1995},
  month = may,
  pages = {4337--4340},
}

@article{SaadSolla1995Committee,
  title = {On-line learning in soft committee machines},
  volume = {52},
  issn = {1095-3787},
  url = {http://dx.doi.org/10.1103/PhysRevE.52.4225},
  doi = {10.1103/physreve.52.4225},
  number = {4},
  journal = {Physical Review E},
  publisher = {American Physical Society (APS)},
  author = {Saad, David and Solla, Sara A.},
  year = {1995},
  month = oct,
  pages = {4225--4243},
}

@article{mcleish1976functional,
  title={Functional and random central limit theorems for the Robbins-Munro process},
  author={McLeish, DL},
  journal={Journal of Applied Probability},
  volume={13},
  number={1},
  pages={148--154},
  year={1976},
  publisher={Cambridge University Press}
}

@misc{WangMattinglyLu2017Scaling,
  title = {Scaling Limit: Exact and Tractable Analysis of Online Learning Algorithms with Applications to Regularized Regression and {PCA}},
  author = {Chuang Wang and Jonathan Mattingly and Yue M. Lu},
  year = {2017},
  eprint = {1712.04332},
  archiveprefix = {arXiv},
  primaryclass = {cs.LG},
  url = {https://arxiv.org/abs/1712.04332},
}

@article{GoldtEtAl2020TeacherStudent,
  title = {Dynamics of stochastic gradient descent for two-layer neural networks in the teacher--student setup},
  volume = {2020},
  issn = {1742-5468},
  url = {http://dx.doi.org/10.1088/1742-5468/abc61e},
  doi = {10.1088/1742-5468/abc61e},
  number = {12},
  journal = {Journal of Statistical Mechanics: Theory and Experiment},
  publisher = {IOP Publishing},
  author = {Goldt, Sebastian and Advani, Madhu S and Saxe, Andrew M and Krzakala, Florent and Zdeborov{\'a}, Lenka},
  year = {2020},
  month = dec,
  pages = {124010},
}

@article{RefinettiEtAl2021Mixtures,
  title = {Classifying high-dimensional {Gaussian} mixtures: Where kernel methods fail and neural networks succeed},
  author = {Refinetti, Maria and Goldt, Sebastian and Krzakala, Florent and Zdeborova, Lenka},
  booktitle = {Proceedings of the 38th International Conference on Machine Learning},
  pages = {8936--8947},
  year = {2021},
}

@article{veiga2023phase,
  title = {Phase diagram of stochastic gradient descent in high-dimensional two-layer neural networks},
  volume = {2023},
  issn = {1742-5468},
  url = {http://dx.doi.org/10.1088/1742-5468/ad01b1},
  doi = {10.1088/1742-5468/ad01b1},
  number = {11},
  journal = {Journal of Statistical Mechanics: Theory and Experiment},
  publisher = {IOP Publishing},
  author = {Veiga, Rodrigo and Stephan, Ludovic and Loureiro, Bruno and Krzakala, Florent and Zdeborov{\'a}, Lenka},
  year = {2023},
  month = nov,
  pages = {114008},
}

@inproceedings{paquette2021sgd,
  title = {{SGD} in the Large: Average-case Analysis, Asymptotics, and Stepsize Criticality},
  author = {Paquette, Courtney and Lee, Kiwon and Pedregosa, Fabian and Paquette, Elliot},
  booktitle = {Proceedings of Thirty Fourth Conference on Learning Theory},
  year = {2021},
}

@article{collinswoodfin2024onepass,
  title = {High-dimensional limit of one-pass {SGD} on least squares},
  volume = {29},
  issn = {1083-589X},
  url = {http://dx.doi.org/10.1214/23-ECP571},
  doi = {10.1214/23-ecp571},
  number = {none},
  journal = {Electronic Communications in Probability},
  publisher = {Institute of Mathematical Statistics},
  author = {Collins--Woodfin, Elizabeth and Paquette, Elliot},
  year = {2024},
  month = jan,
  pages = {1--15},
}

@article{collinswoodfin2024highdim,
  title = {Hitting the High-dimensional notes: an {ODE} for {SGD} learning dynamics on {GLMs} and multi-index models},
  volume = {13},
  issn = {2049-8772},
  url = {http://dx.doi.org/10.1093/imaiai/iaae028},
  doi = {10.1093/imaiai/iaae028},
  number = {4},
  journal = {Information and Inference: A Journal of the IMA},
  publisher = {Oxford University Press (OUP)},
  author = {Collins-Woodfin, Elizabeth and Paquette, Courtney and Paquette, Elliot and Seroussi, Inbar},
  year = {2024},
  month = sep,
  pages = {iaae028},
}

@article{paquette2025homogenization,
  title = {Homogenization of {SGD} in high-dimensions: exact dynamics and generalization properties},
  volume = {214},
  issn = {1436-4646},
  url = {http://dx.doi.org/10.1007/s10107-024-02171-3},
  doi = {10.1007/s10107-024-02171-3},
  number = {1-2},
  journal = {Mathematical Programming},
  publisher = {Springer Science and Business Media LLC},
  author = {Paquette, Courtney and Paquette, Elliot and Adlam, Ben and Pennington, Jeffrey},
  year = {2025},
  pages = {1--90},
}

@misc{GheissariJagannath2025Universality,
  title = {Universality of high-dimensional scaling limits of stochastic gradient descent},
  author = {Reza Gheissari and Aukosh Jagannath},
  year = {2025},
  eprint = {2512.13634},
  archiveprefix = {arXiv},
  primaryclass = {stat.ML},
  url = {https://arxiv.org/abs/2512.13634},
}

@article{BenArousGheissariJagannath2021Online,
  author = {Ben Arous, G{\'e}rard and Gheissari, Reza and Jagannath, Aukosh},
  title = {Online stochastic gradient descent on non-convex losses from high-dimensional inference},
  journal = {Journal of Machine Learning Research},
  year = {2021},
  volume = {22},
  number = {106},
  pages = {1--51},
  url = {http://jmlr.org/papers/v22/20-1288.html},
}

@article{AbbeBoixAdseraMisiakiewicz2022Staircase,
  title = {The merged-staircase property: a necessary and nearly sufficient condition for {SGD} learning of sparse functions on two-layer neural networks},
  author = {Abbe, Emmanuel and Boix Adser{\`a}, Enric and Misiakiewicz, Theodor},
  booktitle = {Proceedings of Thirty Fifth Conference on Learning Theory},
  pages = {4782--4887},
  year = {2022},
}

@article{AbbeBoixAdseraMisiakiewicz2023Leap,
  title = {SGD learning on neural networks: leap complexity and saddle-to-saddle dynamics},
  author = {Abbe, Emmanuel and Boix Adser{\`a}, Enric and Misiakiewicz, Theodor},
  booktitle = {Proceedings of Thirty Sixth Conference on Learning Theory},
  pages = {2552--2623},
  year = {2023},
}

@article{DamianLeeSoltanolkotabi2022Representations,
  title = {Neural Networks can Learn Representations with Gradient Descent},
  author = {Damian, Alexandru and Lee, Jason and Soltanolkotabi, Mahdi},
  booktitle = {Proceedings of Thirty Fifth Conference on Learning Theory},
  pages = {5413--5452},
  year = {2022},
}

@article{anastasiou2019normal,
  title = {Normal Approximation for Stochastic Gradient Descent via Non-Asymptotic Rates of Martingale {CLT}},
  author = {Anastasiou, Andreas and Balasubramanian, Krishnakumar and Erdogdu, Murat A.},
  booktitle = {Proceedings of the Thirty-Second Conference on Learning Theory},
  pages = {115--137},
  year = {2019},

}

@article{paquette2024scaling,
  author = {Paquette, Elliot and Paquette, Courtney and Xiao, Lechao and Pennington, Jeffrey},
  booktitle = {Advances in Neural Information Processing Systems},
  doi = {10.52202/079017-0526},
  pages = {16459--16537},
  title = {4+3 Phases of Compute-Optimal Neural Scaling Laws},
  year = {2024},
}

@article{BenArousEtAl2025Spectral,
  title = {Spectral alignment of stochastic gradient descent for high-dimensional classification tasks},
  volume = {35},
  issn = {2168-8737},
  url = {http://dx.doi.org/10.1214/25-AAP2187},
  doi = {10.1214/25-aap2187},
  number = {4},
  journal = {The Annals of Applied Probability},
  publisher = {Institute of Mathematical Statistics},
  author = {Ben Arous, G{\'e}rard and Gheissari, Reza and Huang, Jiaoyang and Jagannath, Aukosh},
  year = {2025},
  month = aug,
  pages = {2767--2822},
}

@article{BenArousEtAl2026Geometry,
  title = {Local geometry of high-dimensional mixture models: Effective spectral theory and dynamical transitions},
  volume = {54},
  issn = {2168-8966},
  url = {http://dx.doi.org/10.1214/26-AOS2621},
  doi = {10.1214/26-aos2621},
  number = {4},
  journal = {The Annals of Statistics},
  publisher = {Institute of Mathematical Statistics},
  author = {Ben Arous, G{\'e}rard and Gheissari, Reza and Huang, Jiaoyang and Jagannath, Aukosh},
  year = {2026},
  month = aug,
  pages = {1682--1711},
}

@misc{FrankleCarbin2019Lottery,
  title = {The Lottery Ticket Hypothesis: Finding Sparse, Trainable Neural Networks},
  author = {Jonathan Frankle and Michael Carbin},
  year = {2019},
  eprint = {1803.03635},
  archiveprefix = {arXiv},
  primaryclass = {cs.LG},
  url = {https://arxiv.org/abs/1803.03635},
}

@article{kovachki2021continuous,
  author = {Nikola B. Kovachki and Andrew M. Stuart},
  title = {Continuous Time Analysis of Momentum Methods},
  journal = {Journal of Machine Learning Research},
  year = {2021},
  volume = {22},
  number = {17},
  pages = {1--40},
  url = {http://jmlr.org/papers/v22/19-466.html},
}

@article{belotto2020general,
  author = {Andre Belotto da Silva and Maxime Gazeau},
  title = {A General System of Differential Equations to Model First-Order Adaptive Algorithms},
  journal = {Journal of Machine Learning Research},
  year = {2020},
  volume = {21},
  number = {129},
  pages = {1--42},
  url = {http://jmlr.org/papers/v21/18-808.html},
}

@article{barakat2021adam,
  title = {Convergence and Dynamical Behavior of the {ADAM} Algorithm for Nonconvex Stochastic Optimization},
  volume = {31},
  issn = {1095-7189},
  url = {http://dx.doi.org/10.1137/19M1263443},
  doi = {10.1137/19m1263443},
  number = {1},
  journal = {SIAM Journal on Optimization},
  publisher = {Society for Industrial \& Applied Mathematics (SIAM)},
  author = {Barakat, Anas and Bianchi, Pascal},
  year = {2021},
  month = jan,
  pages = {244--274},
}

@article{ma2022qualitative,
  title = {A Qualitative Study of the Dynamic Behavior for Adaptive Gradient Algorithms},
  author = {Ma, Chao and Wu, Lei and E, Weinan},
  booktitle = {Proceedings of the 2nd Mathematical and Scientific Machine Learning Conference},
  pages = {671--692},
  year = {2022},
}

@article{malladi2022sdes,
  author = {Malladi, Sadhika and Lyu, Kaifeng and Panigrahi, Abhishek and Arora, Sanjeev},
  booktitle = {Advances in Neural Information Processing Systems},
  doi = {10.52202/068431-0559},
  pages = {7697--7711},
  title = {On the {SDEs} and Scaling Rules for Adaptive Gradient Algorithms},
  year = {2022},
}

@misc{reddi2019convergence,
  title = {On the Convergence of {Adam} and Beyond},
  author = {Sashank J. Reddi and Satyen Kale and Sanjiv Kumar},
  year = {2019},
  eprint = {1904.09237},
  archiveprefix = {arXiv},
  primaryclass = {cs.LG},
  url = {https://arxiv.org/abs/1904.09237},
}

@article{gess2026exponential,
  title = {Exponential convergence rates for momentum stochastic gradient descent in the overparametrized setting},
  issn = {1436-4646},
  url = {http://dx.doi.org/10.1007/s10107-025-02308-y},
  doi = {10.1007/s10107-025-02308-y},
  journal = {Mathematical Programming},
  publisher = {Springer Science and Business Media LLC},
  author = {Gess, Benjamin and Kassing, Sebastian},
  year = {2026},
  month = jan,
}

@article{liu2021diffusion,
  title = {A Diffusion Approximation Theory of Momentum Stochastic Gradient Descent in Nonconvex Optimization},
  volume = {11},
  issn = {1946-5238},
  url = {http://dx.doi.org/10.1287/stsy.2021.0083},
  doi = {10.1287/stsy.2021.0083},
  number = {4},
  journal = {Stochastic Systems},
  publisher = {Institute for Operations Research and the Management Sciences (INFORMS)},
  author = {Liu, Tianyi and Chen, Zhehui and Zhou, Enlu and Zhao, Tuo},
  year = {2021},
  month = dec,
  pages = {307--323},
}

@misc{wang2021escaping,
  title = {Escaping Saddle Points Faster with Stochastic Momentum},
  author = {Jun-Kun Wang and Chi-Heng Lin and Jacob Abernethy},
  year = {2021},
  eprint = {2106.02985},
  archiveprefix = {arXiv},
  primaryclass = {cs.LG},
  url = {https://arxiv.org/abs/2106.02985},
}

@misc{ghosh2023implicit,
  title = {Implicit regularization in Heavy-ball momentum accelerated stochastic gradient descent},
  author = {Avrajit Ghosh and He Lyu and Xitong Zhang and Rongrong Wang},
  year = {2023},
  eprint = {2302.00849},
  archiveprefix = {arXiv},
  primaryclass = {cs.LG},
  url = {https://arxiv.org/abs/2302.00849},
}

@article{cattaneo2024implicit,
  title = {On the Implicit Bias of {A}dam},
  author = {Cattaneo, Matias D. and Klusowski, Jason Matthew and Shigida, Boris},
  booktitle = {Proceedings of the 41st International Conference on Machine Learning},
  pages = {5862--5906},
  year = {2024},
}

@misc{cattaneo2026minibatch,
  title = {The Effect of Mini-Batch Noise on the Implicit Bias of {Adam}},
  author = {Matias D. Cattaneo and Boris Shigida},
  year = {2026},
  eprint = {2602.01642},
  archiveprefix = {arXiv},
  primaryclass = {cs.LG},
  url = {https://arxiv.org/abs/2602.01642},
}

@article{gadat2018heavyball,
  title = {Stochastic heavy ball},
  volume = {12},
  issn = {1935-7524},
  url = {http://dx.doi.org/10.1214/18-EJS1395},
  doi = {10.1214/18-ejs1395},
  number = {1},
  journal = {Electronic Journal of Statistics},
  publisher = {Institute of Mathematical Statistics},
  author = {Gadat, S{\'e}bastien and Panloup, Fabien and Saadane, Sofiane},
  year = {2018},
  month = jan,
  pages = {461--529},
}

@misc{dereich2026clt,
  title = {Central limit theorem for the averaged {Adam} optimizer},
  author = {Steffen Dereich and Arnulf Jentzen},
  year = {2026},
  eprint = {2606.21433},
  archiveprefix = {arXiv},
  primaryclass = {math.PR},
  url = {https://arxiv.org/abs/2606.21433},
}

@misc{dereich2025ode,
  title = {{ODE} approximation for the {Adam} algorithm: General and overparametrized setting},
  author = {Steffen Dereich and Arnulf Jentzen and Sebastian Kassing},
  year = {2025},
  eprint = {2511.04622},
  archiveprefix = {arXiv},
  primaryclass = {math.OC},
  url = {https://arxiv.org/abs/2511.04622},
}

@inproceedings{paquette2021momentum,
  author = {Paquette, Courtney and Paquette, Elliot},
  booktitle = {Advances in Neural Information Processing Systems},
  title = {Dynamics of Stochastic Momentum Methods on Large-scale, Quadratic Models},
  year = {2021},
}

@inproceedings{lee2022momentum,
  author = {Lee, Kiwon and Cheng, Andrew and Paquette, Elliot and Paquette, Courtney},
  booktitle = {Advances in Neural Information Processing Systems},
  doi = {10.52202/068431-2677},
  title = {Trajectory of Mini-Batch Momentum: Batch Size Saturation and Convergence in High Dimensions},
  volume = {35},
  year = {2022},
}

@inproceedings{collinswoodfin2024highline,
  author = {Collins-Woodfin, Elizabeth and Seroussi, Inbar and Garc\'{\i}a Malaxechebarr\'{\i}a, Bego\~{n}a and Mackenzie, Andrew W. and Paquette, Elliot and Paquette, Courtney},
  booktitle = {Advances in Neural Information Processing Systems},
  doi = {10.52202/079017-0209},
  editor = {A. Globerson and L. Mackey and D. Belgrave and A. Fan and U. Paquet and J. Tomczak and C. Zhang},
  pages = {6500--6548},
  publisher = {Curran Associates, Inc.},
  title = {The High Line: Exact Risk and Learning Rate Curves of Stochastic Adaptive Learning Rate Algorithms},
  url = {https://proceedings.neurips.cc/paper_files/paper/2024/file/0c7ca207a051228f978971447a56464a-Paper-Conference.pdf},
  volume = {37},
  year = {2024},
}

@article{xiao2025signsgd,
  title = {Exact risk curves of sign{SGD} in High-Dimensions: quantifying preconditioning and noise-compression effects},
  author = {Xiao, Ke Liang and Marshall, Noah and Agarwala, Atish and Paquette, Elliot},
  booktitle = {Proceedings of the 42nd International Conference on Machine Learning},
  pages = {68391--68439},
  year = {2025},
}

@article{marshall2025clipping,
  author = {Marshall, Noah and Xiao, Ke Liang and Agarwala, Atish and Paquette, Elliot},
  booktitle = {International Conference on Learning Representations},
  pages = {27381--27417},
  title = {To Clip or not to Clip: the Dynamics of {SGD} with Gradient Clipping in High-Dimensions},
  year = {2025},
}

@misc{paquette2026phases,
  title = {Phases of {Muon}: When {Muon} Eclipses {SignSGD}},
  author = {Elliot Paquette and Noah Marshall and Lucas Benigni and Guangyuan Wang and Atish Agarwala and Courtney Paquette},
  year = {2026},
  eprint = {2605.09552},
  archiveprefix = {arXiv},
  primaryclass = {math.OC},
  url = {https://arxiv.org/abs/2605.09552},
}

@article{BenArousGheissariJagannath2024Effective,
  title = {High-dimensional limit theorems for {SGD}: Effective dynamics and critical scaling},
  volume = {77},
  issn = {1097-0312},
  url = {http://dx.doi.org/10.1002/cpa.22169},
  doi = {10.1002/cpa.22169},
  number = {3},
  journal = {Communications on Pure and Applied Mathematics},
  publisher = {Wiley},
  author = {Ben Arous, G{\'e}rard and Gheissari, Reza and Jagannath, Aukosh},
  year = {2024},
  pages = {2030--2080},
}

@book{pavliotis2008multiscale,
  author = {Pavliotis, Grigorios A. and Stuart, Andrew M.},
  title = {Multiscale Methods: Averaging and Homogenization},
  series = {Texts in Applied Mathematics},
  volume = {53},
  publisher = {Springer},
  year = {2008},
  isbn = {9780387738291},
  doi = {10.1007/978-0-387-73829-1},
  url = {https://link.springer.com/book/10.1007/978-0-387-73829-1},
}

@incollection{papanicolaou1977martingale,
  author = {Papanicolaou, G. C. and Stroock, D.
		 and Varadhan, S. R. S.},
  title = {Martingale approach to some limit theorems},
  booktitle = {Duke turbulence conference},
  editor = {Chow, P. L.},
  series = {Duke University Mathematics Series},
  number = {III},
  publisher = {Duke University},
  address = {Durham, NC},
  year = {1977},
  pages = {ii+120 pp.},
  note = {(Durham, NC, April 23--25, 1976). MR:0461684.},
}

@article{Goldt-et-al-Gaussian-Equivalence-Conjecture,
  title = 	 {The Gaussian equivalence of generative models for learning with shallow neural networks},
  author =       {Goldt, Sebastian and Loureiro, Bruno and Reeves, Galen and Krzakala, Florent and Mezard, Marc and Zdeborova, Lenka},
  booktitle = 	 {Proceedings of the 2nd Mathematical and Scientific Machine Learning Conference},
  year = 	 {2022},
}

@article{Goldt-PRX-universality,
  title = {Modeling the Influence of Data Structure on Learning in Neural Networks: The Hidden Manifold Model},
  author = {Goldt, Sebastian and M\'ezard, Marc and Krzakala, Florent and Zdeborov\'a, Lenka},
  journal = {Phys. Rev. X},
  volume = {10},
  issue = {4},
  pages = {041044},
  numpages = {32},
  year = {2020},
  month = {Dec},
  publisher = {American Physical Society},
  doi = {10.1103/PhysRevX.10.041044},
  url = {https://link.aps.org/doi/10.1103/PhysRevX.10.041044}
}

@article{upcoming,
    author={Aukosh Jagannath and Varnan Sarangian},
    title={High-dimensional limit theorems for adaptive variants of SGD II: Diffusive limits} , 
    note={In preparation}
}

@article{arous2020algorithmic,
  title={Algorithmic thresholds for tensor PCA},
  author={Ben Arous, G{\'e}rard and Gheissari, Reza and Jagannath, Aukosh},
  journal={The Annals of Probability},
  volume={48},
  number={4},
  pages={2052--2087},
  year={2020},
  publisher={JSTOR}
}
\bibliographystyle{amsplain}
\end{document}